\documentclass{article}

\PassOptionsToPackage{numbers,sort&compress}{natbib}

\usepackage[main, final]{neurips_2026}

\usepackage[utf8]{inputenc} %
\usepackage[T1]{fontenc}    %
\usepackage{hyperref}       %
\usepackage{url}            %
\usepackage{booktabs}       %
\usepackage{amsfonts}       %
\usepackage{nicefrac}       %
\usepackage{microtype}      %
\usepackage{xcolor}         %

\usepackage{subcaption}

\usepackage{amsmath}
\usepackage{amssymb}
\usepackage{mathtools}
\usepackage{amsthm}
\usepackage{caption}
\usepackage{etoolbox}
\usepackage[framemethod=default]{mdframed}

\usepackage{titletoc}
\usepackage{tocloft} %
\usepackage{bm}

\usepackage{algorithm}        
\usepackage{algorithmic}

\definecolor{nice_gold}{RGB}{191, 128, 64}
\definecolor{nice_blue}{RGB}{21, 100, 220}
\definecolor{nice_pink}{RGB}{237, 2, 140}

\newcommand{\MMD}{\ensuremath{\mathbf{MMD}}}

\newcommand{\E}{\mathbb E}
\DeclareMathOperator*{\Eop}{\mathbb{E}}

\DeclareMathOperator{\Law}{Law}

\newcommand{\R}{\mathbb{R}}

\newif\ifinappendix
\inappendixfalse
\newif\ifmainfancyresults
\mainfancyresultsfalse
\newif\ifmainitalicresults
\mainitalicresultstrue

\newtheoremstyle{mainresultitalicstyle}
  {\topsep}
  {\topsep}
  {\itshape}
  {}
  {\bfseries}
  {.}
  {0.5em}
  {}

\newcommand{\mainresultframestart}{%
  \ifmainfancyresults
    \par\addvspace{\topsep}%
    \begin{mdframed}[%
      topline=false,
      rightline=false,
      bottomline=false,
      leftline=true,
      linewidth=0.8pt,
      linecolor=black,
      backgroundcolor=white,
      skipabove=0pt,
      skipbelow=0pt,
      innerleftmargin=6pt,
      innerrightmargin=0pt,
      innertopmargin=0pt,
      innerbottommargin=0pt,
      roundcorner=0pt
    ]%
  \fi
}

\newcommand{\mainresultframeend}{%
  \ifmainfancyresults
    \end{mdframed}%
    \par\addvspace{\topsep}%
  \fi
}

\theoremstyle{definition}
\newtheorem{maindefinition}{Definition}[section]

\theoremstyle{definition}
\newtheorem{mainlemmaupright}{Lemma}[section]
\newtheorem{mainpropositionupright}{Proposition}[section]
\newtheorem{maintheoremupright}{Theorem}[section]
\newtheorem{maincorollaryTupright}{Corollary}[maintheoremupright]
\newtheorem{maincorollaryDupright}{Corollary}[maindefinition]
\newtheorem{maincorollaryLupright}{Corollary}[mainlemmaupright]

\theoremstyle{mainresultitalicstyle}
\newtheorem{mainlemmaitalic}{Lemma}[section]
\newtheorem{mainpropositionitalic}{Proposition}[section]
\newtheorem{maintheoremitalic}{Theorem}[section]
\newtheorem{maincorollaryTitalic}{Corollary}[maintheoremitalic]
\newtheorem{maincorollaryDitalic}{Corollary}[maindefinition]
\newtheorem{maincorollaryLitalic}{Corollary}[mainlemmaitalic]

\theoremstyle{definition}
\newtheorem{appdefinition}{Definition}[section]
\newtheorem{applemma}{Lemma}[section]
\newtheorem{appproposition}{Proposition}[section]
\newtheorem{apptheorem}{Theorem}[section]
\newtheorem{appcorollaryT}{Corollary}[apptheorem]
\newtheorem{appcorollaryD}{Corollary}[appdefinition]
\newtheorem{appcorollaryL}{Corollary}[applemma]

\newenvironment{definition}[1][]
  {%
    \ifinappendix
      \begin{appdefinition}[#1]%
    \else
      \mainresultframestart
      \begin{maindefinition}[#1]%
    \fi
  }
  {%
    \ifinappendix
      \end{appdefinition}%
    \else
      \end{maindefinition}%
      \mainresultframeend
    \fi
  }

\newenvironment{lemma}[1][]
  {%
    \ifinappendix
      \begin{applemma}[#1]%
    \else
      \mainresultframestart
      \ifmainitalicresults
        \begin{mainlemmaitalic}[#1]%
      \else
        \begin{mainlemmaupright}[#1]%
      \fi
    \fi
  }
  {%
    \ifinappendix
      \end{applemma}%
    \else
      \ifmainitalicresults
        \end{mainlemmaitalic}%
      \else
        \end{mainlemmaupright}%
      \fi
      \mainresultframeend
    \fi
  }

\newenvironment{proposition}[1][]
  {%
    \ifinappendix
      \begin{appproposition}[#1]%
    \else
      \mainresultframestart
      \ifmainitalicresults
        \begin{mainpropositionitalic}[#1]%
      \else
        \begin{mainpropositionupright}[#1]%
      \fi
    \fi
  }
  {%
    \ifinappendix
      \end{appproposition}%
    \else
      \ifmainitalicresults
        \end{mainpropositionitalic}%
      \else
        \end{mainpropositionupright}%
      \fi
      \mainresultframeend
    \fi
  }

\newenvironment{theorem}[1][]
  {%
    \ifinappendix
      \begin{apptheorem}[#1]%
    \else
      \mainresultframestart
      \ifmainitalicresults
        \begin{maintheoremitalic}[#1]%
      \else
        \begin{maintheoremupright}[#1]%
      \fi
    \fi
  }
  {%
    \ifinappendix
      \end{apptheorem}%
    \else
      \ifmainitalicresults
        \end{maintheoremitalic}%
      \else
        \end{maintheoremupright}%
      \fi
      \mainresultframeend
    \fi
  }

\newenvironment{corollaryT}[1][]
  {%
    \ifinappendix
      \begin{appcorollaryT}[#1]%
    \else
      \mainresultframestart
      \ifmainitalicresults
        \begin{maincorollaryTitalic}[#1]%
      \else
        \begin{maincorollaryTupright}[#1]%
      \fi
    \fi
  }
  {%
    \ifinappendix
      \end{appcorollaryT}%
    \else
      \ifmainitalicresults
        \end{maincorollaryTitalic}%
      \else
        \end{maincorollaryTupright}%
      \fi
      \mainresultframeend
    \fi
  }

\pretocmd{\appendix}{%
  \inappendixtrue
}{}{}

\newcommand{\inlinehead}[1]{\noindent\textbf{#1}\hspace{0.8em}}

\title{Bellman Contraction under MMD: \\
A Unified Framework
}

\author{%
  Baptiste Debes \\
  ESAT-PSI \\
  KU Leuven \\
  \texttt{baptiste.debes@kuleuven.be} \\
  \And
  Tinne Tuytelaars \\
  ESAT-PSI \\
  KU Leuven \\
  \texttt{tinne.tuytelaars@kuleuven.be} \\
}

\begin{document}

\maketitle

\begin{abstract}
 We introduce a mathematical framework for studying contraction of the distributional Bellman operator under Maximum Mean Discrepancy (MMD). We show that contraction reduces to conditional positive definiteness of an associated kernel constructed from the base kernel, which we call the Bellman difference kernel. Focusing on radial kernels, we use Schoenberg's theorem to connect this condition to Bernstein functions. This gives contraction criteria that hold in any dimension and can be verified analytically or tested numerically. Our framework unifies known results for the Gaussian RBF, power-distance, and multiquadric kernels. A numerical search identifies 11 new kernels that induce contraction. We also propose \textit{chaining}, a general construction that produces kernels inducing contraction with any prescribed factor. To reconcile theory with practice, we show that contraction on a set only requires conditional positive definiteness on that set. This explains why the RBF kernel can remain a valid choice when rewards are bounded, despite its lack of global contraction. Finally, we identify a fundamental tradeoff between contraction rate and discriminative power, where kernels that contract faster tend to discriminate less between distributions. Code is available at \url{https://github.com/BaptisteDebes/BellmanContractionMMD}
  \end{abstract}

\section{Introduction}
\label{sec:introduction}

Distributional reinforcement learning (DRL) models the distribution over returns instead of their expectations \citep{bellemare2017distributional,bellemare2023distributional}. This in turn makes it possible to account for irreducible uncertainty in transitions and rewards exhibited by some environments \citep{morimura2010nonparametric}. It also enables risk-aware policy improvement \citep{dabney2018implicit, singh2020improving}. Distributional RL has shown substantial performance gains on a wide variety of tasks \citep{barth2018distributed,dabney2018implicit,hessel2018rainbow} and has been motivated theoretically \citep{lyle2019comparative,rowland2018analysis,rowland2019statistics}. A practical instantiation of DRL depends on both how the return distribution is parametrized and how discrepancies between probability distributions are measured \citep{rowland2019statistics}. A natural and flexible choice is to represent the return distribution as a set of equally weighted particles \citep{nguyen2021distributional,singh2022sample,dabney2018implicit}. This representation enables the use of discrepancies computed from samples, such as Maximum Mean Discrepancy (MMD) \citep{JMLR:v13:gretton12a,killingberg2023the,nguyen2021distributional,zhang2021distributional}, Sinkhorn divergence \citep{feydy2019interpolating,sun2024distributional}, or sliced divergences \citep{debes2026multivariate}.

Maximum Mean Discrepancy depends on a kernel that controls which aspects of the distributions are compared and how sensitively. A central question is then whether the \emph{distributional Bellman operator} \citep{bellemare2017distributional} is contractive under the MMD induced by a given kernel, which governs whether the resulting learning objective is principled \citep{bellemare2017distributional,dabney2018distributional}. Despite the flexibility afforded by kernel choice, existing understanding of which kernels induce contraction remains limited. In particular, it provides little guidance for identifying or constructing new ones. This in turn leaves practitioners with few theoretically motivated options.

The first contribution of this work is to connect contraction under MMD with a definiteness property of an associated kernel, namely conditional positive definiteness. For the important class of radial kernels, we leverage Schoenberg’s theorem \citep{schoenberg1938metric}, which links radial kernels to Bernstein functions \citep{schilling2009bernstein,guella2021generalizationenergydistancebernstein}. This perspective allows us to show that contraction can be verified analytically. This unifies known results from \cite{nguyen2021distributional,killingberg2023the,debes2026distributional} under a single framework. More broadly, this framework provides practical recipes for identifying kernels that induce contraction and for constructing new ones, which we demonstrate in this work. Notably, all results hold in arbitrary dimension, directly covering the multivariate setting \citep{zhang2021distributional,wiltzer2024foundations,debes2026multivariate}.

Secondly, a gap remains between theory and practice, as some kernels such as the RBF are known not to induce contraction in general, yet still perform among the best in practice \citep{nguyen2021distributional}. We reconcile theory and practice by showing numerically that such kernels can in fact induce contraction when the return laws are supported on a bounded set. In reinforcement learning, this assumption arises naturally when rewards are bounded, since discounted returns are then bounded as well.

Finally, we identify a tradeoff between the contraction induced by a kernel and its ability to discriminate between distributions. In general, kernels with faster contraction tend to have weaker discriminative power, so these two aspects cannot be optimized independently. We study this tradeoff empirically by measuring the power of kernel two-sample tests.

\section{Background and related work}
\label{sec:background}

We consider a discounted Markov decision process (MDP) \((\mathcal S,\mathcal A,P,R,\gamma)\), where \(\mathcal S\) is the state space, \(\mathcal A\) the action space, \(P(\cdot\mid s,a)\) the transition kernel, \(R(s,a)\) the random reward, and \(\gamma\in(0,1)\) the discount factor. Rewards may be multivariate, in which case returns take values in \(\mathbb R^d\). For a fixed policy \(\pi\), we define the random return as follows, with its expectation yielding the usual state-action value function:
\begin{equation*}
\label{eq:return-rv}
Z^\pi(s,a):=\sum_{t\ge 0}\gamma^t R_t,
\qquad
Q^\pi(s,a):=\E[Z^\pi(s,a)],
\end{equation*}
where \(S_0=s\), \(A_0=a\), \(R_t=R(S_t,A_t)\), \(S_{t+1}\sim P(\cdot\mid S_t,A_t)\) and \(A_{t+1}\sim \pi(\cdot\mid S_{t+1})\). We represent return laws by maps \(\eta:\mathcal S\times\mathcal A\to\mathcal P(\R^d)\). For a fixed policy \(\pi\), the distributional Bellman operator is defined, for \(S'\sim P(\cdot\mid s,a)\), \(A'\sim \pi(\cdot\mid S')\), and \(X'\sim \eta(S',A')\), by
\begin{equation}
\label{eq:distributional-bellman}
(\mathcal T^\pi \eta)(s,a)
:=
\mathrm{Law}\bigl(R(s,a)+\gamma X'\bigr).
\end{equation}
\inlinehead{Policy evaluation.} This work focuses on distributional policy evaluation for a fixed policy \(\pi\), where one seeks to solve or approximate the fixed point equation \(\eta^\pi=\mathcal T^\pi\eta^\pi\). Our goal is to study the contraction properties of the corresponding distributional Bellman operator \(\mathcal T^\pi\), rather than policy improvement or control.

\inlinehead{Maximum Mean Discrepancy.}
A kernel is a symmetric function \(k:\R^d\times\R^d\to\R\). For laws \(\mu,\nu\in\mathcal P(\R^d)\), the squared Maximum Mean Discrepancy (MMD) induced by \(k\) is defined by \citep{JMLR:v13:gretton12a}
\begin{equation}
\label{eq:mmd-squared}
\MMD_k^2(\mu,\nu)
=
\E_{x,x'\sim\mu}[k(x,x')]
+
\E_{y,y'\sim\nu}[k(y,y')]
-
2\,\E_{x\sim\mu,\,y\sim\nu}[k(x,y)].
\end{equation}
Some background on kernels and MMD is provided in Appendix~\ref{app:background}, including the biased estimator of \(\MMD_k^2\) \citep{JMLR:v13:gretton12a}, which is the one commonly used in distributional RL \citep{nguyen2021distributional,killingberg2023the}.

\inlinehead{Contraction.}
Return laws are conditional distributions, so to compare two of them we use the supremum MMD over state-action pairs \citep{bellemare2017distributional,nguyen2021distributional}, defined by
\begin{equation}
\label{eq:sup-mmd-squared}
\overline{\MMD}_k^2(\eta_1,\eta_2)
:=
\sup_{(s,a)\in \mathcal S\times\mathcal A}
\MMD_k^2\bigl(\eta_1(s,a),\eta_2(s,a)\bigr).
\end{equation}
For \(\tau\in[0,1)\), we say that \(\mathcal T^\pi\) is a \(\tau\)-contraction in \(\overline{\MMD}_k^2\) if
\[
\overline{\MMD}_k^2(\mathcal T^\pi \eta_1,\mathcal T^\pi \eta_2)
\le
\tau\,\overline{\MMD}_k^2(\eta_1,\eta_2)
\qquad
\forall\,\eta_1,\eta_2.
\]
We call any such \(\tau\) a \emph{contraction factor} of \(\mathcal T^\pi\) and denote by \(\tau_k^\star(\mathcal T^\pi)\) the smallest one valid for every MDP with discount \(\gamma\) and every fixed policy. Throughout this work, all contraction factors refer to \(\overline{\MMD}_k^2\). The corresponding smallest contraction factor in \(\overline{\MMD}_k\) is the square root \(\sqrt{\tau_k^\star(\mathcal T^\pi)}\).

In policy evaluation, \(\tau_k^\star(\mathcal T^\pi)<1\) guarantees that the distributional Bellman equation has a unique solution \(\eta^\pi\). Repeated Bellman updates then converge to that solution in the supremum MMD at a geometric rate, \(\overline{\MMD}_k^2\bigl((\mathcal T^\pi)^n\eta,\eta^\pi\bigr)\le\tau_k^\star(\mathcal T^\pi)^n\,\overline{\MMD}_k^2(\eta,\eta^\pi)\) for every \(\eta\) and every \(n\ge 1\). More broadly, contraction is the standard way to show that the distributional Bellman equation is well posed under a given discrepancy \citep{bellemare2017distributional,bellemare2023distributional,rowland2019statistics}.

\inlinehead{Related work and known results.}
Existing work on distributional RL with MMD has established contraction results for a small set of kernels \citep{nguyen2021distributional,killingberg2023the}. In the univariate setting, sufficient conditions for Bellman contraction were established for kernels that are both shift invariant and scale sensitive of order \(\alpha>0\), meaning that for all \(x,y,c\) in the underlying space \citep{nguyen2021distributional},
\begin{equation}
\label{eq:shift-scale-kernel}
k(x+c,y+c)=k(x,y), \qquad k(cx,cy)=|c|^\alpha k(x,y).
\end{equation}
Table~\ref{tab:known_mmd_kernels} summarizes the main contraction results known for specific kernels prior to this work. Distributional RL with MMD was subsequently extended to multivariate rewards \citep{zhang2021distributional,wiltzer2024foundations}. The earlier of the two minimized MMD but proved contraction of the multivariate distributional Bellman operator only in the Wasserstein metric \citep{zhang2021distributional}, leaving contraction in MMD open. A later work laying the foundations of multivariate distributional RL proved contraction in MMD, but only for kernels satisfying the conditions of Equation~\ref{eq:shift-scale-kernel} \citep{wiltzer2024foundations}. Taken together, these works identify a small set of kernels for which contraction can be proved or disproved, but do not provide a general criterion that unifies these results. Providing such a framework is among the contributions of this work.

\begin{table}[t]
\centering
\small
\setlength{\tabcolsep}{5pt}
\renewcommand{\arraystretch}{1.15}
\begin{tabular}{p{0.26\linewidth}p{0.36\linewidth}p{0.24\linewidth}}
\toprule
Kernel & Formula & Contraction in \(\overline{\MMD}_k^2\) \\
\midrule
Power-distance \citep{nguyen2021distributional}
&
\(
k_\alpha(x,y)=-\|x-y\|^\alpha,
\quad 0<\alpha\le 2
\)
&
Yes, factor \(\gamma^\alpha\) \\
[0.4em]

Gaussian RBF \citep{nguyen2021distributional}
&
\(
k_{\mathrm{rbf}}(x,y)=\exp(-h^2\|x-y\|^2)
\)
&
No in general \\
[0.4em]

Multiquadric \citep{killingberg2023the,debes2026distributional}
&
\(
k_{\mathrm{mq},h}(x,y)=-\sqrt{1+h^2\|x-y\|^2}
\)
&
Yes, factor \(\gamma\) \\
\bottomrule
\end{tabular}
\caption{Known contraction results for specific kernels in distributional RL with MMD prior to this work. The multiquadric corresponds to the case \(\beta=1/2\) of the generalized multiquadric \(k_{\beta,h}(x,y)=-(1+h^2\|x-y\|^2)^\beta\).}
\label{tab:known_mmd_kernels}
\end{table}

\section{Contraction in Maximum Mean Discrepancy}
\label{sec:main_contraction_mmd}
In this section, we establish our first main result, showing that contraction of the distributional Bellman operator in \(\overline{\MMD}_k^2\) reduces to a conditional positive definiteness condition on a kernel derived from \(k\). We begin by recalling the notion of conditional positive definiteness.

\begin{definition}\label{def:main-cpd}
Let $k:X\times X\to\R$ be symmetric. We say that $k$ is \emph{conditionally positive definite} (CPD), or \emph{conditionally negative definite} (CND), if for every $n\ge 1$, every $x_1,\dots,x_n\in X$, and every $w_1,\dots,w_n\in\R$ with $\sum_{i=1}^n w_i=0$, one has
\begin{equation*}
\sum_{i,j=1}^n w_i w_j\,k(x_i,x_j)\ge 0
\qquad
(\text{resp. } \le 0).
\end{equation*}
\end{definition}
Theorem~\ref{thm:cpd_contraction} below makes precise the reduction of Bellman contraction to a CPD condition. Conditional positive definiteness is well studied, especially for radial kernels, as we exploit in Section~\ref{sec:main_contraction_radial_kernels}. The proof is deferred to Appendix~\ref{app:general_contraction}.
\begin{theorem}
\label{thm:cpd_contraction}
Let \(k:\R^d\times\R^d\to\R\) be a CPD kernel. For a reward value \(q\in\R^d\) and a factor \(\tau\in[0,1)\), define the \emph{Bellman difference kernel}
\begin{equation}
\label{eq:bellman-difference-kernel}
k_{\mathrm{diff}}^{q,\gamma}(x,y):=\tau k(x,y)-k(q+\gamma x,q+\gamma y).
\end{equation}
Then the following two statements are equivalent.
(i)~\(\mathcal T^\pi\) is a \(\tau\)-contraction in \(\overline{\MMD}_k^2\) for every MDP with discount \(\gamma\) and every fixed policy. (ii)~The kernel \(k_{\mathrm{diff}}^{q,\gamma}\) is CPD for every reward value \(q\in\R^d\).
In particular, the smallest contraction factor \(\tau_k^\star(\mathcal T^\pi)\) is the smallest \(\tau\) for which (ii) holds.
\end{theorem}
The CPD condition is therefore sufficient and necessary for \(\mathcal T^\pi\) to be a \(\tau\)-contraction for every MDP with discount \(\gamma\) and every fixed policy. If it fails, there exist an MDP with discount \(\gamma\) and a fixed policy for which \(\mathcal T^\pi\) is not a \(\tau\)-contraction.

\section{Contraction with radial kernels}
\label{sec:main_contraction_radial_kernels}
A kernel \(k:\R^d\times\R^d\to\R\) is called \emph{radial} if \(k(x,y)=g(\|x-y\|)\) for some function \(g:[0,\infty)\to\R\). We focus on this class for two reasons. First, it covers the main kernels studied in distributional RL with MMD, see Table~\ref{tab:known_mmd_kernels}. Second, radial kernels admit powerful contraction results through Schoenberg's theorem \citep{schoenberg1938metric} and the theory of Bernstein functions \citep{schilling2009bernstein}, which we now develop.

\begin{figure}[H]
\centering
\noindent\makebox[\linewidth][c]{%
  \begin{minipage}[t]{0.55\linewidth}
  \vspace{0pt}
  \begin{definition}
  \label{def:main_bernstein}
  A function \(\phi:[0,\infty)\to[0,\infty)\) is a \emph{Bernstein function} if it admits the Lévy--Khintchine representation
  \[
  \phi(t)=a+b\,t+\int_{(0,\infty)} \bigl(1-e^{-st}\bigr)\,\Lambda(ds),\;\;\; t\ge 0
  \]
  where \(a,b\ge 0\) and \(\Lambda\) is a positive Borel measure on \((0,\infty)\), called the Lévy measure of \(\phi\), with \(\int_{(0,\infty)} \min\{1,s\}\,\Lambda(ds)<\infty\). It is a \emph{complete Bernstein function} if, moreover, \(\Lambda(ds)=m(s)\,ds\) for a completely monotone density \(m:(0,\infty)\to[0,\infty)\). In our setting \(\phi(0)=a=0\).
  \end{definition}
  \end{minipage}\hfill
  \begin{minipage}[t]{0.43\linewidth}
  \vspace{0pt}
  \centering
  \includegraphics[width=\linewidth]{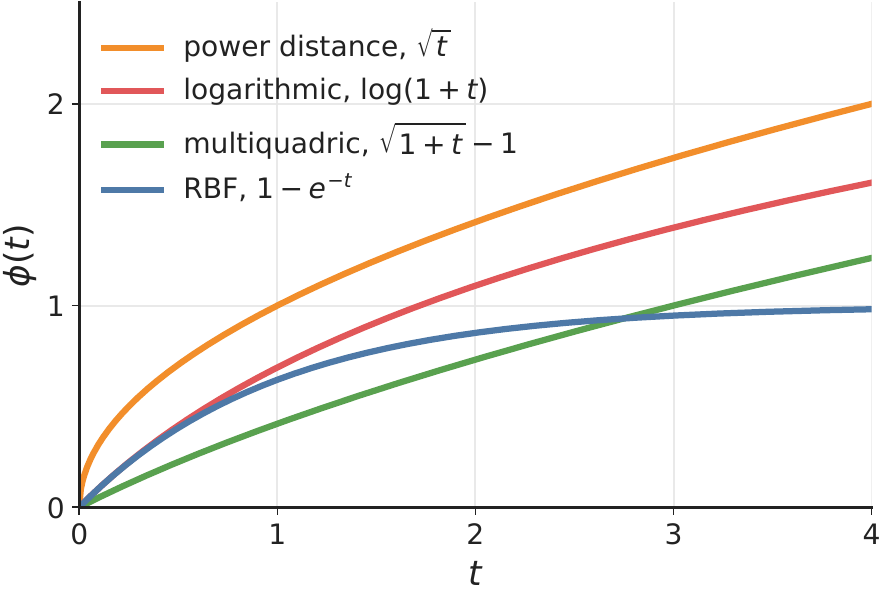}
  \captionsetup{skip=2pt}
  \caption{Examples of Bernstein functions.}
  \label{fig:main_bernstein_functions}
  \end{minipage}%
}
\end{figure}

We now make the bridge from the CPD condition in Theorem~\ref{thm:cpd_contraction} to the Bernstein reduction that follows. To this end, we state Schoenberg's theorem on radial kernels \citep{schoenberg1938metric}, and use the elementary fact that CPD of a kernel \(k\) is equivalent to CND of \(-k\).
\begin{theorem}[Schoenberg \citep{schoenberg1938metric}]
\label{thm:schoenberg-main}
A kernel of the form \((x,y)\mapsto h(\|x-y\|)\), with \(h\) continuous and \(h(0)=0\), is CND on \(\R^d\) for every \(d\ge 1\) if and only if \(h(r)=\phi(r^2)\) for some Bernstein function \(\phi\).
\end{theorem}

\inlinehead{Bernstein reduction.}
The next theorem applies Schoenberg's theorem to the Bellman difference kernel. Its proof is given in Appendix~\ref{app:bernstein_reduction_to_contraction}.

\begin{theorem}
\label{thm:bernstein-reduction-main}
Let \(k(u,v)=g(\|u-v\|)\) be a radial kernel, and assume that \(h(r):=g(0)-g(r)\) can be written as \(h(r)=\phi(r^2)\) for a Bernstein function \(\phi\), which we call the \emph{Bernstein profile} of \(k\). For \(\tau\in[0,1)\), define the \emph{Bellman difference profile}
\begin{equation}
\label{eq:psi-tau-main}
\psi_\tau(t):=\tau\phi(t)-\phi(\gamma^2 t), \qquad t\ge 0.
\end{equation}
If \(\psi_\tau\) is a Bernstein function, then \(\mathcal T^\pi\) is a \(\tau\)-contraction in \(\overline{\MMD}_k^2\). Furthermore, this condition is necessary for \(\tau\)-contraction to hold for every MDP with discount \(\gamma\), every fixed policy, and every ambient dimension \(d\ge 1\).
\end{theorem}
The necessity part of Theorem~\ref{thm:bernstein-reduction-main} concerns every dimension at once. If \(\psi_\tau\) is not a Bernstein function, the Bellman difference kernel fails to be CPD in at least one dimension, but it may still be CPD in others. Theorem~\ref{thm:cpd_contraction} then remains the right tool to check contraction in a given dimension.

\inlinehead{Analytic reduction.}
Theorem~\ref{thm:bernstein-reduction-main} reduces Bellman contraction for radial kernels to checking whether the Bellman difference profile \(\psi_\tau\) is a Bernstein function. Theorem~\ref{thm:levy-side-main} makes this condition explicit through the Lévy--Khintchine representation of the Bernstein profile \(\phi\). In the complete Bernstein case, it also yields an explicit upper bound on the smallest contraction factor. The argument is developed in Appendix~\ref{app:bernstein_reduction_of_difference_kernel}--\ref{app:contraction_complete_bernstein}.

\begin{theorem}
\label{thm:levy-side-main}
Adopt the setting of Theorem~\ref{thm:bernstein-reduction-main}, and let \(b\) and \(\Lambda\) be the components of the Lévy--Khintchine representation of \(\phi\) from Definition~\ref{def:main_bernstein}. Then the Bellman difference profile \(\psi_\tau\) is a Bernstein function if and only if
\[
b(\tau-\gamma^2)\ge 0
\qquad\text{and}\qquad
\tau\,\Lambda(B)\ge \Lambda(B/\gamma^2)
\quad\text{for every Borel }B\subset(0,\infty),
\]
where \(B/\gamma^2:=\{s>0:\gamma^2 s\in B\}\). In particular, if these conditions hold for some \(\tau\in[0,1)\), then \(\mathcal T^\pi\) is a \(\tau\)-contraction in \(\overline{\MMD}_k^2\). Furthermore, if \(\phi\) is a complete Bernstein function, so that \(\Lambda(ds)=m(s)\,ds\) for a density \(m\), then the smallest contraction factor satisfies
\[
\tau_k^\star(\mathcal T^\pi)\le \overline\tau_{\mathrm{cbf}}(\phi,\gamma),
\qquad
\overline\tau_{\mathrm{cbf}}(\phi,\gamma):=
\max\!\left\{
\gamma^2\,\mathbf 1_{\{b>0\}},
\ \sup_{s>0}\frac{m(s/\gamma^2)}{\gamma^2 m(s)}
\right\}.
\]
\end{theorem}

Theorems~\ref{thm:bernstein-reduction-main} and~\ref{thm:levy-side-main} form the backbone of the remainder of this work, unifying known contraction results under a single criterion and providing the tools that the next sections use to identify and construct new kernels that induce contraction.

\inlinehead{Explaining known results.}
Applying Theorem~\ref{thm:levy-side-main} to the power-distance and multiquadric kernels of \citep{nguyen2021distributional,killingberg2023the,debes2026distributional} recovers their contraction results from a single analytic criterion and extends the multiquadric result from \(\beta=1/2\) to the full generalized range \(0<\beta\le 1\). In both cases the radial profile is a complete Bernstein function, so contraction reduces to evaluating \(\overline{\tau}_{\mathrm{cbf}}(\phi,\gamma)\) directly. Proofs are deferred to Propositions~\ref{prop:power_distance_contraction} and~\ref{prop:gmq_contraction} in the appendix.

\begin{proposition}
\label{prop:main_known_positive_kernels}
Consider the power-distance kernels, with parameter \(0<\alpha\le 2\), and the generalized multiquadric kernels, with parameters \(0<\beta\le 1\) and \(h>0\):
\[
k_\alpha(x,y)=-\|x-y\|^\alpha,
\qquad\text{and}\qquad
k_{\beta,h}(x,y)=-\bigl(1+h^2\|x-y\|^2\bigr)^\beta.
\]
Their radial profiles are generated by the Bernstein functions
\[
\phi_\alpha(t)=t^{\alpha/2},
\qquad\text{and}\qquad
\phi_{\beta,h}(t)=(1+h^2t)^\beta-1.
\]
Moreover, both \(\phi_\alpha\) and \(\phi_{\beta,h}\) are complete Bernstein functions, and evaluating the bound \(\overline{\tau}_{\mathrm{cbf}}(\phi,\gamma)\) from Theorem~\ref{thm:levy-side-main} yields
\[
\overline{\tau}_{\mathrm{cbf}}(\phi_\alpha,\gamma)=\gamma^\alpha,
\qquad\text{and}\qquad
\overline{\tau}_{\mathrm{cbf}}(\phi_{\beta,h},\gamma)=\gamma^{2\beta}.
\]
Hence \(\mathcal T^\pi\) is a \(\gamma^\alpha\)-contraction in \(\overline{\MMD}_k^2\) for the power-distance kernels, and a \(\gamma^{2\beta}\)-contraction for the generalized multiquadric kernels.
\end{proposition}

\inlinehead{A first obstruction: saturating profiles.}
Proposition~\ref{prop:main_saturating_obstruction} gives a first explanation for the non-contraction of the RBF kernel established in \citep{nguyen2021distributional}. Saturating Bernstein profiles rule out contraction, and since the RBF profile shown in Figure~\ref{fig:main_bernstein_functions} is saturating, the obstruction applies directly. The proof is given in Proposition~\ref{prop:no_contraction_saturating_BF} in the appendix.
\begin{proposition}
\label{prop:main_saturating_obstruction}
Adopt the setting of Theorem~\ref{thm:bernstein-reduction-main}, and assume that the Bernstein profile \(\phi\) is saturating, meaning that \(\lim_{t\to\infty}\phi(t)<\infty\). If \(\phi\not\equiv 0\), then there is no \(\tau\in(0,1)\) such that \(\tau\)-contraction holds for every MDP with discount \(\gamma\) and every fixed policy.
\end{proposition}

\inlinehead{A second obstruction: finitely many atoms.}
The analytic criterion from Theorem~\ref{thm:levy-side-main} also rules out kernels whose Lévy measure has finitely many atoms, that is, finitely many Dirac masses, as stated in Proposition~\ref{prop:main_finite_atomic_obstruction}. This gives a second explanation for the failure of the RBF kernel. Its Bernstein profile is \(\phi_{\mathrm{rbf},h}(t)=1-\exp(-h^2 t)\), whose Lévy measure \(\Lambda=\delta_{h^2}\) has a single atom at \(h^2\).
\begin{proposition}
\label{prop:main_finite_atomic_obstruction}
Adopt the setting of Theorem~\ref{thm:levy-side-main}, and let \(\Lambda\) be the Lévy measure of \(\phi\). If \(\Lambda\) has at least one but finitely many atoms, then there is no \(\tau\in(0,1)\) such that \(\tau\)-contraction holds for every MDP with discount \(\gamma\), every fixed policy, and every ambient dimension \(d\ge 1\).
\end{proposition}
The same argument also shows that no finite sum of RBF kernels, such as those used in \citep{nguyen2021distributional,zhang2021distributional}, induces contraction. Later, we will show that this obstruction disappears for suitable infinite sums of RBF kernels, which can in fact induce contraction. We refer to Lemma~\ref{lem:finite-atoms} and Theorem~\ref{thm:contraction-drift-density-atoms} in the appendix for the proof of Proposition~\ref{prop:main_finite_atomic_obstruction}, and to Proposition~\ref{prop:rbf_no_contraction} for the case of the RBF kernel.

\section{Contraction on a bounded support}
\label{sec:bounded_support}
The contraction theory developed in Sections~\ref{sec:main_contraction_mmd} and \ref{sec:main_contraction_radial_kernels} requires the Bellman difference kernel to be CPD on the whole ambient space \(\R^d\), which can be unnecessarily restrictive in practice. The RBF kernel, which fails this global condition yet performs well empirically \citep{nguyen2021distributional,zhang2021distributional}, raises the question of whether contraction can hold under weaker conditions. In this section we show that under bounded rewards, CPD need only hold on the bounded region where return laws actually live.

In practical reinforcement learning settings, rewards can often be assumed bounded. In that case, the following proposition gives a bounded set that contains all discounted returns, and shows that return laws supported on this set remain supported on it after a Bellman update. The proof is given in Appendix~\ref{app:contraction_bounded_support}.

\begin{proposition}
\label{prop:main_bounded_return_ball}
Assume the rewards are bounded by \(r_{\max}\), and let \(E\) be the ball of radius \(M\) centered at the origin, \(E:=\{x\in\R^d:\|x\|\le M\}\), with \(M\ge r_{\max}/(1-\gamma)\). Then every return law \(\eta^\pi(s,a)\) is supported on \(E\). Moreover, if \(\eta(s,a)\) is supported on \(E\) for every \((s,a)\), then \((\mathcal T^\pi\eta)(s,a)\) is also supported on \(E\) for every \((s,a)\).
\end{proposition}

This motivates a notion of contraction restricted to return laws supported on \(E\), for which the difference kernel only needs to be CPD on \(E\). The next theorem gives the corresponding criterion for Bellman contraction. 

\begin{theorem}
\label{thm:main_local_support_contraction}
Adopt the setting of Proposition~\ref{prop:main_bounded_return_ball}, so that the return laws are supported on \(E:=\{x\in\R^d:\|x\|\le M\}\) with \(M\ge r_{\max}/(1-\gamma)\). Let \(k\) be a CPD kernel. Fix \(\tau\in[0,1)\). For a reward value \(q\), the Bellman difference kernel \(k_{\mathrm{diff}}^{q,\gamma}\) of Equation~\ref{eq:bellman-difference-kernel} is CPD on \(E\) if for every \(n\ge 1\), every \(x_1,\dots,x_n\in E\),
\[
\sum_{i,j=1}^n w_i w_j\,k_{\mathrm{diff}}^{q,\gamma}(x_i,x_j)\ge 0
\qquad
\forall\,w_1,\dots,w_n\in\R \text{ with }\sum_{i=1}^n w_i=0.
\]
Then the following two statements are equivalent.
(i)~\(\mathcal T^\pi\) is a \(\tau\)-contraction in \(\overline{\MMD}_k^2\) when restricted to return laws supported on \(E\), for every fixed policy and every MDP with discount \(\gamma\) whose rewards are bounded by \(r_{\max}\). (ii)~The kernel \(k_{\mathrm{diff}}^{q,\gamma}\) is CPD on \(E\) for every reward value \(q\) with \(\|q\|\le r_{\max}\).
\end{theorem}

Under this weaker notion, kernels that fail the global condition may still be contractive on bounded sets. The proof of Theorem~\ref{thm:main_local_support_contraction} is given in Appendix~\ref{app:contraction_bounded_support}. We now turn to testing the CPD condition on the ball \(E\) numerically, using the RBF kernel as a practical example.

\inlinehead{Reduction to a generalized eigenvalue problem.}
Assuming bounded rewards, return laws are supported on a fixed bounded set \(E\), so contraction reduces to checking that the associated Bellman difference kernel is CPD on \(E\) rather than on all of \(\R^d\). While this condition cannot be verified numerically on all of \(E\), we can test it on finite subsets of \(E\). As shown in Lemma~\ref{lem:cpd_restriction_monotonicity}, CPD on \(E\) implies CPD on every finite subset of \(E\), so if the CPD condition fails on some finite set \(S\subseteq E\), contraction on \(E\) cannot hold. Moreover, a larger finite subset gives a stricter test, because the condition on a set implies the condition on each of its subsets. On a finite set of points, computing the smallest \(\tau\) for which the CPD condition holds reduces to a generalized eigenvalue problem \citep{parlett1998symmetric}, which can be solved efficiently. The next proposition makes this reduction explicit. Further background and derivation are given in Appendix~\ref{sec:bounded_support_numerical_experiments}.

\begin{proposition}
\label{prop:main_finite_support_gep}
Adopt the setting of Theorem~\ref{thm:main_local_support_contraction}, and assume that the kernel \(k\) is radial. Let \(S=\{z_1,\dots,z_N\}\subseteq E\), let \(Q\in\R^{N\times(N-1)}\) be a matrix whose columns form an orthonormal basis of \(\mathbf 1_N^\perp:=\{w\in\R^N:\mathbf 1_N^\top w=0\}\), and define
\[
K_{ij}:=k(z_i,z_j),
\qquad
(K_\gamma)_{ij}:=k(\gamma z_i,\gamma z_j),
\qquad
A:=Q^\top KQ,
\qquad
B:=Q^\top K_\gamma Q.
\]
Then the following two statements are equivalent.
(i)~The Bellman difference kernel \(k_{\mathrm{diff}}^{q,\gamma}\) is CPD on \(S\). (ii)~The matrix \(\tau A-B\) is positive semidefinite. If \(A\) is strictly positive definite, the smallest value of \(\tau\) for which (ii) holds is the largest generalized eigenvalue \(\lambda_{\max}(B,A)\) of the pair \((B,A)\). We call it the contraction threshold of \(S\):
\[
\tau_S^\star(\gamma)
=
\lambda_{\max}(B,A)
=
\sup_{v\neq 0}\frac{v^\top Bv}{v^\top Av}.
\]
In particular, if \(\tau<\tau_S^\star(\gamma)\) for some \(S\subseteq E\), then, by Theorem~\ref{thm:main_local_support_contraction}, Bellman \(\tau\)-contraction on return laws supported on \(E\) cannot hold.
\end{proposition}

\begin{figure}[t]
    \centering
    \includegraphics[width=0.48\linewidth]{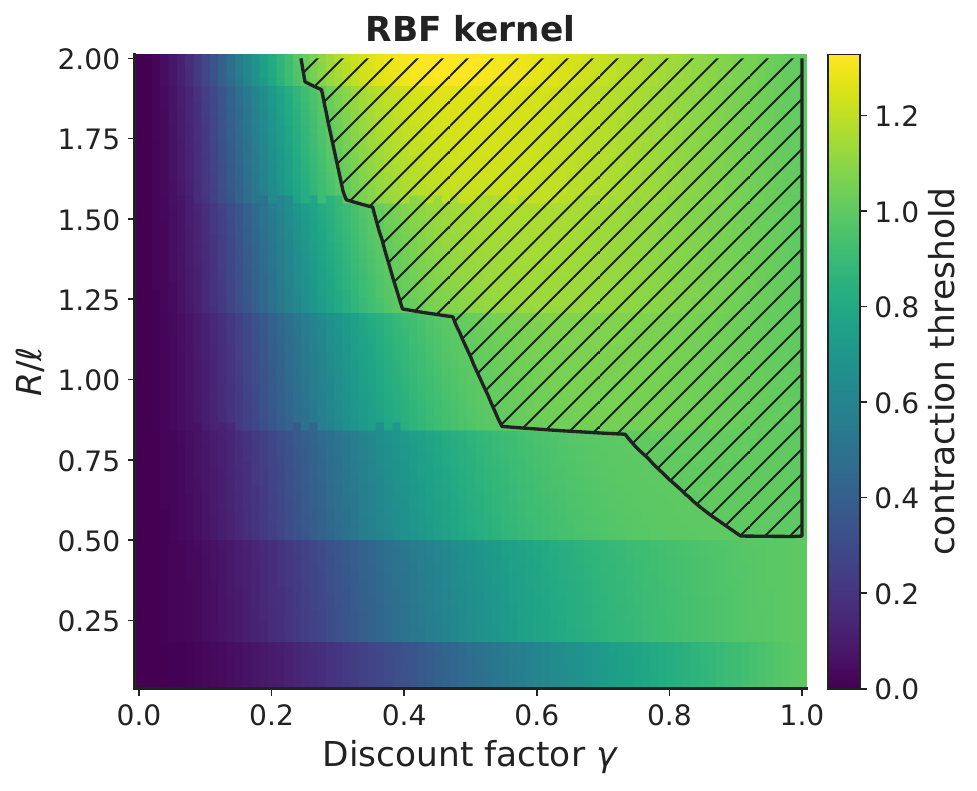}
    \hfill
    \includegraphics[width=0.48\linewidth]{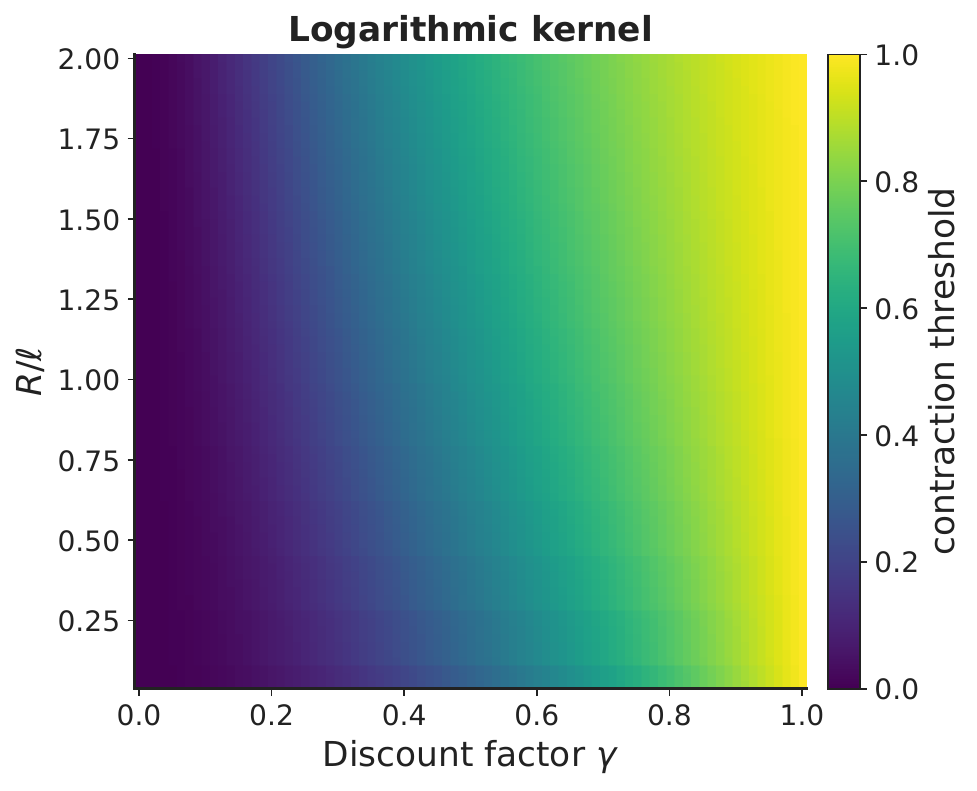}
    \caption{Contraction thresholds on a ball of radius \(R\), for the RBF kernel \(\exp(-\|x-y\|^2/\ell^2)\) (left) and the logarithmic kernel \(-\log(1+\|x-y\|^2/\ell^2)\) (right). For each discount \(\gamma\) and each ratio \(R/\ell\), the color shows the largest contraction threshold \(\tau_S^\star(\gamma)\) of Proposition~\ref{prop:main_finite_support_gep} over \(10\) sets \(S\) of points drawn at random in \(B(0,R)\), in dimension~1. In the hatched region, above the black line, this value exceeds \(1\), so contraction on the ball cannot hold. Where it stays below \(1\), the test does not rule out contraction. The RBF kernel passes the test when \(R/\ell\) is small, even though it does not induce contraction on all of \(\R^d\).}
    \label{fig:main_bounded_support_heatmaps}
\end{figure}

\inlinehead{Numerical experiments and contraction thresholds.}
We now use Proposition~\ref{prop:main_finite_support_gep} to study contraction on a bounded set numerically. For each trial, we draw a set \(S=\{z_1,\dots,z_N\}\) of \(N\) points uniformly at random in the ball \(B(0,R)\), in dimension~1, and compute \(\tau_S^\star(\gamma)\) by solving the generalized eigenvalue problem via Algorithm~\ref{alg:tau-star-mc}. Figure~\ref{fig:main_bounded_support_heatmaps} reports the observed contraction threshold as a function of the discount \(\gamma\) and the ratio \(R/\ell\). For the RBF kernel, the threshold stays below \(1\) when \(R/\ell\) is small enough, which suggests that the kernel induces contraction on a ball of radius \(R\) even though it does not in general. This is numerical evidence on finite sets of points, not a proof on the whole ball. Finding a condition on \(R/\ell\) that guarantees contraction remains an open question. To show that the same method applies beyond the RBF case, we also run it on the logarithmic kernel \(k_\ell(x,y)=-\log\!\left(1+\|x-y\|^2/\ell^2\right)\), which is non-expansive by Proposition~\ref{prop:three_kernels_no_contraction}(iii) and can become contractive on bounded supports. Further details are given in Appendix~\ref{sec:bounded_support_numerical_experiments}.

\section{New kernels inducing contraction}

The criterion of Theorem~\ref{thm:levy-side-main} opens the door to a systematic search for kernels that induce contraction, beyond the two families previously known. Since the kernel determines which aspects of the distributions the MMD objective is sensitive to \citep{sejdinovic2013equivalence}, a broader catalogue allows practitioners to make that choice deliberately rather than by default. In this section we develop two complementary approaches.

\inlinehead{Searching over complete Bernstein functions.}
In the complete Bernstein case, Theorem~\ref{thm:levy-side-main} reduces the contraction test to a one-dimensional supremum that depends only on the L\'evy density \(m(s)\). Standard references such as \citet{schilling2009bernstein} compile many complete Bernstein functions together with their L\'evy densities. We extracted \(43\) entries from this reference, selected as described in Appendix~\ref{app:cbf_search} and listed in Tables~\ref{tab:cbf_screen_candidates_a}--\ref{tab:cbf_screen_candidates_d}. Because the resulting list is too large to handle analytically case by case, we estimate the supremum numerically with the one-dimensional search summarized in Algorithm~\ref{alg:cbf_sup_search}. Appendix~\ref{app:cbf_search} validates this search against the exact contraction factors of Proposition~\ref{prop:main_known_positive_kernels}. This search identifies \(11\) additional kernel families that induce contraction, collected in Table~\ref{tab:accepted-bernstein-functions}. Figure~\ref{fig:cbf_accepted_tau_star_curves} shows their estimated contraction factors. Some of these families are parametric, and contraction holds only on part of the admissible parameter range, as Figure~\ref{fig:cbf_parametric_regimes} in Appendix~\ref{app:positive_results_cmpl} shows.

\inlinehead{Chaining: a construction with prescribed contraction factor.}
In Section~\ref{sec:main_contraction_radial_kernels}, we showed that the RBF kernel does not induce contraction. Surprisingly, a carefully constructed infinite sum of RBF kernels can. We call this construction \emph{chaining} and show that given a discount factor \(\gamma\), it can selectively induce contraction with any prescribed factor in \((\gamma^2, 1)\).
\begin{theorem}
\label{thm:main_gaussian_chain}
Let \(\tau\in(\gamma^2,1)\) be a target contraction factor, and fix a scale \(a>0\) and a weight \(c>0\). For a lengthscale \(\ell>0\), let \(k_{\mathrm{rbf},\ell}(x,y)=\exp(-\|x-y\|^2/\ell^2)\) be the RBF kernel with lengthscale \(\ell\), and define the \emph{chained RBF kernel}
\[
k_{\mathrm{chain}}(x,y):=\sum_{n=0}^{\infty} c\,\tau^{-n}\bigl(k_{\mathrm{rbf},\ell_n}(x,y)-1\bigr),\qquad \ell_n:=a^{-1/2}\gamma^{-n}.
\]
Then \(\mathcal T^\pi\) is a \(\tau\)-contraction in \(\overline{\MMD}_{k_{\mathrm{chain}}}^2\) for every MDP with discount \(\gamma\) and every fixed policy. Moreover, no smaller factor gives a contraction for every MDP with discount \(\gamma\), every fixed policy and every ambient dimension \(d\ge 1\).
\end{theorem}
Its proof combines Theorem~\ref{thm:cpd_contraction} and Proposition~\ref{prop:gaussian_geometric_chaining_series} in the appendix. Proposition~\ref{prop:kernel_geometric_chaining_factory} applies the same construction to kernels other than the RBF. Figures~\ref{fig:gaussian_chain_tau_bernstein}--\ref{fig:gaussian_chain_tau_levy} plot the Bernstein function and the Lévy measure of the chained RBF kernel for three values of \(\tau\). The construction combines RBF kernels whose lengthscales and weights both grow. Since \(\gamma<1\), the lengthscales grow, \(\ell_{n+1}=\gamma^{-1}\ell_n>\ell_n\), so each new term is sensitive to larger distances between returns. Since \(\tau<1\), the weights grow as well, \(c\tau^{-(n+1)}=\tau^{-1}c\tau^{-n}>c\tau^{-n}\), so each new term counts more. In practice, the sum must be truncated, and a few hundred terms are enough to make the truncation error small, as shown numerically in Appendix~\ref{app:chaining_numerical}.

The chaining construction produces a kernel that induces contraction with any prescribed factor, making it a principled new option for distributional RL with MMD. However, a faster contraction rate does not automatically make a kernel more desirable, as we discuss in Section~\ref{sec:tradeoff}. 

\section{A tradeoff perspective}\label{sec:tradeoff}

The preceding sections provided tools for characterizing the smallest contraction factor of a given kernel across dimensions. One might therefore be tempted to compare kernels through this quantity alone and to conclude that the kernel with the smallest contraction factor is the most desirable. This conclusion would be misleading. Changing the kernel also changes the discrepancy itself, so a smaller contraction factor need not translate into better practical behavior when the induced \(\MMD\) is less sensitive to distributional differences that matter in practice. This section makes that tension precise and quantifies it empirically through the lens of statistical testing. Within the family of kernels induced by Bernstein functions, we show that the kernels attaining the smallest contraction factors are precisely those whose induced \(\MMD\) is least informative. We begin by recalling what it means for \(\MMD_k\) to be a metric.
\begin{definition}
A kernel \(k\) is said to induce a metric between probability distributions if \(\MMD_k(P,Q)=0\) implies \(P=Q\) for all \(P,Q\in\mathcal P(X)\).
\end{definition}
For the radial kernels considered in this work, which are built from Bernstein functions, the next two propositions show that these kernels induce a metric except in one case, where the resulting discrepancy reduces to mean comparison. The proofs are given in Appendix~\ref{app:tradeoff_detectability} and Appendix~\ref{app:properties_bernstein_based_kernels}.
\begin{proposition}\label{prop:main_bernstein_mmd_metric}
Let \(k(x,y)=C-\phi(\|x-y\|^2)\), where \(\phi\) is a Bernstein function with Lévy measure \(\Lambda\). Then \(\MMD_k\) is a metric if and only if \(\Lambda\not\equiv 0\). Equivalently, \(\MMD_k\) fails to be a metric exactly when \(\phi(t)=bt\), that is, when \(k(x,y)=C-b\|x-y\|^2\).
\end{proposition}
\begin{proposition}
Let \(k(x,y)=C-b\|x-y\|^2\) with \(b\ge 0\). Then \(\MMD_k^2(P,Q)=2b\|\E_P[X]-\E_Q[X]\|^2\). In particular, \(\MMD_k^2(P,Q)=0\) whenever \(P\) and \(Q\) have the same mean.
\end{proposition}
In this degenerate case, distributional policy evaluation reduces to comparing expected returns, as in classical Q-learning \citep{watkins1992q}. The next proposition is the first illustration of the tension discussed in this section, as it shows that the smallest possible contraction factor \(\gamma^2\) is attained only in this degenerate case. Its proof is given in the appendix as Proposition~\ref{prop:nonzero_levy_strictly_above_drift}.
\begin{proposition}\label{prop:main_nonzero_levy_above_drift}
Let \(k(x,y)=C-\phi(\|x-y\|^2)\), where \(\phi\) is a Bernstein function with Lévy measure \(\Lambda\). If \(\Lambda\not\equiv 0\), that is, if \(\MMD_k\) is a metric (Proposition~\ref{prop:main_bernstein_mmd_metric}), then \(\tau_k^\star(\mathcal T^\pi)>\gamma^2\), so the kernel cannot reach the smallest possible contraction factor \(\gamma^2\). Equivalently, this value can only be reached by the kernel \(k(x,y)=C-b\|x-y\|^2\).
\end{proposition}
\inlinehead{Empirical detectability.}
Beyond its role as a discrepancy, MMD can be used as a test statistic to decide whether two distributions differ \citep{borgwardt2006integrating,gretton2006kernel,JMLR:v13:gretton12a}. Run with each kernel, such a test offers a principled way to compare kernels in practice, through the probability that it detects a difference. In distributional policy evaluation, this means detecting a statistically significant difference between predicted and target return laws from a limited number of samples.
\begin{definition}
A test \(\Delta_k(X_n,Y_m)\in\{0,1\}\) uses an empirical MMD statistic to test \(H_0\colon P=Q\) against \(H_1\colon P\neq Q\), with \(\Delta_k=1\) meaning rejection. It has \emph{level} \(\alpha\) if \(\sup_{P}\,\mathbb{P}_{P\times P}(\Delta_k=1)\le \alpha\). For \(P\neq Q\), its \emph{power} is \(\mathbb{P}_{P\times Q}(\Delta_k=1)\).
\end{definition}

Motivated by Proposition~\ref{prop:main_nonzero_levy_above_drift}, we make the following empirical claim: \emph{For a fixed discount \(\gamma\) and a parametric kernel family \(k_\theta\), a smaller contraction factor on a ball \(E\) containing the data, \(\tau^\star_{k_{\theta_1},E}(\gamma)<\tau^\star_{k_{\theta_2},E}(\gamma)\), empirically comes with reduced power to distinguish return laws.}

To test this claim, we fix two distributions \(P\) and \(Q\) and measure how reliably a test of level \(\alpha\) built from the empirical MMD can detect their difference. We use a permutation test \citep{JMLR:v13:gretton12a} and estimate its power for each kernel configuration. The full protocol is described in Appendix~\ref{app:permutation_test_power}.

We consider several kernel families \(k_\theta\), with parameter \(\theta\in\Theta\). We vary \(\theta\) within \(\Theta\) and, for each resulting kernel and each discount \(\gamma\), use the procedure of Section~\ref{sec:bounded_support} to estimate \(\tau^\star_{k_\theta,E}(\gamma)\), the smallest contraction factor of \(k_\theta\) on \(E\). We also estimate the power of each kernel with the permutation test. For each discount \(\gamma\) and each bound \(\bar{\tau}\) on the contraction factor, we then select, within each family, the kernel with the highest power among those with \(\tau^\star_{k_\theta,E}(\gamma)\leq \bar{\tau}\). This shows how each kernel family trades contraction against empirical detectability.

\begin{figure}[t]
    \centering
    \includegraphics[width=\linewidth]{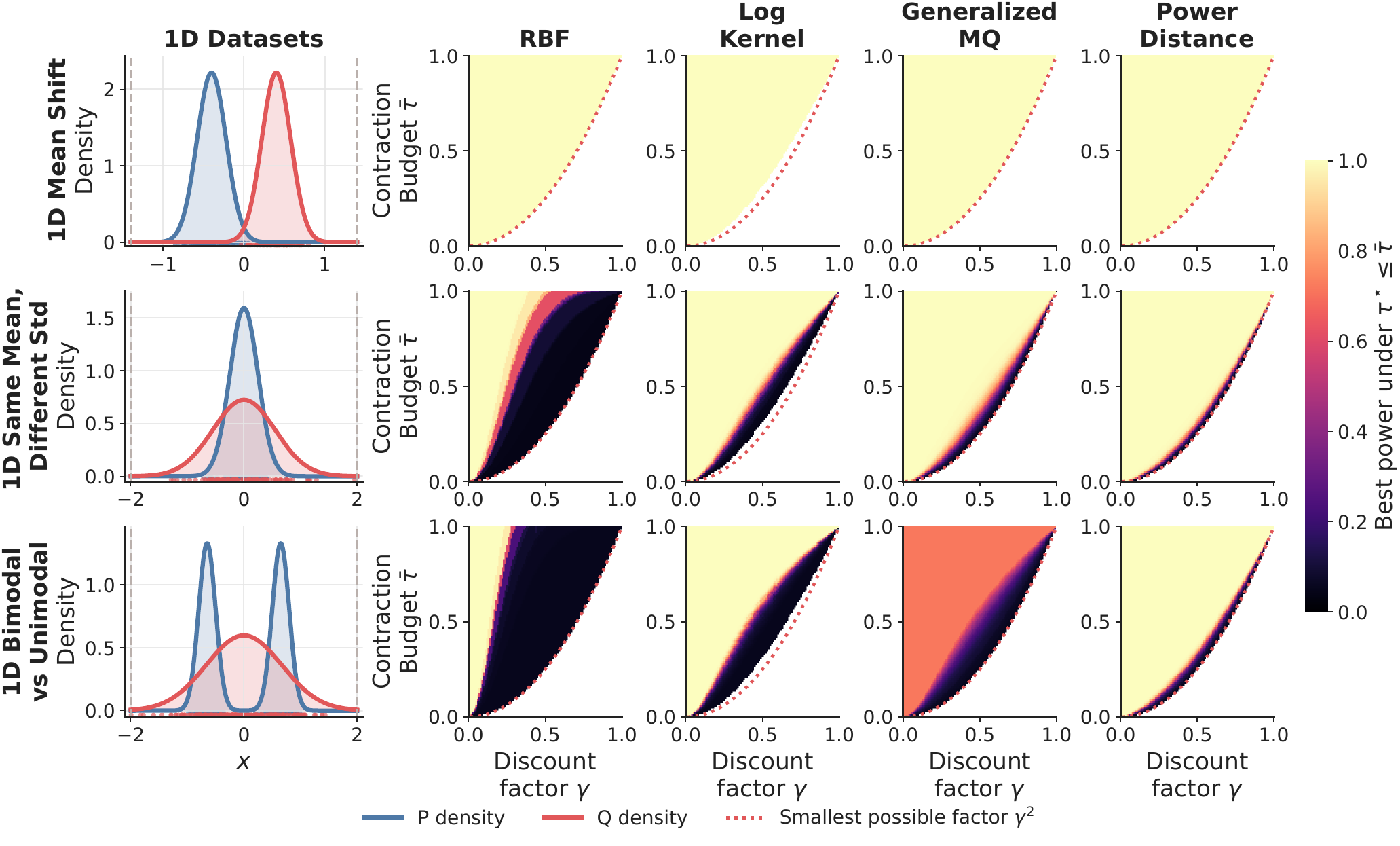}
    \caption{One-dimensional tradeoff experiments. The left column shows the clipped task distributions, with dashed lines marking the clipping radius \(R\). The remaining columns show, for each kernel family, the best empirical test power achievable under the constraint \(\tau^\star_{k_\theta,E}(\gamma)\leq \bar{\tau}\). The dotted curve is the smallest possible contraction factor \(\gamma^2\).}
    \label{fig:empirical_envelope_1d_kernel_comparison}
\end{figure}

Results are shown in Figure~\ref{fig:empirical_envelope_1d_kernel_comparison} for the RBF, logarithmic, generalized multiquadric, and power-distance kernels. The tradeoff depends both on the distributions being compared and on the kernel family. When the two distributions differ only in their mean, the tradeoff essentially disappears across all kernels. Comparing means is then enough, and each family has parameter regimes close to comparing means only. When they have the same mean but different spreads, or when one is bimodal and the other unimodal, the tradeoff becomes much stronger, especially for some kernel families, showing that a contraction factor close to \(\gamma^2\) can come at a substantial loss of power. Details about the experiments and additional two-dimensional examples, shown in Figure~\ref{fig:empirical_envelope_2d_kernel_comparison}, are provided in Appendix~\ref{app:tradeoff_detectability}. In Appendix~\ref{app:tradeoff_dimension}, we show that these observations carry over to dimensions up to \(64\), where the tradeoff stays negligible for a mean shift and grows for a checkerboard, until the dimension is too large for \(100\) samples per distribution to reveal the difference with any kernel.

\section{Conclusion and Further Directions}
\label{sec:conclusion}

This work develops a mathematical framework for studying contraction of the distributional Bellman operator under MMD. The central result is a general reduction showing that Bellman contraction is equivalent to the conditional positive definiteness of a kernel derived from the base kernel. For radial kernels, this criterion reduces further to a one-dimensional condition on the Lévy measure of the Bernstein function that generates the kernel, which can be verified analytically or tested numerically. The framework simultaneously explains and unifies existing results \citep{nguyen2021distributional,zhang2021distributional,killingberg2023the,debes2026distributional} and yields new ones. We recover contraction results for the power-distance and multiquadric kernels, identify numerically 11 additional kernel families that induce contraction, and introduce the chaining construction, which produces a kernel that induces contraction with any prescribed factor by aggregating RBF kernels.

We also study the RBF kernel, which does not induce contraction in general, yet performs well in practice \citep{nguyen2021distributional,zhang2021distributional,killingberg2023the}. When rewards are bounded, return laws remain supported on a ball, and contraction only requires the Bellman difference kernel to be conditionally positive definite on that ball. We test this condition on random finite sets of points in the ball, where it reduces to a generalized eigenvalue problem. The results suggest that the RBF kernel induces contraction on balls whose radius is small relative to the lengthscale, which may explain part of its success.

We further show that contraction rate and discriminative power are in tension. Among the radial kernels of this work, those with the smallest contraction factors are those whose MMD is least sensitive to differences beyond the mean. This tradeoff suggests that the contraction rate alone is not a sufficient criterion for kernel selection in distributional RL, and that practitioners should also consider discriminative power.

Finally, this work leaves several further directions open. We focused on contraction in distributional policy evaluation. The natural next step is to run experiments on RL environments and compare kernels on control tasks. The tradeoff between contraction rate and discriminative power depends on the distributions being compared. In distributional RL, these are the predicted and target return laws, which depend on the environment and on the policy. Therefore, measuring the tradeoff on the return laws that arise during training could guide kernel selection.

\begin{ack}
This project has received funding from the European Research Council (ERC) under the European Union's Horizon 2020 research and innovation programme (grant agreement n\textdegree~101021347).
\end{ack}

{
\small
\bibliographystyle{plainnat}
\bibliography{references}
}

\newpage
\appendix
\section*{Appendix}
\addcontentsline{toc}{section}{Appendix}

\startcontents[appendix]
\printcontents[appendix]{l}{2}{\setcounter{tocdepth}{2}}

\newpage
\section{Background on Maximum Mean Discrepancy}\label{app:background}
\subsection{Reproducing kernel Hilbert spaces} \label{app:rkhs}
\begin{definition}
A \emph{Hilbert space} \(\mathcal H\) is a vector space of functions equipped with an inner product \(\langle u, v\rangle_{\mathcal H}\). The inner product measures angles between functions, and the norm it induces, \(\|v\|_{\mathcal H}:=\sqrt{\langle v,v\rangle_{\mathcal H}}\), measures their lengths.
\end{definition}

\noindent For the formal definition, which also requires completeness, see \cite[Ch.~1]{conway1990course} or \cite[Ch.~3]{kreyszig1989introductory}.

\begin{definition}[Kernel]
Let \(X\) be a nonempty set. A \emph{kernel} on \(X\) is any function \(k: X\times X \to \R\).
We say \(k\) is \emph{symmetric} if \(k(x,y)=k(y,x)\) for all \(x,y\in X\).
\end{definition}

\begin{definition}[Positive Semidefinite Kernel {\cite[Ch.~1]{berlinet2011reproducing}}]\label{def:positive_semidefinite}
Let \(X\) be a nonempty set and \(k\colon X\times X\to\R\) be symmetric.  We say \(k\) is \emph{positive semidefinite (PSD)} if for every integer \(n\ge1\), any points \(x_1,\dots,x_n\in X\) and real coefficients \(c_1,\dots,c_n\),
\[
\sum_{i,j=1}^n c_i\,c_j\,k(x_i,x_j)
\;\ge\;0.
\]
Equivalently,
\[
\iint_{X\times X} k(x,y)\,d\sigma(x)\,d\sigma(y)\ge 0
\]
for every finite signed measure \(\sigma\) on \(X\) for which this integral is well defined.
\end{definition}

\begin{definition}[Reproducing Kernel and RKHS {\cite[p.~343]{aronszajn1950theory}}]
A function \(k\colon X\times X\to\R\) is called a \emph{reproducing kernel} if there exists a Hilbert space \(\mathcal{H}\) of functions \(f\colon X \to \R\) such that \(k(x,\cdot)\in\mathcal{H}\) for each \(x\in X\), and such that the \emph{reproducing property} holds,
\[
f(x)
= \bigl\langle f,\,k(x,\cdot)\bigr\rangle_{\mathcal{H}}
\qquad\text{for all }f\in\mathcal{H}\text{ and }x\in X.
\]
The Hilbert space \(\mathcal{H}\) is uniquely determined by \(k\) and is called the \emph{reproducing kernel Hilbert space} (RKHS) of \(k\).
\end{definition}

\begin{proposition}[Reproducing kernels are the symmetric PSD kernels {\cite[p.~344]{aronszajn1950theory}}]\label{prop:reproducing-psd}
A function \(k\colon X\times X\to\R\) is a reproducing kernel if and only if it is symmetric and positive semidefinite.
\end{proposition}

\paragraph{Feature map and mean embeddings.}
Let \(\mathcal{H}\) be the RKHS of \(k\). Then there is a canonical \emph{feature map}
\[
\Phi:X\to\mathcal H,\qquad \Phi(x)=k(x,\cdot),
\qquad
k(x,x')=\langle \Phi(x),\Phi(x')\rangle_{\mathcal H}\quad\forall x,x'\in X.
\]
For any \(P\in\mathcal P(X)\), the set of probability measures on \(X\), the \emph{mean embedding} of \(P\) is
\[
\mu_P:=\int_X \Phi(x)\,dP(x)\in\mathcal H.
\]
Since \(\mu_P\in\mathcal H\), we may apply the reproducing property. For any \(x'\in X\),
\[
\begin{aligned}
\mu_P(x')
&= \langle \mu_P,\,k(x',\cdot)\rangle_{\mathcal H}
\quad\text{(reproducing property)}\\
&= \Bigl\langle \int_X \Phi(x)\,dP(x),\;k(x',\cdot)\Bigr\rangle_{\mathcal H}
\quad\text{(definition of \(\mu_P\))}\\
&= \int_X \langle \Phi(x),\,k(x',\cdot)\rangle_{\mathcal H}\,dP(x)
\quad\text{(linearity of inner product)}\\
&= \int_X k(x',x)\,dP(x)
\quad\text{(reproducing property)}\\
&= \E_{x\sim P}\bigl[k(x',x)\bigr]
\quad\text{(definition of expectation).}
\end{aligned}
\]
Hence \(\mu_P(x')\) can be evaluated without computing the feature map, using kernel values alone, by averaging \(k(x',x)\) over points \(x\) drawn from \(P\).

\paragraph{Expectation as an inner product.}
The RKHS turns integration under \(P\) into a single inner product against the mean embedding \(\mu_P\),
\[
\begin{aligned}
\E_{x\sim P}[f(x)]
&= \int_X f(x)\,dP(x) \quad\text{(definition of expectation)}\\
&= \int_X \langle f,\,k(x,\cdot)\rangle_{\mathcal H}\,dP(x) \quad\text{(reproducing property)}\\
&= \Bigl\langle f,\;\int_X k(x,\cdot)\,dP(x)\Bigr\rangle_{\mathcal H}
   \quad\text{(linearity of inner product)}\\
&= \langle f,\;\mu_P\rangle_{\mathcal H}
   \quad\text{(definition of \(\mu_P\)).}
\end{aligned}
\]

\paragraph{Distance between feature maps.}
 Since 
    \(\displaystyle k(x,y)=\bigl\langle\Phi(x),\Phi(y)\bigr\rangle_{\mathcal H}\)
    is an inner product, the squared distance in \(\mathcal H\) between feature maps is
    \[
    \begin{aligned}
    \bigl\|\Phi(x)-\Phi(y)\bigr\|_{\mathcal H}^2
    &=\bigl\langle \Phi(x)-\Phi(y),\,\Phi(x)-\Phi(y)\bigr\rangle_{\mathcal H} \\[6pt]
    &=\bigl\langle \Phi(x),\Phi(x)\bigr\rangle_{\mathcal H}
      -2\,\bigl\langle \Phi(x),\Phi(y)\bigr\rangle_{\mathcal H}
      +\bigl\langle \Phi(y),\Phi(y)\bigr\rangle_{\mathcal H} \\[6pt]
    &=k(x,x)\;-\;2\,k(x,y)\;+\;k(y,y).
    \end{aligned}
    \]

\subsection{Maximum mean discrepancy}\label{app:mmd}

\begin{definition}[CPD and CSPD {\cite[Def.~3.1.1]{berg1984harmonic}, \cite[Def.~3.2]{killingberg2023the}}]\label{def:cpd-cspd}
Let $X$ be a measurable space and let $k:X\times X\to\R$ be symmetric.
We say that $k$ is \emph{conditionally positive definite} (CPD) if
\[
\iint_{X\times X} k(x,y)\,d\sigma(x)\,d\sigma(y)\;\ge\;0
\]
for every finite signed measure $\sigma$ on $X$ with $\sigma(X)=0$ for which this integral is well defined.
If the inequality is strict for every nonzero such $\sigma$, then $k$ is
\emph{conditionally strictly positive definite} (CSPD).
For continuous kernels, CPD is equivalent to the condition with finite sums of Definition~\ref{def:main-cpd} \cite[Lemma~3.4]{guella2021generalizationenergydistancebernstein}.
\end{definition}
\begin{definition}[Maximum mean discrepancy]\label{def:mmd}
Let $X$ be a measurable space and let $k\colon X\times X\to\R$ be symmetric.
\begin{itemize}
\item If $k$ is a reproducing (positive semidefinite) kernel with RKHS $\mathcal H_k$,
then for $P,Q\in\mathcal P(X)$ the squared maximum mean discrepancy \cite[Def.~2 and Lemma~4]{JMLR:v13:gretton12a} is defined by
\[
\MMD_k^2(P,Q):=\bigl\|\mu_P-\mu_Q\bigr\|_{\mathcal H_k}^2,
\qquad
\mu_P:=\int_X k(x,\cdot)\,dP(x).
\]
\item More generally, if $k$ is CPD (Definition~\ref{def:cpd-cspd}), we \emph{define}
the squared maximum mean discrepancy by the quadratic form \cite[Eq.~(2.6) and Thm.~22]{sejdinovic2013equivalence}
\[
\MMD_k^2(P,Q)
:=\iint_{X\times X} k(x,y)\,d(P-Q)(x)\,d(P-Q)(y),
\]
whenever the integral is well defined. Equivalently, writing $x,x'\sim P$ and
$y,y'\sim Q$, all independent, one has
\[
\MMD_k^2(P,Q)
=
\E\,k(x,x')+\E\,k(y,y')-2\,\E\,k(x,y),
\]
whenever the expectations are finite.
\end{itemize}
When \(k\) is a reproducing kernel, the expressions as a quadratic form and as expectations coincide with
the RKHS definition \cite[Lemma~6]{JMLR:v13:gretton12a}.
\end{definition}

\begin{proposition}[Centering CPD kernels {\cite{sejdinovic2013equivalence}}]\label{prop:cpd-centering}
Let $X$ be a measurable space and let $k:X\times X\to\R$ be symmetric and CPD.
Fix $x_0\in X$ and define the kernel centered at $x_0$,
\[
k^\circ(x,y):=k(x,y)-k(x,x_0)-k(x_0,y)+k(x_0,x_0).
\]
Then $k^\circ$ is positive semidefinite and hence admits an RKHS.
Moreover, for any $P,Q\in\mathcal P(X)$ such that the expressions below are well defined,
\[
\boxed{
\MMD_k^2(P,Q)
=
\iint_{X\times X} k^\circ(x,y)\,d(P-Q)(x)\,d(P-Q)(y)
=
\MMD_{k^\circ}^2(P,Q).
}
\]
\end{proposition}

\begin{definition}[Characteristic kernel]\label{def:characteristic-kernel}
A reproducing kernel $k$ is \emph{characteristic} if
\[
\MMD_k(P,Q)=0\quad\Longrightarrow\quad P=Q
\qquad\text{for all }P,Q\in\mathcal P(X).
\]
\end{definition}
\begin{proposition}[MMD is a metric under CSPD]\label{prop:mmd-metric}
Let $X$ be a measurable space and let $k:X\times X\to\R$ be symmetric and CPD.
Assume $\MMD_k^2(P,Q)$ is defined by Definition~\ref{def:mmd} for the pairs considered, and set $\MMD_k(P,Q):=\sqrt{\MMD_k^2(P,Q)}$.
Then $\MMD_k$ satisfies the metric axioms, except possibly identity of indiscernibles.
More precisely, for any $P,Q,R\in\mathcal P(X)$ such that all the quantities below are finite, the following hold.
\begin{enumerate}
\item \textbf{Nonnegativity:} $\MMD_k(P,Q)\ge 0$.
\item \textbf{Symmetry:} $\MMD_k(P,Q)=\MMD_k(Q,P)$.
\item \textbf{Triangle inequality:} $\MMD_k(P,Q)\le \MMD_k(P,R)+\MMD_k(R,Q)$.
\item \textbf{Identity of indiscernibles:}
\[
\MMD_k(P,Q)=0 \Rightarrow P=Q
\quad\text{holds if and only if}\quad
k\ \text{is CSPD}.
\]
\end{enumerate}

\end{proposition}

\begin{proof}
When $k$ is a reproducing kernel, $\MMD_k$ is a distance in the RKHS norm \cite[Lemma~4]{JMLR:v13:gretton12a}, hence it satisfies
nonnegativity, symmetry, and the triangle inequality.
For a CPD kernel $k$, Proposition~\ref{prop:cpd-centering} yields a centered positive semidefinite kernel $k^\circ$
such that $\MMD_k^2=\MMD_{k^\circ}^2$, so $\MMD_k=\MMD_{k^\circ}$ is again an RKHS norm distance.
Moreover, identity of indiscernibles holds exactly when $k$ is CSPD \cite{sejdinovic2013equivalence}.
\end{proof}

\subsection{Gram matrices and empirical MMD}\label{app:mmd_gram_matrices}

\paragraph{General formulation for discrete signed measures.}
Fix points $z_1,\dots,z_N\in X$ and coefficients $w\in\R^N$, and define the finite signed measure
\[
\sigma := \sum_{i=1}^N w_i\,\delta_{z_i}.
\]
Let $K\in\R^{N\times N}$ be the Gram matrix $K_{ij}:=k(z_i,z_j)$.
Then
\[
\iint_{X\times X} k(x,y)\,d\sigma(x)\,d\sigma(y)
=
\sum_{i=1}^N\sum_{j=1}^N w_i w_j\,k(z_i,z_j)
=
w^\top K w.
\]
In particular, if $\sum_{i=1}^N w_i=0$ and $k$ is CPD in the sense of Definition~\ref{def:cpd-cspd},
then $w^\top K w\ge 0$.

\paragraph{Two-sample empirical estimator.}
Given samples $x_{1:n}=(x_1,\dots,x_n)$ from $P$ and $y_{1:m}=(y_1,\dots,y_m)$ from $Q$, define the Gram matrices
\begin{alignat*}{2}
K_{xx} &\in \R^{n\times n}, \qquad & (K_{xx})_{ij} &:= k(x_i,x_j), \\
K_{yy} &\in \R^{m\times m}, \qquad & (K_{yy})_{ij} &:= k(y_i,y_j), \\
K_{xy} &\in \R^{n\times m}, \qquad & (K_{xy})_{ij} &:= k(x_i,y_j).
\end{alignat*}
Let $\mathbf 1_n\in\R^n$ denote the vector of ones, and similarly $\mathbf 1_m\in\R^m$.

The standard biased empirical estimator of $\MMD_k^2(P,Q)$ is
\begin{align*}
\widehat{\MMD}_{k,\mathrm{b}}^2(x_{1:n},y_{1:m})
:=
\;&
\frac{1}{n^2}\sum_{i,j=1}^n k(x_i,x_j)
+
\frac{1}{m^2}\sum_{i,j=1}^m k(y_i,y_j)
-
\frac{2}{nm}\sum_{i=1}^n\sum_{j=1}^m k(x_i,y_j),
\end{align*}
which can be written with Gram matrices as
\begin{align*}
\widehat{\MMD}_{k,\mathrm{b}}^2(x_{1:n},y_{1:m})
=
\;&
\frac{1}{n^2}\,\mathbf 1_n^\top K_{xx}\,\mathbf 1_n
+
\frac{1}{m^2}\,\mathbf 1_m^\top K_{yy}\,\mathbf 1_m
-
\frac{2}{nm}\,\mathbf 1_n^\top K_{xy}\,\mathbf 1_m.
\end{align*}

The estimator can also be written with a single Gram matrix, built from all $n+m$ points.
Let $N:=n+m$ and define the aggregate sample $z_{1:N}=(z_1,\dots,z_N)$ by
\[
z_i:=x_i\quad(1\le i\le n),
\qquad
z_{n+j}:=y_j\quad(1\le j\le m).
\]
Define the two-sample weight vector $\ell\in\R^N$ by
\[
\ell
:=
\begin{pmatrix}
\frac{1}{n}\mathbf 1_n\\[2pt]
-\frac{1}{m}\mathbf 1_m
\end{pmatrix},
\qquad\text{so that }\ \mathbf 1_N^\top \ell = 0,
\]
and let $K\in\R^{N\times N}$ be the Gram matrix of the aggregate sample, $K_{ij}:=k(z_i,z_j)$, which has the blocks
\[
K=\begin{pmatrix} K_{xx} & K_{xy}\\ K_{xy}^\top & K_{yy}\end{pmatrix}.
\]
Multiplying out $\ell^\top K\ell$ block by block gives the expression with Gram matrices above, so
\[
\widehat{\MMD}_{k,\mathrm{b}}^2(x_{1:n},y_{1:m})
=
\ell^\top K\ell.
\]

\paragraph{Discrete measures on a common support.}
Let $S=\{z_1,\dots,z_N\}\subset X$ and consider two probability measures supported on $S$,
\[
\mu=\sum_{i=1}^N a_i\,\delta_{z_i},
\qquad
\nu=\sum_{i=1}^N b_i\,\delta_{z_i},
\qquad
\sum_{i=1}^N a_i=\sum_{i=1}^N b_i=1.
\]
Define $w\in\R^N$ by $w_i:=a_i-b_i$, so that $\mathbf 1_N^\top w=0$, and set
$\sigma:=\mu-\nu=\sum_{i=1}^N w_i\delta_{z_i}$.
Then the same identity for discrete measures yields
\[
\MMD_k^2(\mu,\nu)
=
\iint_{X\times X} k(x,y)\,d(\mu-\nu)(x)\,d(\mu-\nu)(y)
=
w^\top K w,
\qquad
K_{ij}:=k(z_i,z_j).
\]
The constraint $\mathbf 1_N^\top w=0$ is the discrete counterpart of the condition
$(\mu-\nu)(X)=0$.

\paragraph{Centering and restriction to vectors whose entries sum to zero.}
In several arguments in this paper, we must control quadratic forms $w^\top K w$ for all
vectors $w$ whose entries sum to zero.
It is convenient to encode the constraint $\mathbf 1_N^\top w=0$ using the centering projector
\[
H_N:=I_N-\frac{1}{N}\mathbf 1_N\mathbf 1_N^\top,
\]
which satisfies $H_N w=w$ whenever $\mathbf 1_N^\top w=0$. Consequently, for any such $w$,
\[
w^\top K w
=
(H_N w)^\top K (H_N w)
=
w^\top (H_N K H_N) w.
\]
Moreover, the condition
\[
w^\top K w \ge 0
\qquad\text{for all }w\in\R^N\text{ with }\mathbf 1_N^\top w=0
\]
is equivalent to the matrix inequality
\[
H_N K H_N \succeq 0.
\]
Indeed, assume first that $H_N K H_N\succeq 0$ and let $w\in\R^N$ satisfy $\mathbf 1_N^\top w=0$.
Since $H_N w=w$, we have
\[
w^\top K w=(H_N w)^\top K (H_N w)=w^\top(H_N K H_N)w\ge 0.
\]
Conversely, assume that $w^\top K w\ge 0$ for every $w\in\R^N$ with $\mathbf 1_N^\top w=0$.
Fix $v\in\R^N$ and set $w:=H_N v$, so that $\mathbf 1_N^\top w=0$. Then
\[
v^\top(H_N K H_N)v=(H_N v)^\top K (H_N v)=w^\top K w\ge 0.
\]
Since this holds for all $v\in\R^N$, it follows that $H_N K H_N\succeq 0$.

\newpage
\section{Basic identities for $\MMD^2$}

\begin{lemma}[Equivalence of nonnegativity of \MMD$^2$ on a set and CPD on that set]\label{lem:MMD_CPD}
Let \(k\colon X\times X\to\R\) be a symmetric kernel on a measurable space \(X\), and let \(E\subseteq X\) be measurable. Then
\[
\boxed{
\begin{array}{c}
\MMD_k^2(P,Q)\ge 0
\quad
\forall\,P,Q\in\mathcal P(X)
\text{ with }\mathrm{supp}\,P,\mathrm{supp}\,Q\subseteq E
\\[4pt]
\text{if and only if}
\\[4pt]
k\text{ is CPD on }E
\end{array}
}
\]
Here, “CPD on \(E\)” means that
\[
\iint_{E\times E} k(x,y)\,d\mu(x)\,d\mu(y)\ge 0
\]
for every finite signed measure \(\mu\) supported on \(E\) with \(\mu(E)=0\) for which this integral is well defined.
\end{lemma}

\begin{proof}
\noindent\paragraph{CPD on \(E\) implies nonnegativity.}

Take any two probability measures \(P,Q\) supported on \(E\), and set
\[
\sigma:=P-Q.
\]
Then \(\sigma\) is supported on \(E\) and satisfies \(\sigma(E)=1-1=0\). Hence, by the assumption that \(k\) is CPD on \(E\),
\[
\MMD_k^2(P,Q)
=
\iint k(x,y)\,d\sigma(x)\,d\sigma(y)
\ge 0.
\]

\medskip
\noindent\paragraph{Nonnegativity implies CPD on \(E\).}
Assume that \(\MMD_k^2(P,Q)\ge 0\) for all probability measures \(P,Q\) supported on \(E\) for which \(\MMD_k^2(P,Q)\) is well defined. Let \(\mu\) be any finite signed measure supported on \(E\) with \(\mu(E)=0\) for which \(\iint k\,d\mu\,d\mu\) is well defined. By the Hahn--Jordan decomposition,
\[
\mu=\mu^+-\mu^-,
\qquad
\mu^+(E)=\mu^-(E)=:A,
\]
where the equality follows from \(\mu(E)=0\). If \(A=0\), then \(\mu^+(E)=\mu^-(E)=0\), hence \(\mu^+=\mu^-=0\), so that \(\mu=0\). In that case,
\[
\iint k(x,y)\,d\mu(x)\,d\mu(y)=0,
\]
and the conclusion is immediate. Assume therefore that \(A>0\). Define
\[
P=\frac{1}{A}\mu^+,
\qquad
Q=\frac{1}{A}\mu^-.
\]
These are probability measures supported on \(E\), and they satisfy
\[
P-Q=\frac{1}{A}\mu.
\]
By bilinearity of the quadratic form and the identity \(\mu=A(P-Q)\),
\begin{align*}
\iint k(x,y)\,d\mu(x)\,d\mu(y)
&=
A^2\iint k(x,y)\,d(P-Q)(x)\,d(P-Q)(y) \\
&=
A^2\,\MMD_k^2(P,Q) \\
&\ge 0.
\end{align*}
Hence \(k\) is CPD on \(E\).
\end{proof}

\begin{lemma}[\(\MMD^2\) of pushforward measures]\label{lem:pushforward}
Let \(f\colon X\to Y\) be a measurable map, let \(\nu,\mu\) be probability measures on \(X\), let \(k\colon Y\times Y\to\R\) be any kernel, and define \(k_f(x,x'):=k\bigl(f(x),f(x')\bigr)\). Then
\[
\boxed{\;
\MMD^2_{k}\bigl(f\#\nu,\;f\#\mu\bigr)
\;=\;
\MMD^2_{k_f}\bigl(\nu,\;\mu\bigr).\;}
\]
\end{lemma}

\begin{proof}
Define the pushforward measures
\[
(f\#\nu)(B)\;=\;\nu\bigl(f^{-1}(B)\bigr),
\quad
(f\#\mu)(B)\;=\;\mu\bigl(f^{-1}(B)\bigr).
\]
By the definition of \MMD\(^2\),
\[
\begin{aligned}
\MMD^2_{k}\bigl(f\#\nu,f\#\mu\bigr)
&=
\Eop_{u,u'\sim f\#\nu}\bigl[k(u,u')\bigr]
+ \Eop_{v,v'\sim f\#\mu}\bigl[k(v,v')\bigr] \\[-2pt]
&\quad
-2\,\Eop_{u\sim f\#\nu,\;v\sim f\#\mu}\bigl[k(u,v)\bigr]
\\[6pt]
&=
\Eop_{x,x'\sim\nu}\bigl[k\bigl(f(x),f(x')\bigr)\bigr]
+ \Eop_{y,y'\sim\mu}\bigl[k\bigl(f(y),f(y')\bigr)\bigr] \\[-2pt]
&\quad
-2\,\Eop_{x\sim\nu,\;y\sim\mu}\bigl[k\bigl(f(x),f(y)\bigr)\bigr]
\\[6pt]
&=
\Eop_{x,x'\sim\nu}\bigl[k_f(x,x')\bigr]
+ \Eop_{y,y'\sim\mu}\bigl[k_f(y,y')\bigr]
-2\,\Eop_{x\sim\nu,\;y\sim\mu}\bigl[k_f(x,y)\bigr]
\\[4pt]
&=
\MMD^2_{k_f}\bigl(\nu,\mu\bigr)\,.
\end{aligned}
\]
\end{proof}

\begin{lemma}[\(\MMD^2\) is linear in the kernel]\label{lem:MMD_linearity}
Let $\nu,\mu$ be probability measures on a measurable space $X$, let
$k_1,k_2:X \times X\to \R$ be kernels, and let $\alpha_1,\alpha_2\in \R$. Then
\[
\boxed{\;
\MMD^2_{\alpha_1\,k_1+\alpha_2\,k_2}\bigl(\nu,\mu\bigr)
=\alpha_1\,\MMD^2_{k_1}\bigl(\nu,\mu\bigr)
+\alpha_2\,\MMD^2_{k_2}\bigl(\nu,\mu\bigr).\;}
\]
In particular, setting $k_2\equiv0$ gives
\[
\MMD^2_{\alpha_1\,k_1}\bigl(\nu,\mu\bigr)
=\alpha_1\,\MMD^2_{k_1}\bigl(\nu,\mu\bigr).
\]
\end{lemma}
\begin{proof}
This follows from
\[
\MMD^2_{k}\bigl(\nu,\mu\bigr)
=\iint_{X\times X}k(x,y)\,\bigl(\nu-\mu\bigr)(dx)\,\bigl(\nu-\mu\bigr)(dy)
\]
by linearity of the integral.
\end{proof}

\newpage
\section{Bellman contraction in Maximum Mean Discrepancy}
\label{app:general_contraction}
\paragraph{Preamble.}
This section provides the contraction theory underlying all subsequent arguments. By Theorem~\ref{thm:mmd-contraction-cpd}, for a fixed map \(f\), an MMD inequality of the form
\[
\MMD^2_{k}\bigl(f\#\nu,f\#\mu\bigr)\le \tau\,\MMD^2_{k}(\nu,\mu)
\]
holds for all \(\nu,\mu\) if and only if the \emph{difference kernel} \(\tau k-k_f\) is CPD, where \(k_f(x,y)=k(f(x),f(y))\). This criterion is then extended to the full distributional Bellman operator by representing \(\mathcal T^\pi\) as a mixture of affine pushforwards and exploiting convexity of the CPD quadratic form under averaging, yielding a condition that is both sufficient and necessary for \(\tau\)-contraction in \(\overline{\MMD}_k^2\) across all MDPs with discount \(\gamma\) (Theorem~\ref{thm:bellman_contraction_cpd}). Section~\ref{app:contraction_bounded_support} then extends the analysis to return laws supported on a bounded set, where the CPD condition is required only on that set. It is therefore weaker, and holds for some kernels that fail it on all of \(\R^d\). For radial kernels on \(\R^d\) with \(d\) fixed, as in Proposition~\ref{prop:radial_contraction}, the same criterion can be equivalently expressed as a conditionally negative definite (CND) property for an associated radial kernel. This connects Bellman contraction in MMD to CND kernels and to the characterizations of Schoenberg and Bernstein used in the next section.
\subsection{Contraction via difference kernel}

\begin{theorem}[Contraction of MMD on a set under CPD kernels]
\label{thm:mmd-contraction-cpd}
Let \(f\colon X\to X\) be measurable, let \(k\colon X\times X\to\R\) be a symmetric, conditionally positive definite (CPD) kernel, fix a measurable set \(E\subseteq X\), and define
\[
k_f(x,y)=k\bigl(f(x),f(y)\bigr),
\qquad x,y\in X.
\]
Fix \(\tau\in\R\). Then
\[
\boxed{%
\begin{gathered}
\MMD^2_{k}\bigl(f\#\nu,f\#\mu\bigr)
\;\le\;
\tau\,\MMD^2_{k}(\nu,\mu)
\qquad
\forall\,\mu,\nu\in\mathcal P(X)\ \text{with }\mathrm{supp}\,\mu,\mathrm{supp}\,\nu\subseteq E
\\[4pt]
\text{if and only if}
\\[4pt]
\tau\,k \;-\; k_f
\text{ is CPD on }E
\end{gathered}
}
\]
Here, “CPD on \(E\)” means that
\[
\iint_{E\times E} \bigl(\tau k-k_f\bigr)(x,y)\,d\sigma(x)\,d\sigma(y)\ge 0
\]
for every finite signed measure \(\sigma\) supported on \(E\) with \(\sigma(E)=0\) for which this integral is well defined.

Accordingly, for laws supported on \(E\), the smallest such \(\tau\) is
\[
\tau_E^\star:=\inf\{\tau>0:\ \tau k-k_f\text{ is CPD on }E\}.
\]
Contraction on laws supported on \(E\) holds exactly when \(\tau_E^\star<1\). In that case, \(\tau_E^\star\) is the smallest contraction factor on \(E\) in \(\MMD^2\), and the corresponding smallest contraction factor in \(\MMD\) is \(\sqrt{\tau_E^\star}\). The choice \(E=X\) yields contraction on all probability laws on \(X\), while proper subsets \(E\) restrict the comparison to laws supported on a prescribed region, such as the bounded sets used later.
\end{theorem}

\begin{proof}
Fix \(\mu,\nu\in\mathcal P(X)\) with
\[
\mathrm{supp}\,\mu,\mathrm{supp}\,\nu\subseteq E.
\]
By Lemma~\ref{lem:pushforward}, we have
\[
\MMD^2_{k}\bigl(f\#\nu,f\#\mu\bigr)
=\MMD^2_{k_f}(\nu,\mu).
\]
By Lemma~\ref{lem:MMD_linearity},
\[
\tau\,\MMD^2_{k}(\nu,\mu)=\MMD^2_{\tau k}(\nu,\mu).
\]
Hence
\[
\tau\,\MMD^2_{k}(\nu,\mu)
-\MMD^2_{k}\bigl(f\#\nu,f\#\mu\bigr)
=
\MMD^2_{\tau k}(\nu,\mu)
-\MMD^2_{k_f}(\nu,\mu).
\]
By Lemma~\ref{lem:MMD_linearity}, this difference is
\[
\MMD^2_{\tau k-k_f}(\nu,\mu).
\]
Therefore,
\[
\MMD^2_{k}\bigl(f\#\nu,f\#\mu\bigr)\le
\tau\,\MMD^2_{k}(\nu,\mu)
\quad\Longleftrightarrow\quad
\MMD^2_{\tau k-k_f}(\nu,\mu)\ge 0.
\]
Thus
\[
\MMD^2_{k}\bigl(f\#\nu,f\#\mu\bigr)\le
\tau\,\MMD^2_{k}(\nu,\mu)
\qquad
\forall\,\mu,\nu\in\mathcal P(X)\text{ with }\mathrm{supp}\,\mu,\mathrm{supp}\,\nu\subseteq E
\]
if and only if
\[
\MMD^2_{\tau k-k_f}(\nu,\mu)\ge 0
\qquad
\forall\,\mu,\nu\in\mathcal P(X)\text{ with }\mathrm{supp}\,\mu,\mathrm{supp}\,\nu\subseteq E.
\]
Finally, by Lemma~\ref{lem:MMD_CPD}, this is equivalent to \(\tau k-k_f\) being CPD on \(E\).
\end{proof}

\begin{lemma}[The Bellman operator averages pushforwards under affine maps]
\label{lem:bellman_affine_representation}
Fix \((s,a)\in\mathcal S\times\mathcal A\), and let
\[
C := \bigl(R(s,a),S',A'\bigr)
\]
with law \(\rho_{s,a}\). For each realization \(C=(q,s',a')\), define
\[
f_q:\R^d\to\R^d,
\qquad
f_q(x):=q+\gamma x.
\]
Let \(\eta:\mathcal S\times\mathcal A\to\mathcal P(\R^d)\), and let \(X'\mid C\sim \eta(S',A')\). Given \(C\) (the reward, next state and next action), the Bellman update maps the next return law to its pushforward under the affine map \(f_q\),
\[
\Law(q+\gamma X' \mid C)=f_q\#\eta(s',a').
\]
Consequently,
\[\boxed{
(\mathcal T^\pi\eta)(s,a)
=
\int f_q\#\eta(s',a')\,\rho_{s,a}(dC).
}\]
\end{lemma}

\begin{proof}
Fix a realization \(C=(q,s',a')\). Conditionally on \(C\), the random variable \(X'\) has law \(\eta(s',a')\). Hence, for every Borel set \(B\subseteq\R^d\),
\[
\Pr(q+\gamma X'\in B\mid C)
=
\Pr(X'\in f_q^{-1}(B)\mid C)
=
\eta(s',a')\bigl(f_q^{-1}(B)\bigr)
=
(f_q\#\eta(s',a'))(B).
\]
Therefore,
\[
\Law(q+\gamma X' \mid C)=f_q\#\eta(s',a').
\]
Integrating this conditional law with respect to \(\rho_{s,a}\) yields
\[
(\mathcal T^\pi\eta)(s,a)
=
\int f_q\#\eta(s',a')\,\rho_{s,a}(dC),
\]
which is the claimed representation.
\end{proof}

\begin{lemma}[A Bellman operator can reduce to an affine map]
\label{lem:deterministic_bellman_embedding}
Fix \(q\in\R^d\) and \(\gamma\in(0,1)\). Then there exists a discounted MDP with a single state \(s\), a single action \(a\) whose transition always returns to \(s\), and deterministic reward \(q\), together with its unique fixed policy \(\pi\), such that
\[\boxed{
(\mathcal T^\pi\eta)(s,a)=f_q\#\eta(s,a),
\qquad
f_q(x)=q+\gamma x,
}\]
for every law \(\eta(s,a)\in\mathcal P(\R^d)\). Moreover, since \(\mathcal S\times\mathcal A=\{(s,a)\}\),
\[
\overline{\MMD}_k^2(\eta_1,\eta_2)
=
\MMD_k^2\bigl(\eta_1(s,a),\eta_2(s,a)\bigr)
\]
for all laws \(\eta_1(s,a),\eta_2(s,a)\in\mathcal P(\R^d)\).
\end{lemma}

\begin{proof}
Consider the MDP with one state \(s\), one action \(a\), transition kernel
\[
P(\{s\}\mid s,a)=1,
\]
and deterministic reward equal to \(q\). Let \(\pi\) be the unique policy on this MDP. For any law \(\eta(s,a)\in\mathcal P(\R^d)\), let \(X'\sim \eta(s,a)\). By definition of the distributional Bellman operator,
\[
(\mathcal T^\pi\eta)(s,a)
=
\Law(q+\gamma X')
=
f_q\#\eta(s,a).
\]
The identity for \(\overline{\MMD}_k^2\) is immediate because there is only one state-action pair.
\end{proof}

\begin{lemma}[Averaging does not increase the CPD quadratic form]
\label{lem:cpd_mixture_quadratic_form}
Let \(k\colon \R^d\times\R^d\to\R\) be a symmetric CPD kernel, let \((\Omega,\mathcal F,\rho)\) be a probability space, and let \((\sigma_c)_{c\in\Omega}\) be a measurable family of finite signed measures on \(\R^d\), each with total mass \(0\). Define
\[
\bar\sigma:=\int_\Omega \sigma_c\,\rho(dc).
\]
Then
\[
\boxed{
\iint k(x,y)\,d\bar\sigma(x)\,d\bar\sigma(y)
\le
\int_\Omega
\left(
\iint k(x,y)\,d\sigma_c(x)\,d\sigma_c(y)
\right)\rho(dc)
}
\]
whenever the quadratic forms are well defined.
\end{lemma}

\begin{proof}
Define
\[
Q(\sigma):=\iint k(x,y)\,d\sigma(x)\,d\sigma(y).
\]
Since \(k\) is CPD and each \(\sigma_c-\bar\sigma\) has total mass \(0\), we have
\[
Q(\sigma_c-\bar\sigma)\ge 0
\qquad\text{for every }c\in\Omega.
\]
Integrating with respect to \(\rho\) yields
\[
0\le \int_\Omega Q(\sigma_c-\bar\sigma)\,\rho(dc).
\]
Expanding the quadratic form and using bilinearity,
\begin{align*}
\int_\Omega Q(\sigma_c-\bar\sigma)\,\rho(dc)
&=
\int_\Omega Q(\sigma_c)\,\rho(dc)
-2\int_\Omega\iint k(x,y)\,d\sigma_c(x)\,d\bar\sigma(y)\,\rho(dc) \\
&\qquad
+Q(\bar\sigma).
\end{align*}
To simplify the mixed term, write
\[
\bar\sigma=\int_\Omega \sigma_{c'}\,\rho(dc'),
\]
where \(c'\) is a dummy variable. Then, using bilinearity to expand \(\bar\sigma\) and Fubini to interchange the \(c\) and \(c'\) integrals, we obtain
\begin{align*}
\int_\Omega\iint k(x,y)\,d\sigma_c(x)\,d\bar\sigma(y)\,\rho(dc)
&=
\int_\Omega\int_\Omega
\left(
\iint k(x,y)\,d\sigma_c(x)\,d\sigma_{c'}(y)
\right)\rho(dc')\,\rho(dc) \\
&=
\iint k(x,y)\,
d\!\left(\int_\Omega \sigma_c\,\rho(dc)\right)(x)\,
d\!\left(\int_\Omega \sigma_{c'}\,\rho(dc')\right)(y) \\
&=
\iint k(x,y)\,d\bar\sigma(x)\,d\bar\sigma(y) \\
&=
Q(\bar\sigma).
\end{align*}
Therefore
\[
0\le \int_\Omega Q(\sigma_c)\,\rho(dc)-Q(\bar\sigma),
\]
which is exactly the claimed inequality.
\end{proof}

\begin{lemma}[\(\MMD^2\) contraction from affine maps to the Bellman operator]
\label{lem:bellman_pointwise_lift}
Let \(k\colon \R^d\times\R^d\to\R\) be a symmetric CPD kernel as in Definition~\ref{def:cpd-cspd}, let \(E\subseteq\R^d\), and let \(\eta_1,\eta_2:\mathcal S\times\mathcal A\to\mathcal P(\R^d)\) satisfy
\[
\mathrm{supp}\,\eta_i(s,a)\subseteq E
\qquad
\forall\,(s,a)\in\mathcal S\times\mathcal A,\quad i=1,2.
\]
Assume that there exists \(\tau\ge 0\) such that, for every \(q\in\R^d\), the affine map
\[
f_q:\R^d\to\R^d,
\qquad
f_q(x)=q+\gamma x,
\]
satisfies
\[
\MMD_k^2\bigl(f_q\#\nu,f_q\#\mu\bigr)
\le
\tau\,\MMD_k^2(\nu,\mu)
\qquad
\forall\,\nu,\mu\in\mathcal P(\R^d)
\text{ with }\mathrm{supp}\,\nu,\mathrm{supp}\,\mu\subseteq E,
\]
for which the quantities are well defined. Then, for every \((s,a)\in\mathcal S\times\mathcal A\),
\[\boxed{
\MMD_k^2\bigl((\mathcal T^\pi\eta_1)(s,a),(\mathcal T^\pi\eta_2)(s,a)\bigr)
\le
\tau\,\overline{\MMD}_k^2(\eta_1,\eta_2).
}\]
\end{lemma}
\begin{proof}
Fix \((s,a)\in\mathcal S\times\mathcal A\), and let
\[
C:=\bigl(R(s,a),S',A'\bigr)
\]
denote the reward, next state and next action drawn from \((s,a)\), with law
\[
\rho_{s,a}:=\Law(C).
\]
By Lemma~\ref{lem:bellman_affine_representation}, for each realization \(C=(q,s',a')\), define
\[
\mu_C:=f_q\#\eta_1(s',a'),
\qquad
\nu_C:=f_q\#\eta_2(s',a').
\]
Then
\[
(\mathcal T^\pi\eta_1)(s,a)=\int \mu_C\,\rho_{s,a}(dC),
\qquad
(\mathcal T^\pi\eta_2)(s,a)=\int \nu_C\,\rho_{s,a}(dC).
\]

\noindent\textbf{Comparing quadratic forms for mixtures.}

Write
\[
\sigma_C:=\mu_C-\nu_C,
\qquad
\bar\sigma:=\int \sigma_C\,\rho_{s,a}(dC).
\]
Since
\[
\bar\sigma
=
(\mathcal T^\pi\eta_1)(s,a)-(\mathcal T^\pi\eta_2)(s,a),
\]
Lemma~\ref{lem:cpd_mixture_quadratic_form} gives
\[
\iint k(x,y)\,d\bar\sigma(x)\,d\bar\sigma(y)
\le
\int
\left(
\iint k(x,y)\,d\sigma_C(x)\,d\sigma_C(y)
\right)\rho_{s,a}(dC).
\]
By the definition of \(\MMD_k^2\) as a quadratic form for CPD kernels in Definition~\ref{def:mmd}, this yields
\[
\MMD_k^2\bigl((\mathcal T^\pi\eta_1)(s,a),(\mathcal T^\pi\eta_2)(s,a)\bigr)
\le
\int \MMD_k^2(\mu_C,\nu_C)\,\rho_{s,a}(dC).
\]

\noindent\textbf{Applying the assumed contraction on \(E\).}

Since \(\eta_1(s',a')\) and \(\eta_2(s',a')\) are supported on \(E\), the inequality assumed for \(f_q\) applies and gives
\[
\MMD_k^2(\mu_C,\nu_C)
\le
\tau\,\MMD_k^2\bigl(\eta_1(s',a'),\eta_2(s',a')\bigr).
\]
Therefore,
\begin{align*}
\MMD_k^2\bigl((\mathcal T^\pi\eta_1)(s,a),(\mathcal T^\pi\eta_2)(s,a)\bigr)
&\le
\int \MMD_k^2(\mu_C,\nu_C)\,\rho_{s,a}(dC) \\
&\le
\tau \int \MMD_k^2\bigl(\eta_1(s',a'),\eta_2(s',a')\bigr)\,\rho_{s,a}(dC) \\
&\le
\tau\,\overline{\MMD}_k^2(\eta_1,\eta_2),
\end{align*}
which proves the claim.
\end{proof}

\begin{theorem}[CPD criterion for Bellman contraction in \(\overline{\MMD}_k^2\)]
\label{thm:bellman_contraction_cpd}
Let \(k\colon \R^d\times\R^d\to\R\) be a symmetric CPD kernel as in Definition~\ref{def:cpd-cspd}, and fix \(\tau\in[0,1)\). For each \(q\in\R^d\), define the Bellman difference kernel
\[
k_{\mathrm{diff}}^{q,\gamma}(x,y):=\tau k(x,y)-k(q+\gamma x,q+\gamma y).
\]
If \(k_{\mathrm{diff}}^{q,\gamma}\) is CPD for every \(q\in\R^d\), then \(\mathcal T^\pi\) is a \(\tau\)-contraction in \(\overline{\MMD}_k^2\), that is,
\[\boxed{
\overline{\MMD}_k^2(\mathcal T^\pi\eta_1,\mathcal T^\pi\eta_2)
\le
\tau\,\overline{\MMD}_k^2(\eta_1,\eta_2)
\qquad
\forall\,\eta_1,\eta_2:\mathcal S\times\mathcal A\to\mathcal P(\R^d).
}\]
Furthermore, this condition is necessary for \(\tau\)-contraction to hold for the Bellman operator of every MDP with discount \(\gamma\) under every fixed policy.
\end{theorem}

\begin{proof}
\noindent\paragraph{Sufficiency.}
Let \(\eta_1,\eta_2:\mathcal S\times\mathcal A\to\mathcal P(\R^d)\).
By Theorem~\ref{thm:mmd-contraction-cpd}, applied with \(X=\R^d\), \(E=\R^d\), and
\[
f_q(x)=q+\gamma x,
\]
the assumption that \(k_{\mathrm{diff}}^{q,\gamma}\) is CPD for every \(q\in\R^d\) implies that
\[
\MMD_k^2\bigl(f_q\#\nu,f_q\#\mu\bigr)
\le
\tau\,\MMD_k^2(\nu,\mu)
\qquad
\forall\,\nu,\mu\in\mathcal P(\R^d),
\]
for every \(q\in\R^d\) and for all quantities that are well defined. Applying Lemma~\ref{lem:bellman_pointwise_lift} with \(E=\R^d\), we obtain, for every \((s,a)\in\mathcal S\times\mathcal A\),
\[
\MMD_k^2\bigl((\mathcal T^\pi\eta_1)(s,a),(\mathcal T^\pi\eta_2)(s,a)\bigr)
\le
\tau\,\overline{\MMD}_k^2(\eta_1,\eta_2).
\]
Taking the supremum over \((s,a)\) yields
\[
\overline{\MMD}_k^2(\mathcal T^\pi\eta_1,\mathcal T^\pi\eta_2)
\le
\tau\,\overline{\MMD}_k^2(\eta_1,\eta_2),
\]
so \(\mathcal T^\pi\) is a \(\tau\)-contraction in \(\overline{\MMD}_k^2\).

\medskip

\noindent\paragraph{Necessity.}
Assume now that \(\tau\)-contraction holds for the Bellman operator of every MDP with discount \(\gamma\) under every fixed policy. Fix \(q\in\R^d\). By Lemma~\ref{lem:deterministic_bellman_embedding}, there exists a discounted MDP and a fixed policy for which
\[
(\mathcal T^\pi\eta)(s,a)=f_q\#\eta(s,a)
\]
at the unique state-action pair \((s,a)\). Since in this problem
\[
\overline{\MMD}_k^2(\eta_1,\eta_2)
=
\MMD_k^2\bigl(\eta_1(s,a),\eta_2(s,a)\bigr),
\]
the assumed Bellman \(\tau\)-contraction implies
\[
\MMD_k^2\bigl(f_q\#\nu,f_q\#\mu\bigr)
\le
\tau\,\MMD_k^2(\nu,\mu)
\qquad
\forall\,\nu,\mu\in\mathcal P(\R^d).
\]
Applying Theorem~\ref{thm:mmd-contraction-cpd} with \(X=\R^d\) and \(E=\R^d\), we conclude that \(k_{\mathrm{diff}}^{q,\gamma}\) is CPD. Since \(q\) was arbitrary, this holds for every \(q\in\R^d\).
\end{proof}

\newpage
\subsection{Contraction on a bounded support}\label{app:contraction_bounded_support}

\begin{proposition}[The ball as a natural set under bounded rewards]
\label{prop:bounded_return_ball}
Assume that there exists \(r_{\max}\ge 0\) such that
\[
\|R(s,a)\|\le r_{\max}
\qquad\text{for every }(s,a)\in\mathcal S\times\mathcal A.
\]
Let
\[
E:=B_M=\{x\in\R^d:\|x\|\le M\},
\qquad
M\ge \frac{r_{\max}}{1-\gamma}.
\]
Then the following hold.
\begin{enumerate}
\item For every \((s,a)\in\mathcal S\times\mathcal A\), the return law \(\eta^\pi(s,a)\) is supported on \(E\).
\item If \(\eta:\mathcal S\times\mathcal A\to\mathcal P(\R^d)\) satisfies
\[
\mathrm{supp}\,\eta(s,a)\subseteq E
\qquad\forall\,(s,a)\in\mathcal S\times\mathcal A,
\]
then also
\[
\mathrm{supp}\,(\mathcal T^\pi\eta)(s,a)\subseteq E
\qquad\forall\,(s,a)\in\mathcal S\times\mathcal A.
\]
In other words, the class of laws supported on \(E\) is invariant under \(\mathcal T^\pi\).
\end{enumerate}
\end{proposition}

\begin{proof}
\leavevmode\par
\noindent\textbf{1. Bounded rewards imply bounded returns.}

Fix \((s,a)\in\mathcal S\times\mathcal A\), and consider the random return
\[
Z^\pi(s,a)=\sum_{t\ge 0}\gamma^t R_t.
\]
For almost every realization of this random variable, the triangle inequality and the reward bound give
\[
\|Z^\pi(s,a)\|
\le
\sum_{t\ge 0}\gamma^t \|R_t\|
\le
\sum_{t\ge 0}\gamma^t r_{\max}
=
\frac{r_{\max}}{1-\gamma}
\le M.
\]
Thus almost every realization of \(Z^\pi(s,a)\) belongs to \(E=B_M\). Therefore,
\[
\mathrm{supp}\,\eta^\pi(s,a)\subseteq E.
\]

\noindent\textbf{2. Bellman invariance of the ball.}

Assume now that
\[
\mathrm{supp}\,\eta(s,a)\subseteq E
\qquad\forall\,(s,a)\in\mathcal S\times\mathcal A.
\]
Fix \((s,a)\), and let \(X'\sim \eta(S',A')\) in the Bellman update. Since \(\eta(S',A')\) is supported on \(E\), almost every realization of \(X'\) satisfies
\[
\|X'\|\le M.
\]
Hence
\[
\|R(s,a)+\gamma X'\|
\le
\|R(s,a)\|+\gamma\|X'\|
\le
r_{\max}+\gamma M
\le
M,
\]
where the last inequality follows from \(M\ge r_{\max}/(1-\gamma)\). Thus almost every realization of \(R(s,a)+\gamma X'\) lies in \(E\), so
\[
\mathrm{supp}\,(\mathcal T^\pi\eta)(s,a)\subseteq E.
\]
Since \((s,a)\) was arbitrary, the conclusion holds for all state-action pairs.
\end{proof}

\begin{theorem}[CPD criterion for Bellman contraction on a set]
\label{thm:bellman_contraction_cpd_support}
Let \(k\colon \R^d\times\R^d\to\R\) be a symmetric CPD kernel as in Definition~\ref{def:cpd-cspd}, let \(E\subseteq \R^d\) be measurable, and fix \(\tau\in[0,1)\). For each \(q\in\R^d\), define the Bellman difference kernel
\[
k_{\mathrm{diff}}^{q,\gamma}(x,y):=\tau k(x,y)-k(q+\gamma x,q+\gamma y).
\]
Assume that \(k_{\mathrm{diff}}^{q,\gamma}\) is CPD on \(E\) for every \(q\in\R^d\). Let
\[
\mathcal C_E:=\bigl\{\eta:\mathcal S\times\mathcal A\to\mathcal P(\R^d)\;:\;
\mathrm{supp}\,\eta(s,a)\subseteq E
\ \ \forall\,(s,a)\in\mathcal S\times\mathcal A\bigr\}.
\]
Then, for all \(\eta_1,\eta_2\in\mathcal C_E\),
\[\boxed{
\overline{\MMD}_k^2(\mathcal T^\pi\eta_1,\mathcal T^\pi\eta_2)
\le
\tau\,\overline{\MMD}_k^2(\eta_1,\eta_2).
}\]
If, in addition, \(\mathcal T^\pi(\mathcal C_E)\subseteq \mathcal C_E\), then \(\mathcal T^\pi\) is a \(\tau\)-contraction on \(\mathcal C_E\). Conversely, assume that for every MDP with discount \(\gamma\) under every fixed policy, whenever \(\mathcal T^\pi(\mathcal C_E)\subseteq \mathcal C_E\), the Bellman operator is a \(\tau\)-contraction on \(\mathcal C_E\). Then, for every \(q\in\R^d\) such that \(q+\gamma E\subseteq E\), the kernel \(k_{\mathrm{diff}}^{q,\gamma}\) is CPD on \(E\). Proposition~\ref{prop:bounded_return_ball} gives one natural situation in which such an invariant set \(E\) exists.
\end{theorem}

\begin{proof}
\noindent\paragraph{Sufficiency.}
Fix \(\eta_1,\eta_2\in\mathcal C_E\). By Theorem~\ref{thm:mmd-contraction-cpd}, applied with \(X=\R^d\), the set \(E\), and
\[
f_q(x)=q+\gamma x,
\]
the assumption that \(k_{\mathrm{diff}}^{q,\gamma}\) is CPD on \(E\) implies that, for every \(q\in\R^d\),
\[
\MMD_k^2\bigl(f_q\#\nu,f_q\#\mu\bigr)
\le
\tau\,\MMD_k^2(\nu,\mu)
\qquad
\forall\,\mu,\nu\in\mathcal P(\R^d)\text{ with }\mathrm{supp}\,\mu,\mathrm{supp}\,\nu\subseteq E.
\]
Since \(\eta_1,\eta_2\in\mathcal C_E\), the assumptions of Lemma~\ref{lem:bellman_pointwise_lift} are satisfied with the same set \(E\). Applying that lemma, we obtain, for every \((s,a)\in\mathcal S\times\mathcal A\),
\[
\MMD_k^2\bigl((\mathcal T^\pi\eta_1)(s,a),(\mathcal T^\pi\eta_2)(s,a)\bigr)
\le
\tau\,\overline{\MMD}_k^2(\eta_1,\eta_2).
\]
Taking the supremum over \((s,a)\) yields
\[
\overline{\MMD}_k^2(\mathcal T^\pi\eta_1,\mathcal T^\pi\eta_2)
\le
\tau\,\overline{\MMD}_k^2(\eta_1,\eta_2),
\]
which proves the boxed inequality.

If, in addition, \(\mathcal T^\pi(\mathcal C_E)\subseteq \mathcal C_E\), then \(\mathcal T^\pi\) maps \(\mathcal C_E\) into itself and satisfies the above inequality on \(\mathcal C_E\). Hence \(\mathcal T^\pi\) is a \(\tau\)-contraction on \(\mathcal C_E\).

\noindent\paragraph{Necessity.}
Suppose that, for every MDP with discount \(\gamma\) under every fixed policy, the Bellman operator is a \(\tau\)-contraction on \(\mathcal C_E\) whenever it maps \(\mathcal C_E\) into itself. We show that this forces \(k_{\mathrm{diff}}^{q,\gamma}\) to be CPD on \(E\) for every \(q\in\R^d\) satisfying
\[
q+\gamma E\subseteq E.
\]

Fix such a \(q\). By Lemma~\ref{lem:deterministic_bellman_embedding}, there exists a discounted MDP with a single state \(s\), a single action \(a\) whose transition always returns to \(s\), deterministic reward \(q\), and its unique fixed policy \(\pi\), such that
\[
(\mathcal T^\pi\eta)(s,a)=f_q\#\eta(s,a),
\qquad
f_q(x)=q+\gamma x.
\]
Moreover, this Bellman operator preserves laws supported on \(E\). Indeed, if \(\eta\in\mathcal C_E\), then \(\eta(s,a)\) is supported on \(E\), and since \(q+\gamma E\subseteq E\), the pushforward \(f_q\#\eta(s,a)\) is also supported on \(E\). Hence
\[
\mathcal T^\pi(\mathcal C_E)\subseteq \mathcal C_E.
\]
Therefore, the assumption applies to this MDP. Since \(\mathcal S\times\mathcal A=\{(s,a)\}\), one has
\[
\overline{\MMD}_k^2(\eta_1,\eta_2)
=
\MMD_k^2\bigl(\eta_1(s,a),\eta_2(s,a)\bigr)
\]
for all \(\eta_1,\eta_2\in\mathcal C_E\). The assumed \(\tau\)-contraction on \(\mathcal C_E\) therefore implies
\[
\MMD_k^2\bigl(f_q\#\nu,f_q\#\mu\bigr)
\le
\tau\,\MMD_k^2(\nu,\mu)
\qquad
\forall\,\mu,\nu\in\mathcal P(\R^d)\text{ with }\mathrm{supp}\,\mu,\mathrm{supp}\,\nu\subseteq E.
\]
Applying Theorem~\ref{thm:mmd-contraction-cpd} with the set \(E\), we conclude that \(k_{\mathrm{diff}}^{q,\gamma}=\tau k-k_{f_q}\) is CPD on \(E\). Since \(q\) was arbitrary among those satisfying \(q+\gamma E\subseteq E\), this proves the necessity statement.
\end{proof}

\begin{lemma}[CPD on a set implies CPD on all subsets]
\label{lem:cpd_restriction_monotonicity}
Let \(X\) be a measurable space, let \(k:X\times X\to\R\) be symmetric, and let \(F\subseteq E\subseteq X\) be measurable. If \(k\) is CPD on \(E\), then \(k\) is CPD on \(F\).
\end{lemma}

\begin{proof}
Assume that \(k\) is CPD on \(E\). Let \(\mu\) be any finite signed measure supported on \(F\) with \(\mu(F)=0\). Since \(F\subseteq E\), the measure \(\mu\) is also supported on \(E\). Moreover,
\[
\mu(E)=\mu(F)=0,
\]
because \(\mu\) vanishes outside \(F\). Since \(k\) is CPD on \(E\), it follows that
\[
\iint_{E\times E} k(x,y)\,d\mu(x)\,d\mu(y)\ge 0.
\]
As \(\mu\) is supported on \(F\), this is the same as
\[
\iint_{F\times F} k(x,y)\,d\mu(x)\,d\mu(y)\ge 0.
\]
Hence \(k\) is CPD on \(F\).
\end{proof}

\newpage

\section{Contraction on a bounded support: numerical experiments}\label{sec:bounded_support_numerical_experiments}
\paragraph{Preamble.}
This section turns the criterion of Theorem~\ref{thm:bellman_contraction_cpd_support}, CPD on the bounded set \(E\), into a numerical procedure for estimating \(\tau_E^\star(\lambda)\). Section~\ref{app:from_bounded_support_to_discrete_support} replaces \(E=B(0,R)\) by random finite sets of points in \(E\), which by Lemma~\ref{lem:cpd_restriction_monotonicity} give lower bounds on \(\tau_E^\star(\lambda)\) that can only improve as more sets are drawn. Section~\ref{app:from_support_to_matrix} reduces the CPD condition on a finite support to a generalized eigenvalue problem, solved by the symmetric reduction of Algorithm~\ref{alg:tau-star-mc}. Section~\ref{app:bounded_results} validates the procedure against the power-distance and generalized multiquadric kernels of \cite{nguyen2021distributional,killingberg2023the}, whose contraction factors are known. Section~\ref{app:experimental_setup} gives the settings behind Figure~\ref{fig:main_bounded_support_heatmaps}, which applies the procedure to the RBF and logarithmic kernels.
\subsection{From a bounded to a discrete support condition}\label{app:from_bounded_support_to_discrete_support}
By Theorem~\ref{thm:mmd-contraction-cpd}, for a fixed map, \(\MMD^2\) contraction on a set \(E\subseteq\mathbb R^d\) is equivalent to the corresponding difference kernel being CPD on \(E\). Theorem~\ref{thm:bellman_contraction_cpd_support} gives Bellman contraction on \(E\) when this holds for the map \(x\mapsto q+\gamma x\) of every reward value \(q\).

For a translation-invariant kernel, \(k(q+\lambda x,q+\lambda y)=k(\lambda x,\lambda y)\), so the difference kernel is the same for every reward value \(q\). We therefore take \(q=0\) and \(f(x)=\lambda x\), and the relevant difference kernel is
\[
k_{\mathrm{diff},\tau}^{0,\lambda}(x,y):=\tau k(x,y)-k(\lambda x,\lambda y),
\qquad \lambda\in(0,1).
\]
For each measurable set \(E\subseteq\mathbb R^d\), define
\[
\tau_E^\star(\lambda)
:=
\inf\Bigl\{\tau>0:\ k_{\mathrm{diff},\tau}^{0,\lambda}\text{ is CPD on }E\Bigr\}.
\]
By Lemma~\ref{lem:cpd_restriction_monotonicity}, CPD on a set implies CPD on every subset, so \(\tau_E^\star(\lambda)\) can only increase when \(E\) grows. In particular,
\[
\tau_{\mathbb R^d}^\star(\lambda)\ge \tau_E^\star(\lambda)
\qquad\text{for every measurable }E\subseteq\mathbb R^d.
\]
By Theorem~\ref{thm:cpd_contraction}, for a translation-invariant kernel and \(\lambda=\gamma\), \(\tau_{\mathbb R^d}^\star(\gamma)\) is the smallest contraction factor \(\tau_k^\star(\mathcal T^\pi)\) of the main text.
Thus, requiring CPD only on a bounded set \(E\) is less restrictive than requiring it on all of \(\mathbb R^d\), so the condition can hold on \(E\) for a kernel that fails it on all of \(\mathbb R^d\). In reinforcement
learning, Proposition~\ref{prop:bounded_return_ball} shows that under bounded rewards, return laws are
supported in a ball \(E=B_M\) with \(M\ge r_{\max}/(1-\gamma)\).

The CPD condition on \(E\) concerns all signed measures supported on \(E\), so it cannot be checked numerically. We test it instead on \(J\) finite sets of points drawn at random in \(E\),
\[
S_j=\{z_1^{(j)},\dots,z_N^{(j)}\}\subseteq E,
\qquad j=1,\dots,J,
\]
and define the contraction threshold of each set,
\[
\tau_{S_j}^\star(\lambda)
:=
\inf\Bigl\{\tau>0:\ k_{\mathrm{diff},\tau}^{0,\lambda}\text{ is CPD on }S_j\Bigr\}.
\]
Since each \(S_j\) is a subset of \(E\), Lemma~\ref{lem:cpd_restriction_monotonicity} gives
\[
\tau_{\mathbb R^d}^\star(\lambda)\ge \tau_E^\star(\lambda)\ge \max_{1\le j\le J}\tau_{S_j}^\star(\lambda).
\]
So the largest of these thresholds is a lower bound on \(\tau_E^\star(\lambda)\). It can only increase with \(J\), because the maximum is taken over more sets.

\subsection{From a support condition to a matrix test}\label{app:from_support_to_matrix}

Fix a finite set of points \(S=\{z_1,\dots,z_N\}\subseteq E\). We call it a finite support, because the CPD condition on \(S\) is the contraction inequality for laws whose mass sits on these points. The threshold
\(\tau_S^\star(\lambda)\) of \(S\) is determined by whether the difference kernel
\[
k_{\mathrm{diff},\tau}^{0,\lambda}(x,y)=\tau k(x,y)-k(\lambda x,\lambda y)
\]
is CPD on \(S\). We now express this condition in matrix form.

\paragraph{Checking the CPD condition on \(S\).}
Define the Gram matrices
\[
K\in\mathbb R^{N\times N},
\qquad
K_{ij}:=k(z_i,z_j),
\qquad
K_\lambda\in\mathbb R^{N\times N},
\qquad
(K_\lambda)_{ij}:=k(\lambda z_i,\lambda z_j),
\]
and the corresponding \emph{difference matrix}
\[
D_\tau^{(\lambda)}:=\tau K-K_\lambda.
\]
A signed measure on \(S\) is \(\sigma=\sum_{i=1}^N w_i\,\delta_{z_i}\), with \(\sigma(S)=\mathbf 1_N^\top w\) and \(\iint k_{\mathrm{diff},\tau}^{0,\lambda}\,d\sigma\,d\sigma=w^\top D_\tau^{(\lambda)} w\). The CPD condition on \(S\) is therefore
\[
w^\top D_\tau^{(\lambda)} w \ge 0
\qquad
\text{for all }w\in\mathbf 1_N^\perp,
\qquad
\mathbf 1_N^\perp:=\{w\in\mathbb R^N:\mathbf 1_N^\top w=0\}.
\]
Using the centering projector \(H_N=I_N-\frac{1}{N}\mathbf 1_N\mathbf 1_N^\top\) introduced in
Section~\ref{app:mmd_gram_matrices}, this is equivalent to
\[
H_N D_\tau^{(\lambda)} H_N \succeq 0.
\]
A symmetric matrix is positive semidefinite if and only if its smallest eigenvalue is nonnegative. Since \(H_N D_\tau^{(\lambda)} H_N\) is symmetric, the CPD condition on \(S\) therefore holds if and only if
\[
\lambda_{\min}\!\bigl(H_N D_\tau^{(\lambda)} H_N\bigr)\ge 0.
\]
\paragraph{Monotonicity in \(\tau\).}
Fix \(\tau'\ge \tau\) and any \(w\in \mathbf 1_N^\perp\). Since \(D_\tau^{(\lambda)}=\tau K-K_\lambda\), we have
\[
w^\top D_{\tau'}^{(\lambda)} w
=
w^\top D_\tau^{(\lambda)} w + (\tau'-\tau)\,w^\top K w.
\]
Thus, if \(\tau'>\tau\) and \(w^\top K w>0\), increasing \(\tau\) strictly increases \(w^\top D_\tau^{(\lambda)} w\).
More generally, if the CPD condition on \(S\) holds for \(\tau\), it also holds for every \(\tau'\ge\tau\) whenever
\[
w^\top K w\ge 0
\qquad
\forall\,w\in \mathbf 1_N^\perp,
\]
that is, whenever \(k\) is CPD on \(S\). This is the case for every sampled support when \(k\) is CPD on \(\mathbb R^d\).
\paragraph{Working directly with vectors whose entries sum to zero.}
The CPD condition on \(S\) concerns the quadratic form \(w^\top D_\tau^{(\lambda)} w\) only for
weights \(w\) satisfying \(\mathbf 1_N^\top w=0\), that is, \(w\in \mathbf 1_N^\perp\).
Although the centering projector \(H_N\) encodes this constraint, we do not form it explicitly.
Instead, we choose an orthonormal basis \(Q\in\mathbb R^{N\times(N-1)}\) of \(\mathbf 1_N^\perp\), so that
\[
Q^\top Q = I_{N-1},
\qquad
\mathrm{Range}(Q)=\mathbf 1_N^\perp.
\]
Since every \(w\in\mathbf 1_N^\perp\) is \(w=Qv\) for some \(v\in\mathbb R^{N-1}\), and \((Qv)^\top D_\tau^{(\lambda)}(Qv)=v^\top\bigl(Q^\top D_\tau^{(\lambda)} Q\bigr)v\), the CPD condition on \(S\) holds if and only if \(Q^\top D_\tau^{(\lambda)} Q \succeq 0\).
\paragraph{Generalized eigenvalues for vectors whose entries sum to zero.}
Write the two Gram matrices restricted to \(\mathbf 1_N^\perp\) in the basis \(Q\),
\[
A:=Q^\top KQ,
\qquad
B:=Q^\top K_\lambda Q,
\]
so that
\[
Q^\top D_\tau^{(\lambda)}Q=\tau A-B.
\]
Hence the CPD condition on \(S\) is equivalent to
\[
\tau A-B\succeq 0.
\]
If \(A\) is strictly positive definite, meaning
\[
v^\top A v>0
\qquad
\text{for every }v\neq 0,
\]
or equivalently, as shown in the next paragraph,
\[
w^\top K w>0
\qquad
\text{for every nonzero }w\in \mathbf 1_N^\perp,
\]
then, for every nonzero \(v\),
\[
v^\top(\tau A-B)v\ge 0
\qquad\Longleftrightarrow\qquad
\tau \ge \frac{v^\top Bv}{v^\top Av}.
\]
Therefore,
\[
\tau_S^\star(\lambda)
=
\sup_{v\neq 0}\frac{v^\top Bv}{v^\top Av}
=
\lambda_{\max}(B,A),
\]
that is, the smallest \(\tau\) for which the CPD condition holds on \(S\) is exactly the largest generalized eigenvalue of \((B,A)\).

\paragraph{CSPD and strict positivity of \(A\).}
The assumption \(A\succ 0\) is equivalent to requiring that \(k\) be CSPD on the finite support \(S\) in the sense of Definition~\ref{def:cpd-cspd}. Indeed, since the columns of \(Q\) form an orthonormal basis of \(\mathbf 1_N^\perp\), every \(w\in \mathbf 1_N^\perp\) can be written uniquely as \(w=Qv\) with \(v\in\mathbb R^{N-1}\), and \(w\neq 0\) if and only if \(v\neq 0\). Therefore,
\[
v^\top A v
=
v^\top Q^\top KQv
=
(Qv)^\top K(Qv)
=
w^\top K w.
\]
Hence \(A\) is strictly positive definite if and only if
\[
w^\top K w>0
\qquad
\text{for every nonzero }w\in \mathbf 1_N^\perp,
\]
that is, if and only if \(k\) is CSPD on \(S\). In particular, this condition is automatically satisfied if \(k\) is CSPD globally. By Proposition~\ref{prop:mmd-metric}, this is equivalent to \(\MMD_k\) being a metric, and we assume it in the rest of this section.

\paragraph{Reduction to an ordinary symmetric eigenvalue problem.}
Under the assumption that \(k\) is CSPD, \(A\) is strictly positive definite. We can therefore reduce the generalized eigenvalue problem to an ordinary symmetric one, using the eigendecomposition of \(A\)
\cite[Theorem~15.3.2, Chapter~15.4, method~2]{parlett1998symmetric},
\[
A = P \Delta^2 P^\top,
\qquad
\Delta := \mathrm{diag}(\mu_1^{1/2},\dots,\mu_{N-1}^{1/2}),
\qquad \mu_i>0.
\]
We then introduce the change of coordinates \(u := \Delta P^\top v\), or equivalently \(v = P \Delta^{-1} u\). With \(C:=\Delta^{-1}P^\top B P \Delta^{-1}\), the generalized eigenvalue problem \(Bv=\tau Av\) is transformed into the ordinary symmetric eigenvalue problem \(Cu=\tau u\).
Indeed,
\begin{align*}
Bv
&= B P \Delta^{-1} u \\
&= \tau A v \\
&= \tau A P \Delta^{-1} u \\
&= \tau P \Delta^2 P^\top P \Delta^{-1} u \\
&= \tau P \Delta u.
\end{align*}
Applying \(\Delta^{-1}P^\top\) to both sides of the identity
\[
B P \Delta^{-1} u = \tau P \Delta u
\]
yields
\[
\Delta^{-1}P^\top B P \Delta^{-1} u = \tau u,
\]
that is,
\[
Cu=\tau u.
\]
So the threshold of \(S\) is the largest eigenvalue of \(C\),
\[
\tau_S^\star(\lambda)=\lambda_{\max}(C).
\]
\paragraph{Truncation of very small eigenvalues.}
Although \(A\) is strictly positive definite, some of its eigenvalues can be extremely small in practice. The factors \(\mu_i^{-1/2}\) in the reduction above then become very large and make the reduced problem numerically unstable, a difficulty also discussed in \cite[Chapter~15.4.1, Remark~5]{parlett1998symmetric}. We therefore discard very small eigenvalues before forming the reduced matrix, following the idea of \cite[Chapter~15.5]{parlett1998symmetric}.

Concretely, we compute
\[
A = P \,\mathrm{diag}(\mu_1,\dots,\mu_{N-1})\,P^\top,
\]
choose a relative tolerance \(\varepsilon_{\mathrm{eig}}>0\), and retain only the eigenvalues satisfying
\[
\mu_i > \varepsilon_{\mathrm{eig}}\,\mu_{\max},
\qquad
\mu_{\max}:=\max_i \mu_i.
\]
Setting \(\mu_i^{-1/2}=0\) for discarded eigenvalues, we define
\[
\Delta_{\varepsilon}^{-1}
:=
\mathrm{diag}(\nu_1,\dots,\nu_{N-1}),
\qquad
\nu_i=
\begin{cases}
\mu_i^{-1/2}, & \mu_i > \varepsilon_{\mathrm{eig}}\,\mu_{\max},\\
0, & \text{otherwise}.
\end{cases}
\]
We then form
\[
C_{\varepsilon} := \Delta_{\varepsilon}^{-1} P^\top B P \Delta_{\varepsilon}^{-1}.
\]
This gives the truncated reduced problem, and we define the corresponding numerical estimate by
\[
\widehat{\tau}_S^\star(\lambda):=\lambda_{\max}(C_{\varepsilon}).
\]
If no eigenvalues are discarded, then \(\widehat{\tau}_S^\star(\lambda)=\tau_S^\star(\lambda)\). Otherwise, let \(V\) be the space spanned by the eigenvectors of \(A\) that are kept. Then
\[\widehat{\tau}_S^\star(\lambda)=\max_{v\in V,\ v\neq 0}\frac{v^\top Bv}{v^\top Av}\le\max_{v\neq 0}\frac{v^\top Bv}{v^\top Av}=\tau_S^\star(\lambda),\]
since the first maximum is taken over fewer vectors. So the estimate can only underestimate the threshold.

\subsection{Experimental setup}\label{app:experimental_setup}
Motivated by Proposition~\ref{prop:bounded_return_ball}, we take
\[
E = B(0,R) := \{x\in\R^d:\|x\|\le R\}.
\]
Fix a support size \(N\) and a trial count \(J\). For each \(j\in\{1,\dots,J\}\), we draw an independent random support
\[
S_j=\{z^{(j)}_1,\dots,z^{(j)}_N\}\subseteq B(0,R),
\qquad
z^{(j)}_i \overset{\mathrm{i.i.d.}}{\sim} \mathrm{Unif}\bigl(B(0,R)\bigr).
\]
Each sampled support \(S_j\) induces a threshold \(\tau_{S_j}^\star(\lambda)\). Our implementation computes instead the numerical approximation
\[
\widehat{\tau}_{S_j}^\star(\lambda)\approx \tau_{S_j}^\star(\lambda)
\]
obtained from the truncated symmetric problem described above. We report the largest of these values,
\[
\widehat{\tau}_E^\star(\lambda):=\max_{1\le j\le J}\widehat{\tau}_{S_j}^\star(\lambda),
\]
which estimates the lower bound \(\max_{j}\tau_{S_j}^\star(\lambda)\) on \(\tau_E^\star(\lambda)\) of Section~\ref{app:from_bounded_support_to_discrete_support}.
Algorithm~\ref{alg:tau-star-mc} summarizes this process, and Table~\ref{tab:fixed_support_rejection_heatmap_settings} gives the numerical settings used for the heatmaps of Figure~\ref{fig:main_bounded_support_heatmaps}.

\begin{table}[t]
\centering
\caption{Numerical settings for the heatmaps of Figure~\ref{fig:main_bounded_support_heatmaps}.}
\label{tab:fixed_support_rejection_heatmap_settings}
\begin{tabular}{ll}
\toprule
Setting & Value \\
\midrule
Radius & \(R=1\) \\
Dimension & \(d=1\) \\
Support sampler & Uniform on the ball of radius \(R\) \\
Support size & \(N=512\) \\
Number of sampled supports & \(J=10\) \\
Parameter grid & \(R/\ell \in [0.05,2.00]\), 81 linearly spaced values \\
Scaling grid & \(\lambda \in [0,1]\), 81 linearly spaced values \\
Eigenvalue tolerance \(\varepsilon_{\mathrm{eig}}\) & \(10^{-9}\) \\
Hatched cells (contraction ruled out) & \(\widehat{\tau}_E^{\star}(\lambda)>1+10^{-6}\) \\
\bottomrule
\end{tabular}
\end{table}
\paragraph{Compute resources.}
The experiments of this section were implemented in JAX \citep{jax2018github} and run on CPU using float64 precision. For a fixed kernel configuration, computing one contraction curve took at most one minute.

\begin{algorithm}[t]
\caption{Monte Carlo sampling of \(\widehat{\tau}_{S_j}^\star(\lambda)\) from CPD tests on finite supports}
\label{alg:tau-star-mc}
\begin{algorithmic}[1]
  \STATE {\bfseries Require:} Dimension \(d\ge 1\), support size \(N\ge 2\), trial count \(J\ge 1\)
  \STATE {\bfseries Require:} Radius \(R>0\) defining \(E=B(0,R)\subset\R^d\)
  \STATE {\bfseries Require:} Scaling parameter \(\lambda\in(0,1)\)
  \STATE {\bfseries Require:} Symmetric kernel \(k:\R^d\times\R^d\to\R\)
  \STATE {\bfseries Require:} Relative tolerance \(\varepsilon_{\mathrm{eig}}>0\)
  \STATE {\bfseries Output:} Samples \(\{\widehat{\tau}_{S_j}^\star(\lambda)\}_{j=1}^J\)

  \STATE Compute an orthonormal basis \(Q\in\R^{N\times(N-1)}\) of \(\mathbf 1_N^\perp\)

  \FOR{$j=1,\dots,J$}
    \STATE \textcolor{green!50!black}{\textit{Sample a support and build the corresponding restricted matrices.}}
    \STATE Draw \(S_j=\{z_1^{(j)},\dots,z_N^{(j)}\}\) with \(z_i^{(j)} \overset{\mathrm{i.i.d.}}{\sim}\mathrm{Unif}(B(0,R))\)

    \STATE Form the Gram matrices
    \[
      K_{ab}\leftarrow k\!\left(z_a^{(j)},z_b^{(j)}\right),
      \qquad
      (K_\lambda)_{ab}\leftarrow k\!\left(\lambda z_a^{(j)},\lambda z_b^{(j)}\right)
    \]

    \STATE Restrict to \(\mathbf 1_N^\perp\)
    \[
      A \leftarrow Q^\top K Q,
      \qquad
      B \leftarrow Q^\top K_\lambda Q
    \]

    \STATE \textcolor{green!50!black}{\textit{Reduce the generalized eigenvalue problem to a truncated ordinary symmetric one.}}
    \STATE Compute an eigendecomposition
    \[
      A = P\,\mathrm{diag}(\mu_1,\dots,\mu_{N-1})\,P^\top
    \]

    \STATE Set
    \[
      \mu_{\max}\leftarrow \max_{1\le i\le N-1}\mu_i
    \]

    \STATE Set the truncated diagonal matrix
    \[
      \Delta_{\varepsilon}^{-1}
      \leftarrow
      \mathrm{diag}(\nu_1,\dots,\nu_{N-1}),
      \qquad
      \nu_i =
      \begin{cases}
        \mu_i^{-1/2}, & \mu_i > \varepsilon_{\mathrm{eig}}\,\mu_{\max},\\
        0, & \text{otherwise}
      \end{cases}
    \]

    \STATE Form the reduced symmetric matrix
    \[
      C_\varepsilon \leftarrow \Delta_{\varepsilon}^{-1} P^\top B P \Delta_{\varepsilon}^{-1}
    \]

    \STATE \textcolor{green!50!black}{\textit{Extract the threshold estimate from the largest eigenvalue.}}
    \STATE Compute
    \[
      \widehat{\tau}_{S_j}^\star(\lambda)\leftarrow \lambda_{\max}(C_\varepsilon)
    \]
  \ENDFOR

  \STATE {\bfseries Return:} \(\{\widehat{\tau}_{S_j}^\star(\lambda)\}_{j=1}^J\)
\end{algorithmic}
\end{algorithm}

\subsection{Results}\label{app:bounded_results}
\begin{figure}[t]
    \centering
    \includegraphics[width=\textwidth]{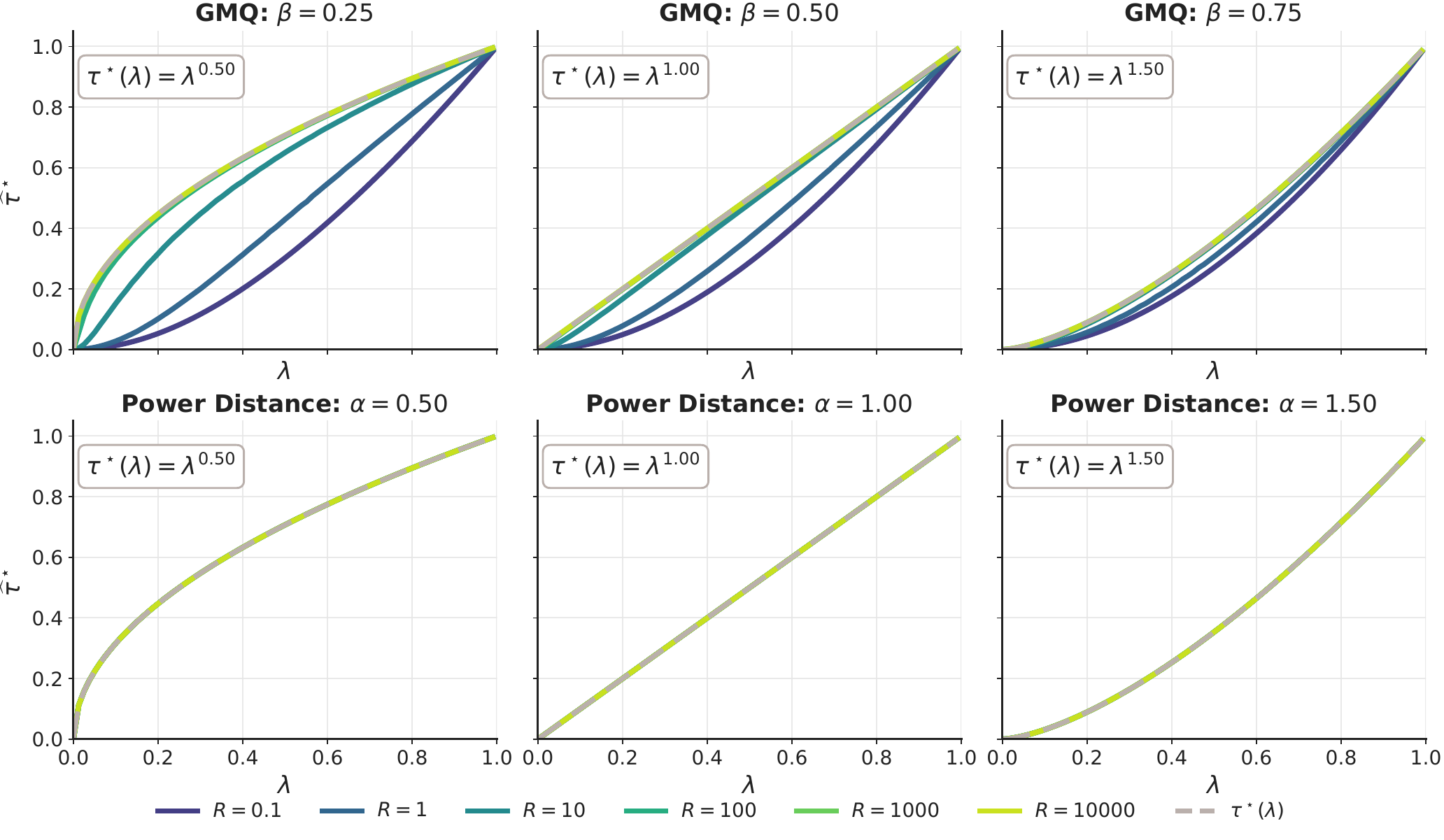}
    \caption{Estimates on balls \(B(0,R)\) for kernels with known smallest contraction factors across dimensions. As \(R\) increases, the estimates approach the theoretical factors.}
    \label{fig:bounded_support_theoretical_confirmation}
\end{figure}

\paragraph{Verification on known kernels.}
The method introduced in this section was designed to estimate a lower bound on
\(\tau^\star_E(\lambda)\), where \(E=B(0,R)\). As \(R\) increases, the set
\(E\) approaches the ambient space, so the estimate on \(B(0,R)\) should approach the
smallest contraction factor on all of \(\mathbb R^d\),
\[
    \lim_{R\to\infty} \tau^\star_{B(0,R)}(\lambda) = \tau^\star_{\mathbb R^d}(\lambda).
\]
To validate Algorithm~\ref{alg:tau-star-mc}, we apply it to the power-distance and generalized multiquadric kernels,
whose smallest contraction factor across dimensions \(\tau^\star(\lambda)\) is known (Propositions~\ref{prop:power_distance_contraction} and~\ref{prop:gmq_contraction}) and bounds \(\tau^\star_{\mathbb R^d}(\lambda)\) from above. Figure~\ref{fig:bounded_support_theoretical_confirmation}
shows the estimates obtained on \(B(0,R)\) for increasing values of \(R\). For
the power-distance kernels, the estimates already match
\(\tau^\star(\lambda)\). For the generalized multiquadric kernels, the estimates are
smaller on small balls and approach \(\tau^\star(\lambda)\) as \(R\) increases. This
is consistent with the fact that CPD on \(B(0,R)\) is weaker than CPD on the
whole ambient space.

\clearpage
\newpage

\section{Global contraction under radial kernels via Bernstein functions}\label{app:contraction_radial_kernels_via_Bernstein}
\paragraph{Preamble.}
This section studies contraction under radial kernels. It turns the CPD criterion of Appendix~\ref{app:general_contraction} into conditions on the Bernstein function that generates the kernel.

Section~\ref{app:motivations_for_radial_kernels} explains why we focus on radial kernels and shows that contraction under an affine map then reduces to a CND condition on the difference profile (Proposition~\ref{prop:radial_contraction}). Section~\ref{app:bernstein_schoenberg} collects the necessary background, namely the definitions of Bernstein and complete Bernstein functions and Schoenberg's theorem (Theorem~\ref{thm:schoenberg-bernstein}), by which a radial kernel is CND exactly when its profile is a Bernstein function. In Section~\ref{app:bernstein_reduction_of_difference_kernel}, Theorem~\ref{thm:bernstein_reduction} shows that contraction under affine maps, in every dimension, holds exactly when the difference profile $\psi_\tau = \tau\phi - \phi(\lambda^2\,\cdot)$ is a Bernstein function, and Proposition~\ref{prop:no_contraction_saturating_BF} shows that saturating profiles cannot yield strict contraction.

Section~\ref{app:measure_domination} proves the results on L\'evy measures needed to make the Bernstein criterion explicit. Section~\ref{app:bernstein_reduction_to_contraction} uses them to express the criterion as pointwise inequalities on the drift and the L\'evy measure (Theorem~\ref{thm:contraction-drift-density-atoms}). Section~\ref{app:contraction_complete_bernstein} treats complete Bernstein functions, for which the smallest contraction factor across dimensions reduces to a single supremum of the L\'evy density ratio (Theorem~\ref{thm:cbf_contraction_sup}). Section~\ref{app:chaining_general_mechanism} introduces geometric chaining, which builds Bernstein profiles that induce contraction with a chosen contraction factor (Theorem~\ref{thm:general_geometric_chaining_factory}).

\subsection{Motivations for radial kernels}\label{app:motivations_for_radial_kernels}
For scalar returns (\(d=1\)), the following proposition shows that a symmetric kernel that is invariant under translations is radial. Hence, in one dimension, focusing on radial kernels covers every translation-invariant kernel.

\begin{proposition}[A symmetric, translation-invariant univariate kernel is radial]\label{prop:1d_ti_sym_radial}
Let \(k:\R\times\R\to\R\).

\begin{enumerate}\itemsep0.2em
\item[(i)] \textbf{Translation invariance:} \(k(x+t,y+t)=k(x,y)\) for all \(x,y,t\in\R\).
\item[(ii)] \textbf{Symmetry:} \(k(x,y)=k(y,x)\) for all \(x,y\in\R\).
\end{enumerate}

Then \((i)\) and \((ii)\) hold if and only if there exists a function \(g:[0,\infty)\to\R\) such that
\[
k(x,y)=g(|x-y|)\qquad \forall\,x,y\in\R.
\]
\end{proposition}

\begin{proof}
\noindent\paragraph{(\(\Rightarrow\)) Necessity.}
Assume translation invariance. Define \(\kappa:\R\to\R\) by \(\kappa(r):=k(r,0)\).
Then for any \(x,y\in\R\), applying translation invariance with \(t=-y\) gives
\[
k(x,y)=k(x-y,0)=\kappa(x-y).
\]
Assume also that \(k\) is symmetric. Then for all \(r\in\R\),
\[
\kappa(r)=k(r,0)=k(0,r)=k(-r,0)=\kappa(-r),
\]
where the step \(k(0,r)=k(-r,0)\) uses translation invariance with \(t=-r\).
Thus \(\kappa\) is even. Define \(g:[0,\infty)\to\R\) by \(g(u):=\kappa(u)\) for \(u\ge 0\).
Since \(\kappa\) is even, \(\kappa(r)=g(|r|)\) for all \(r\in\R\), and therefore
\[
k(x,y)=\kappa(x-y)=g(|x-y|).
\]

\noindent\paragraph{(\(\Leftarrow\)) Sufficiency.}
Conversely, if \(k(x,y)=g(|x-y|)\), then for any \(t\in\R\),
\[
k(x+t,y+t)=g(|(x+t)-(y+t)|)=g(|x-y|)=k(x,y),
\]
so \(k\) is translation invariant. Also,
\[
k(y,x)=g(|y-x|)=g(|x-y|)=k(x,y),
\]
so \(k\) is symmetric.
\end{proof}

\begin{proposition}[For radial kernels, contraction under an affine map reduces to a CND condition]\label{prop:radial_contraction}
Fix \(d\ge 1\). Assume \(f\colon \R^d\to\R^d\) is an affine map of the form
\[
f(x)=q+\lambda x,
\qquad q\in\R^d,\ \lambda\in\R.
\]
Let \(k(u,v)=g(\|u-v\|)\) be a radial kernel on \(\R^d\), and define
\[
h(r)=g(0)-g(r),\quad r\ge0.
\]
As in Theorem~\ref{thm:mmd-contraction-cpd}, assume \(k\) is CPD on \(\R^d\), so that
\((u,v)\mapsto h(\|u-v\|)\) is CND on \(\R^d\).
\begin{mdframed}[linewidth=0.4pt,leftmargin=1.5em,rightmargin=1.5em,innerleftmargin=8pt,innerrightmargin=8pt,innertopmargin=6pt,innerbottommargin=6pt,skipabove=8pt,skipbelow=8pt]
Then, for any \(\tau\in\R\), the contraction inequality
\[
\MMD^2_{k}\bigl(f\#\nu,f\#\mu\bigr)
\;\le\;
\tau\,\MMD^2_{k}(\nu,\mu)
\]
holds for every \(\mu,\nu\in\mathcal P(\R^d)\) if and only if
\[
\Gamma_\tau(x,y)
=
\tau\,h\bigl(\|x-y\|\bigr)
-\;h\bigl(|\lambda|\,\|x-y\|\bigr)
\ \text{is CND on }\R^d.
\]
\end{mdframed}
Moreover, \(\Gamma_\tau(x,x)=0\).
\end{proposition}

\begin{proof}
Since \(f(x)=q+\lambda x\), we have
\[
\|f(x)-f(y)\| = |\lambda|\,\|x-y\|
\quad\text{for all }x,y\in\R^d.
\]

By Theorem~\ref{thm:mmd-contraction-cpd}, applied with \(E=\R^d\), the contraction statement
\[
\MMD^2_{k}\bigl(f\#\nu,f\#\mu\bigr)
\;\le\;
\tau\,\MMD^2_{k}(\nu,\mu)
\qquad \forall\,\mu,\nu\in\mathcal P(\R^d)
\]
is equivalent to \(\Gamma_\tau\) being CND on \(\R^d\), where
\[
\Gamma_\tau(x,y)
=
\tau\,h\bigl(\|x-y\|\bigr)
-\;h\bigl(\|f(x)-f(y)\|\bigr).
\]
Substituting \(\|f(x)-f(y)\|=|\lambda|\,\|x-y\|\) yields
\[
\Gamma_\tau(x,y)
=
\tau\,h\bigl(\|x-y\|\bigr)
-\;h\bigl(|\lambda|\,\|x-y\|\bigr).
\]

Finally, \(h(0)=g(0)-g(0)=0\), hence
\[
\Gamma_\tau(x,x)=\tau\,h(0)-h(0)=0.
\]
\end{proof}

\subsection{Definitions and Schoenberg's theorem}
\label{app:bernstein_schoenberg}
\begin{definition}[Bernstein function]\label{def:bernstein}
A function \(\phi\colon [0,\infty)\to[0,\infty)\) is called a \emph{Bernstein function} \cite[Def.~3.1]{schilling2009bernstein} if it is continuous at \(0\), infinitely differentiable on \((0,\infty)\), and its derivative is completely monotone,
\[
\phi\in C^\infty((0,\infty)),
\quad
(-1)^{n-1}\,\frac{d^n}{dt^n}\phi(t)\;\ge\;0
\quad\forall\,n\ge1,\;t>0.
\]
Equivalently, it admits the L\'evy--Khintchine representation \cite[Thm.~3.2]{schilling2009bernstein}
\[
\phi(t)
=
a + b\,t
+
\int_{(0,\infty)}\bigl(1 - e^{-t\,s}\bigr)\,\Lambda(d s),
\qquad a\ge 0,\; b\ge 0,
\]
together with the integrability condition
\[
\int_{(0,\infty)}\min\{1,s\}\,\Lambda(d s) < \infty,
\]
and moreover, by \cite[Rem.~3.3(iv)]{schilling2009bernstein}, one has
\[
a = \phi(0+),
\qquad
b = \lim_{t\to\infty}\frac{\phi(t)}{t}.
\]
Thus \(b\), called the \emph{drift} of \(\phi\), determines whether \(\phi\) has linear or sublinear growth.
\end{definition}

\begin{definition}[Complete Bernstein function]\label{def:complete-bernstein}
A Bernstein function \(\phi\) (Definition~\ref{def:bernstein}) is called a \emph{complete Bernstein function}
\cite[Def.~6.1]{schilling2009bernstein} if its L\'evy measure \(\Lambda\) is absolutely continuous with a completely monotone density \(m\),
\[
\Lambda(d s)=m(s)\,d s,
\]
where
\[
m\in C^\infty((0,\infty)),
\quad
(-1)^n\,m^{(n)}(s)\;\ge\;0
\quad\forall\,n\ge0,\;s>0.
\]
\end{definition}

\begin{theorem}[Schoenberg's characterization of radial CND kernels, restated via Bernstein functions {\cite{schoenberg1938metric}}]\label{thm:schoenberg-bernstein}
Let \(h\colon[0,\infty)\to[0,\infty)\) be continuous with \(h(0)=0\), and define the radial kernel
\[
k(x,y)=h\bigl(\|x-y\|\bigr),
\quad x,y\in\R^d.
\]
Then \(k\) is conditionally negative definite on \(\R^d\) for \emph{every} \(d\ge1\) if and only if there exists a Bernstein function \(\phi\) such that
\[
h(r)=\phi(r^{2}),\quad r\ge0.
\]
In particular, \(\phi\) admits a L\'evy--Khintchine representation
\[
\phi(t)
= a + b\,t
+ \int_{(0,\infty)}\bigl(1 - e^{-s\,t}\bigr)\,\Lambda(ds),
\quad a,b\ge0.
\]
Moreover, since \(h(0)=0\), evaluating at \(r=0\) gives \(0=h(0)=\phi(0)=a\), which in turn yields
\[
\boxed{
\phi(t)
= b\,t
+ \int_{(0,\infty)}\bigl(1 - e^{-s\,t}\bigr)\,\Lambda(ds),
\quad
h(r)
= b\,r^{2}
+ \int_{(0,\infty)}\bigl(1 - e^{-s\,r^{2}}\bigr)\,\Lambda(ds).
}
\]
\end{theorem}

\subsection{Bernstein reduction of difference kernel}
\label{app:bernstein_reduction_of_difference_kernel}
\paragraph{Preamble.}
In this subsection, the contraction condition on the difference kernel is translated into a condition on a Bernstein function.
The key point is that, in order for Schoenberg's theorem to apply, the relevant radial kernel must be conditionally negative definite on \(\R^d\) for every \(d\ge 1\), not only in one fixed dimension. Theorem~\ref{thm:bernstein_reduction} shows that, when this holds in every dimension, the contraction condition is equivalent to the difference profile being a Bernstein function. Theorem~\ref{thm:levy_scaled_difference} then rewrites this criterion directly in terms of the drift term and the L\'evy measure in the L\'evy--Khintchine representation. Finally, Proposition~\ref{prop:no_contraction_saturating_BF} shows that saturating profiles cannot yield strict MMD contraction under affine maps.

\begin{theorem}[For radial kernels, contraction in every dimension reduces to a Bernstein condition]
\label{thm:bernstein_reduction}
Fix \(\lambda\in\R\). Let \(g:[0,\infty)\to\R\) be continuous, and assume that, for every \(d\ge 1\), the radial kernel
\[
k(u,v)=g(\|u-v\|),
\qquad u,v\in\R^d,
\]
is CPD on \(\R^d\). Define
\[
h(r)=g(0)-g(r),
\qquad r\ge 0.
\]
Then, for every \(d\ge 1\), the kernel \((u,v)\mapsto h(\|u-v\|)\) is CND on \(\R^d\). Hence, by Theorem~\ref{thm:schoenberg-bernstein}, there exists a Bernstein function \(\phi\) such that
\[
h(r)=\phi(r^2),
\qquad r\ge 0.
\]
For each \(d\ge 1\) and \(q\in\R^d\), consider the affine map
\[
f(x)=q+\lambda x,
\qquad x\in\R^d.
\]
In the first statement below, the kernel \(k\) is understood as the above radial kernel on the ambient space \(\R^d\). Then, for any \(\tau\in\R\), the following are equivalent.
\[
\boxed{%
\begin{gathered}
\MMD_k^2\bigl(f\#\nu,f\#\mu\bigr)
\;\le\;
\tau\,\MMD_k^2(\nu,\mu)
\\
\text{for every }d\ge 1\text{ and every }\mu,\nu\in\mathcal P(\R^d)
\\[1pt]
\text{if and only if}
\\[1pt]
\Gamma_\tau(x,y)
=
\tau\,h(\|x-y\|)
-\;h(|\lambda|\,\|x-y\|)
\ \text{is CND}
\\
\text{on }\R^d\text{ for every }d\ge 1
\\[1pt]
\text{if and only if}
\\[1pt]
\psi_\tau(t)=\tau\,\phi(t)-\phi(\lambda^2 t)
\ \text{is a Bernstein function.}
\end{gathered}
}
\]
\end{theorem}
\begin{proof}
Since, for every \(d\ge 1\), the kernel
\[
k(u,v)=g(\|u-v\|)
\]
is CPD on \(\R^d\), the kernel
\[
(u,v)\mapsto h(\|u-v\|)=g(0)-g(\|u-v\|)
\]
is CND on \(\R^d\) for every \(d\ge 1\). Theorem~\ref{thm:schoenberg-bernstein} therefore yields a Bernstein function
\(\phi\) such that
\[
h(r)=\phi(r^2),
\qquad r\ge 0.
\]

Fix \(\tau\in\R\), and let \(d\ge 1\) and \(q\in\R^d\). By Proposition~\ref{prop:radial_contraction}, applied in the fixed ambient dimension \(d\) to the affine map \(f(x)=q+\lambda x\), the first statement is equivalent to
\[
\Gamma_\tau(x,y)
=
\tau\,h(\|x-y\|)
-\;h(|\lambda|\,\|x-y\|)
\]
being CND on \(\R^d\). As \(d\ge 1\) was arbitrary, the first two statements are equivalent.

Finally, using \(h(r)=\phi(r^2)\), we may rewrite
\[
\Gamma_\tau(x,y)
=
\tau\,\phi(\|x-y\|^2)-\phi(\lambda^2\|x-y\|^2).
\]
Thus \(\Gamma_\tau(x,y)\) depends on \(x\) and \(y\) only through \(\|x-y\|^2\). We therefore define the difference profile \(\psi_\tau:[0,\infty)\to\R\) by
\[
\psi_\tau(t)=\tau\,\phi(t)-\phi(\lambda^2 t).
\]
With this notation,
\[
\Gamma_\tau(x,y)=\psi_\tau(\|x-y\|^2).
\]
Applying Theorem~\ref{thm:schoenberg-bernstein} once more, \(\Gamma_\tau\) is CND on \(\R^d\) for every \(d\ge 1\) if and only if \(\psi_\tau\) is a Bernstein function.
\end{proof}

\begin{theorem}[For the difference kernel, CND in every dimension reduces to a L\'evy--Khintchine criterion]\label{thm:levy_scaled_difference}
Let $\phi:[0,\infty)\to[0,\infty)$ be a Bernstein function with $\phi(0)=0$, admitting the L\'evy--Khintchine representation
\[
\phi(t)
=\;b\,t
\;+\;
\int_{0}^{\infty}\bigl(1-e^{-s\,t}\bigr)\,\Lambda(ds),
\]
where $b\ge0$ and $\Lambda$ is a nonnegative Borel measure on $(0,\infty)$.
Fix $\lambda\in\R\setminus\{0\}$ and $\tau\in\R$, and define the difference profile $\psi_\tau$ and the kernel $\Gamma_\tau$ built from it by
\[
\psi_\tau(t):=\tau\,\phi(t)-\phi(\lambda^2 t),
\qquad
\Gamma_\tau(x,y)=\psi_\tau\bigl(\|x-y\|^2\bigr),
\qquad x,y\in\R^d.
\]
Then
\[
\boxed{%
\begin{gathered}
(i)\quad \Gamma_\tau \text{ is CND on }\R^d\text{ for every }d\ge1,
\\[6pt]
\text{if and only if}
\\[6pt]
(ii)\quad
b(\tau-\lambda^2)\;\ge\;0
\quad\text{and}\quad
\tau\,\Lambda(B)\;\ge\;\Lambda\bigl(B/\lambda^2\bigr)
\quad\text{for every Borel set }B\subset(0,\infty).
\end{gathered}
}
\]
\end{theorem}

\begin{proof}
\noindent\paragraph{(i)\(\Rightarrow\)(ii).}
Assume \(\Gamma_\tau\) is CND on \(\R^d\) for every \(d\ge1\).
Using the L\'evy--Khintchine representation of \(\phi\),
\[
\psi_\tau(t)
=
\underbrace{b(\tau-\lambda^2)}_{\text{drift}}\,t
+
\int_{(0,\infty)}\bigl(1-e^{-s t}\bigr)\,
\underbrace{[\tau\,\Lambda-\Lambda_\lambda]}_{\text{L\'evy measure}}(ds),
\qquad
\Lambda_\lambda(B):=\Lambda(B/\lambda^2).
\]
Since \(\Gamma_\tau\) is CND on \(\R^d\) for every \(d\ge1\), Theorem~\ref{thm:schoenberg-bernstein} implies that \(\psi_\tau\) is Bernstein. Its drift and its L\'evy measure are therefore nonnegative, hence
\[
b(\tau-\lambda^2)\ge 0
\quad\text{and}\quad
\tau\,\Lambda(B)\ge \Lambda(B/\lambda^2)\ \ \text{for every Borel }B\subset(0,\infty).
\]

\noindent\paragraph{(ii)\(\Rightarrow\)(i).}
Assume
\[
b(\tau-\lambda^2)\ge 0
\quad\text{and}\quad
\tau\,\Lambda(B)\ge \Lambda(B/\lambda^2)\ \ \text{for every Borel }B\subset(0,\infty).
\]
Then the same L\'evy--Khintchine representation shows that
\[
\psi_\tau(t)=\tau\phi(t)-\phi(\lambda^2 t)
\]
is a Bernstein function. Applying Schoenberg's theorem again shows that
\[
\Gamma_\tau(x,y)=\psi_\tau\bigl(\|x-y\|^2\bigr)
\]
is CND on \(\R^d\) for every \(d\ge1\).
\end{proof}

\begin{proposition}[A saturating radial profile cannot induce contraction]\label{prop:no_contraction_saturating_BF}
Adopt the setting of Theorem~\ref{thm:bernstein_reduction}. Fix \(d\ge 1\) and \(\lambda\in(0,1)\), and define the affine map
\[
f(x)=q+\lambda x,
\qquad x\in\R^d.
\]
Let \(\phi:[0,\infty)\to[0,\infty)\) satisfy \(\phi(0)=0\) and define the radial kernel
\[
k(x,y)=\phi(0)-\phi(\|x-y\|^2)=-\phi(\|x-y\|^2).
\]
Assume that \(\phi\) is \emph{saturating}, meaning that \(\lim_{t\to\infty}\phi(t)<\infty\).
Then, for every \(\tau\in(0,1)\), no strict \(\MMD^2\) contraction under \(f\) can hold unless \(\phi\equiv 0\).
More precisely, for any \(\tau\in(0,1)\), define the difference profile \(\psi_\tau\) and the kernel \(\Gamma_\tau\) built from it by
\[
\psi_\tau(t)=\tau\,\phi(t)-\phi(\lambda^2 t),
\qquad
\Gamma_\tau(x,y)=\psi_\tau(\|x-y\|^2).
\]
If \(\phi\not\equiv 0\) and \(\phi\) is saturating, then \(\Gamma_\tau\) is \emph{not} conditionally negative definite on \(\R^d\).
As a consequence, there is no \(\tau\in(0,1)\) such that
\[
\MMD_k^2(f_{\#}\nu,f_{\#}\mu)\ \le\ \tau\,\MMD_k^2(\nu,\mu)
\]
holds for all \(\mu,\nu\in\mathcal P(\R^d)\).
\end{proposition}

\begin{proof}
Fix \(\tau\in(0,1)\). If \(\Gamma_\tau\) were conditionally negative definite on \(\R^d\), then for every choice of two points
\(x_1,x_2\in\R^d\) and weights \((w_1,w_2)=(1,-1)\) with \(w_1+w_2=0\), we would have
\[
\sum_{i,j=1}^2 w_i w_j\,\Gamma_\tau(x_i,x_j)\ \le\ 0.
\]
Let \(r=\|x_1-x_2\|\). Since \(\Gamma_\tau(x_1,x_2)=\psi_\tau(r^2)\) and
\(\Gamma_\tau(x_1,x_1)=\Gamma_\tau(x_2,x_2)=\psi_\tau(0)\), a direct expansion gives
\[
\sum_{i,j=1}^2 w_i w_j\,\Gamma_\tau(x_i,x_j)
=
2\psi_\tau(0)-2\psi_\tau(r^2).
\]
Thus CND would imply
\[
\psi_\tau(r^2)\ \ge\ \psi_\tau(0)\qquad\text{for all }r\ge 0.
\]
Since \(\phi(0)=0\), we have \(\psi_\tau(0)=\tau\phi(0)-\phi(0)=0\), hence
\[
\psi_\tau(t)\ \ge\ 0\qquad\text{for all }t\ge 0,
\]
that is,
\[
\tau\,\phi(t)\ \ge\ \phi(\lambda^2 t)\qquad\text{for all }t\ge 0.
\]
Now assume \(\phi\) is saturating and \(\phi\not\equiv 0\). Then the limit
\[
L:=\lim_{t\to\infty}\phi(t)
\]
exists and satisfies \(L\in(0,\infty)\). Letting \(t\to\infty\) in \(\tau\phi(t)\ge \phi(\lambda^2 t)\) yields
\[
\tau L\ \ge\ L.
\]
Since \(L>0\), this forces \(\tau\ge 1\), contradicting \(\tau\in(0,1)\). Therefore, if \(\tau\in(0,1)\) and
\(\phi\) is saturating with \(\phi\not\equiv 0\), the kernel \(\Gamma_\tau\) cannot be CND on \(\R^d\).

Finally, by Proposition~\ref{prop:radial_contraction}, strict \(\MMD^2\) contraction under the map \(f(x)=q+\lambda x\)
with factor \(\tau\in(0,1)\) would require \(\Gamma_\tau\) to be CND on \(\R^d\). This is impossible when \(\phi\) is saturating and \(\phi\not\equiv 0\),
so no such strict contraction can hold unless \(\phi\equiv 0\).
\end{proof}

\subsection{Results on measures and the measure domination condition}
\label{app:measure_domination}
\paragraph{Preamble.} This subsection studies the condition
\[
\tau\,\Lambda(B)\ \ge\ \Lambda(B/\rho)\qquad\text{for all Borel sets }B\subset(0,\infty),
\]
with $\rho=\lambda^2$. Lemma~\ref{lem:tau-domination} reduces it to two pointwise inequalities, one on the density $m(s)$ and one on the mass $\Lambda(\{x\})$ at every point $x>0$. Lemma~\ref{lem:countable-atoms} shows that a L\'evy measure has at most countably many atoms. The remaining results describe the atoms of a measure that satisfies this condition. Lemma~\ref{lem:finite-atoms} shows that a measure with at least one but finitely many atoms cannot satisfy the condition, so a measure that satisfies it has either no atoms or infinitely many. Lemma~\ref{lem:single_chain_example} shows that a single geometric chain can occur, and Lemma~\ref{lem:chain_representative} partitions $(0,\infty)$ into disjoint geometric chains $\{\rho^n r : n\in\mathbb{Z}\}$, indexed by their representative $r\in(\rho,1]$. Proposition~\ref{prop:bernstein-levy} combines these results. When $\tau\le\rho$, $\Lambda$ has no atoms, and when $\tau>\rho$, atoms can exist but the set of atoms must be a union of disjoint tails of geometric chains.

\begin{lemma}[Atoms of a L\'evy measure are countable]\label{lem:countable-atoms}
Let $\Lambda$ be a L\'evy measure on $(0,\infty)$, that is,
\[
\int_{(0,\infty)}\min\{1,s\}\,\Lambda(ds)<\infty.
\]
Then the set of atoms $\{x>0:\Lambda(\{x\})>0\}$ is at most countable.
\end{lemma}

\begin{proof}
For each \(n\in\mathbb{N}\) let \(E_n=(1/n,\infty)\).
By the L\'evy condition,
\[
\int_{(1,\infty)} 1\,\Lambda(ds) \;=\; \int_{(1,\infty)} \min\{1,s\}\,\Lambda(ds) \;<\;\infty,
\]
so \(\Lambda((1,\infty))<\infty\). Moreover,
\[
\int_{(1/n,1]} 1\,\Lambda(ds)
\;\le\;
n\!\int_{(1/n,1]} s\,\Lambda(ds)
\;\le\;
n\!\int_{(0,1]} s\,\Lambda(ds)
\;\le\;
n\!\int_{(0,\infty)} \min\{1,s\}\,\Lambda(ds)
\;<\;\infty,
\]
hence \(\Lambda((1/n,1])<\infty\). Therefore \(\Lambda(E_n)<\infty\) for every \(n\).

Fix \(n\). We show that the set of atoms in \(E_n\) is countable by grouping atoms according to a lower bound on their mass.
For each \(\ell\in\mathbb{N}\), define
\[
A_{n,\ell} \;:=\; \{\,x\in E_n : \Lambda(\{x\})\ge 1/\ell\,\}.
\]
Each \(A_{n,\ell}\) contains the atoms in \(E_n\) whose mass is at least \(1/\ell\).

We claim that \(A_{n,\ell}\) is finite for every \(\ell\).
Indeed, if \(A_{n,\ell}\) had infinitely many points, then for any \(M\in\mathbb{N}\) we could choose distinct
\(x_1,\dots,x_M\in A_{n,\ell}\). Since the singletons \(\{x_i\}\) are disjoint and contained in \(E_n\),
\[
\Lambda(E_n)\ \ge\ \sum_{i=1}^M \Lambda(\{x_i\})\ \ge\ M\cdot \tfrac{1}{\ell}.
\]
Since \(M\) is arbitrary, letting \(M\to\infty\) forces \(\Lambda(E_n)=\infty\), contradicting \(\Lambda(E_n)<\infty\).
Hence each \(A_{n,\ell}\) is finite.

Now let \(x\in E_n\) be an atom with \(\Lambda(\{x\})=\alpha>0\). Choose \(\ell>\tfrac{1}{\alpha}\).
Then \(\Lambda(\{x\})\ge 1/\ell\), so \(x\in A_{n,\ell}\). Therefore every atom in \(E_n\) belongs to some \(A_{n,\ell}\), and
\[
\{x\in E_n:\Lambda(\{x\})>0\}\;=\;\bigcup_{\ell=1}^\infty A_{n,\ell}.
\]
Since this is a countable union of finite sets, the set of atoms in \(E_n\) is countable. Finally,
\[
\{x>0:\Lambda(\{x\})>0\}
\;=\;
\bigcup_{n=1}^{\infty} \{x\in E_n:\Lambda(\{x\})>0\}
\]
is a countable union of countable sets, hence at most countable.

\end{proof}

\begin{lemma}[The measure domination condition reduces to an inequality on the density and one on the atoms]\label{lem:tau-domination}
Let $\Lambda$ be a nonnegative Borel measure on $(0,\infty)$,
\emph{locally finite} in the sense that $\Lambda([a,b])<\infty$ for all $0<a<b$, 
admitting the decomposition
\[
\Lambda
\;=\;
\underbrace{m(s)\,ds}_{\text{absolutely continuous part}}
\;+\;
\underbrace{\sum_{j\ge1}c_{j}\,\delta_{s_{j}}}_{\text{atomic part}}
\quad(\text{no singular continuous part}),
\]
Here $m(s)\ge0$ for almost every $s>0$, and each $c_{j}>0$. Since $\Lambda$ is locally finite, $m$ is locally integrable on $(0,\infty)$, that is, $\int_a^b m(s)\,ds\le\Lambda([a,b])<\infty$ for all $0<a<b$.  

Fix $\lambda\in(0,1]$, set $\rho=\lambda^{2}$, and define, for every Borel set \(B\subset(0,\infty)\),
\[
\Lambda_{\lambda}(B)\;:=\;\Lambda\bigl(B/\rho\bigr),
\qquad
B/\rho=\{\,s>0:\rho s\in B\,\}.
\]

Let $\tau\in(0,1)$. Then
\[
\boxed{%
\begin{gathered}
(i)\quad\tau\,\Lambda(B)\;\ge\;\Lambda_{\lambda}(B)
    \quad\text{for every Borel set }B\subset(0,\infty),\\[6pt]
\text{if and only if}\\[6pt]
(ii)\quad
\begin{cases}
  (1)\quad
      \tau\,\rho\,m(s)\;\ge\;m\bigl(s/\rho\bigr)
      &\text{for almost every }s>0,\\[4pt]
  (2)\quad
      \tau\,\Lambda(\{x\})\;\ge\;\Lambda\bigl(\{x/\rho\}\bigr)
      &\text{for every }x>0.
\end{cases}
\end{gathered}
}
\]
\end{lemma}
\begin{proof}
\noindent\paragraph{Necessity (i)\(\;\Rightarrow\;\)(ii).}

\paragraph{1. The inequality on the atoms.}
For any $x>0$, take $B=\{x\}$. Since \(\{x\}/\rho=\{x/\rho\}\),
\[
\Lambda_\lambda(\{x\})=\Lambda(\{x/\rho\}).
\]
Thus the hypothesis $\tau\,\Lambda(B)\ge \Lambda_\lambda(B)$ gives
\[
\tau\,\Lambda(\{x\})\;\ge\;\Lambda(\{x/\rho\})\qquad\text{for all }x>0.
\]

\paragraph{2. The inequality on the density.}
Let \(S=\{x>0:\Lambda(\{x\})>0\}=\{s_j:j\ge 1\}\) be the set of atoms, which is countable. Take a Borel set \(B\subset(0,\infty)\) that contains no point of \(S\) and no point of \(\rho S=\{\rho x:x\in S\}\). Then neither \(B\) nor \(B/\rho\) contains an atom, so only the absolutely continuous part contributes, and we have
\[
\Lambda(B) = \int_B m(s)\,ds,
\qquad
\Lambda_\lambda(B) = \Lambda(B/\rho)
= \int_{B/\rho} m(s)\,ds.
\]

Hence the assumption gives
\[
\tau\!\int_B m(s)\,ds
\;\ge\;
\int_{B/\rho}m(s)\,ds.
\]
Perform the change of variables $u=\rho s$, so $s=u/\rho$ and $ds=\tfrac1\rho\,du$.  Then
\[
\int_{B/\rho}m(s)\,ds
=\int_{\{s:\rho s\in B\}}m(s)\,ds
=\frac1\rho\!\int_{B}m(u/\rho)\,du.
\]
Thus for every Borel set \(B\) that contains no point of \(S\cup\rho S\),
\[
\tau\!\int_B m(s)\,ds
\;\ge\;
\frac1\rho\!\int_B m(s/\rho)\,ds.
\]
Finally, choose $B$ to be a small interval around a point $s>0$, minus the countably many points of \(S\cup\rho S\). A countable set has length zero, so this changes neither $|B|$ nor the integrals of $m$. Divide both sides by $|B|$, and let the interval shrink to $\{s\}$. By the Lebesgue differentiation theorem, in the limit one obtains
\[
\tau\,m(s)
\;\ge\;
\frac1\rho\,m(s/\rho)
\quad\text{for almost every }s>0,
\]
or equivalently
\[
\tau\,\rho\,m(s)\;\ge\;m(s/\rho).
\]

\medskip

\noindent\textbf{Sufficiency (ii)\(\;\Rightarrow\;\)(i).}  
Assume
\[
\tau\,\rho\,m(s)\;\ge\;m(s/\rho)
\quad\text{almost everywhere},
\qquad
\tau\,\Lambda(\{x\})\;\ge\;\Lambda(\{x/\rho\})
\quad\forall\,x>0.
\]
Set
\[
\Delta \Lambda \;=\;\tau\,\Lambda \;-\;\Lambda_\lambda
   \;=\;
   \bigl[\tau\,m(s)-\tfrac1\rho\,m(s/\rho)\bigr]\,ds
   \;+\;
   \sum_{\substack{x>0:\\ \Lambda(\{x\})+\Lambda(\{x/\rho\})>0}}\bigl[\tau\,\Lambda(\{x\})-\Lambda(\{x/\rho\})\bigr]\,\delta_x.
\]
Here the sum runs over the countably many points \(x\) such that \(x\) or \(x/\rho\) is an atom of \(\Lambda\), and each such \(x\) carries the mass \(\tau\,\Lambda(\{x\})-\Lambda(\{x/\rho\})\).

\noindent\textbf{1. Absolutely continuous part.}  
By assumption (ii.1), 
$\tau m(s) - \tfrac{1}{\rho} m(s/\rho) \ge 0$ almost everywhere, 
so the absolutely continuous part of $\Delta\Lambda$ is nonnegative.

\medskip

\noindent\textbf{2. Atomic part.}  
By assumption (ii.2), each mass 
$\tau\,\Lambda(\{x\})-\Lambda(\{x/\rho\})$ is nonnegative, 
so all atoms of $\Delta\Lambda$ are nonnegative.

\medskip

Hence \(\Delta \Lambda\) is a nonnegative measure and \(\tau\,\Lambda=\Lambda_\lambda+\Delta \Lambda\ge\Lambda_\lambda\).
\end{proof}

\begin{lemma}[A measure with at least one but finitely many atoms cannot satisfy the measure domination condition]\label{lem:finite-atoms}
Fix \(\lambda\in(0,1)\) and \(\tau\in(0,1)\), and let \(\rho=\lambda^2\).
Assume that a Borel measure \(\Lambda\) on \((0,\infty)\) satisfies the measure domination condition
\[
\tau\,\Lambda(B)\;\ge\;\Lambda_{\lambda}(B)
\quad\text{for every Borel set }B\subset(0,\infty),
\qquad
\Lambda_{\lambda}(B):=\Lambda(B/\rho).
\]
Assume in addition that \(\Lambda\) has finitely many atoms, that is, \(\Lambda = m(s)\,ds+\sum_{j=1}^{N}c_{j}\,\delta_{s_{j}}\) with \(m(s)\ge 0\) and \(c_j\ge 0\).
Then \(\Lambda\) has no atoms.
\end{lemma}

\begin{proof}
Suppose for contradiction that \(\Lambda\) has an atom at some \(s^\star>0\), and let \(c^\star:=\Lambda(\{s^\star\})>0\).
To use the measure domination condition, we choose the singleton \(B=\{\rho s^\star\}\), since \(B/\rho=\{s^\star\}\), so the right-hand side is exactly the mass of the atom at \(s^\star\). Indeed, by definition of \(\Lambda_\lambda\),
\[
\Lambda_\lambda(\{\rho s^\star\})
=\Lambda(\{s^\star\})
=c^\star.
\]
Applying the measure domination condition with \(B=\{\rho s^\star\}\) gives
\[
\tau\,\Lambda(\{\rho s^\star\}) \;\ge\; \Lambda_\lambda(\{\rho s^\star\}) \;=\; c^\star,
\]
so \(\Lambda(\{\rho s^\star\})>0\). Thus, for the measure domination condition to hold, \(\rho s^\star\) must also be an atom of \(\Lambda\).
Iterating, we obtain
\[
\Lambda(\{\rho^n s^\star\})>0
\qquad\text{for every }n\ge 1.
\]
The locations \(\rho^n s^\star\) are all distinct because \(\rho\in(0,1)\), so \(\Lambda\) has infinitely many atoms.
This contradicts the assumption that \(\Lambda\) has finitely many atoms.
Hence \(\Lambda\) has no atoms.
\end{proof}

\begin{lemma}[Atoms on a single geometric chain satisfy the measure domination condition]\label{lem:single_chain_example}
Fix \(\lambda\in(0,1)\), set \(\rho=\lambda^2\), and let \(\tau\in(0,1)\).
Fix \(s_0>0\) and \(c>0\), and define the measure
\[
\Lambda\ =\ \sum_{n=0}^\infty a_n\,\delta_{s_n},
\qquad
s_n=\rho^n s_0,
\qquad
a_n=c\,\tau^{-n}.
\]
Define, for every Borel set \(B\subset(0,\infty)\),
\[
\Lambda_{\lambda}(B)\;:=\;\Lambda\bigl(B/\rho\bigr),
\qquad
B/\rho=\{\,s>0:\rho s\in B\,\}.
\]
Then the measure domination condition
\[
\tau\,\Lambda(B)\ \ge\ \Lambda_{\lambda}(B)\qquad\text{for every Borel set }B\subset(0,\infty)
\]
holds.

Moreover, \(\Lambda\) is a L\'evy measure if and only if \(\tau>\rho\). In particular, when \(\tau>\rho\), a single geometric chain gives an example satisfying the measure domination condition.
\end{lemma}

\begin{proof}
\noindent\paragraph{Step 1: Verification of the measure domination condition.}

We first compute \(\Lambda_\lambda\) on the singletons that support the chain. For \(n\ge 1\), the previous point of the chain is \(s_{n-1}=s_n/\rho\), so
\begin{align*}
\Lambda_\lambda(\{s_n\})
&= \Lambda(\{s_n\}/\rho)
= \Lambda(\{s_n/\rho\})
= \Lambda(\{s_{n-1}\})
= a_{n-1}
= \tau\,a_n
= \tau\,\Lambda(\{s_n\}).
\end{align*}
For \(n=0\), we instead have
\begin{align*}
\Lambda_\lambda(\{s_0\})
&= \Lambda(\{s_0\}/\rho)
= \Lambda(\{s_0/\rho\})
= 0,
\end{align*}
since \(s_0/\rho\) does not belong to the support of \(\Lambda\), whereas \(\tau\,\Lambda(\{s_0\})=\tau a_0>0\).

We now pass from these singleton identities to an arbitrary Borel set \(B\subset(0,\infty)\). Since \(\Lambda\) is a sum of atoms,
\begin{align*}
\Lambda_\lambda(B)
&= \Lambda(B/\rho)
\qquad(\text{definition of }\Lambda_\lambda)\\
&= \sum_{n=0}^\infty a_n\,\mathbf{1}_{B/\rho}(s_n)
\qquad\bigl(\Lambda=\sum_{n=0}^\infty a_n\,\delta_{s_n}\bigr)\\
&= \sum_{n=0}^\infty a_n\,\mathbf{1}_B(\rho s_n)
\qquad\bigl(\mathbf{1}_{B/\rho}(s_n)=\mathbf{1}_B(\rho s_n)\bigr)\\
&= \sum_{n=0}^\infty a_n\,\mathbf{1}_B(s_{n+1})
\qquad(\rho s_n=s_{n+1})\\
&= \sum_{j=1}^\infty a_{j-1}\,\mathbf{1}_B(s_j)
\qquad(\text{reindex } j=n+1)\\
&= \sum_{j=1}^\infty \tau\,a_j\,\mathbf{1}_B(s_j)
\qquad(a_{j-1}=\tau a_j)\\
&\le \tau\sum_{j=0}^\infty a_j\,\mathbf{1}_B(s_j)
\qquad(\text{add the nonnegative }j=0\text{ term})\\
&= \tau\,\Lambda(B).
\end{align*}
Thus the measure domination condition holds for all Borel \(B\subset(0,\infty)\). Moreover, equality holds whenever \(s_0\notin B\), since the only extra term added in the inequality above is the \(j=0\) contribution.

\noindent\paragraph{Step 2: Verification of the L\'evy condition.}

By definition, \(\Lambda\) is a L\'evy measure when it satisfies the L\'evy integrability condition
\[
\int_{(0,\infty)}\min\{1,s\}\,\Lambda(ds)<\infty.
\]
Since \(\Lambda\) puts the mass \(a_n\) at \(s_n\), this integral equals
\[
\sum_{n\ge0} a_n\min\{1,s_n\}.
\]
Since \(\rho<1\), the atoms \(s_n=\rho^n s_0\) decrease to \(0\), so there is an index \(K\) after which they are all at most \(1\), with \(K=0\) when \(s_0\le 1\). For \(n\ge K\), \(\min\{1,s_n\}=s_n\). The sum above therefore differs from
\[
\sum_{n\ge0} a_n s_n
= s_0 c\,\sum_{n\ge0}\Bigl(\frac{\rho}{\tau}\Bigr)^n,
\qquad\text{using } a_n s_n=c\,\tau^{-n}\rho^n s_0,
\]
only in its first \(K\) terms, each of them finite, so one is finite exactly when the other is.
This geometric series converges if and only if \(\rho/\tau<1\), that is, if and only if \(\tau>\rho\). Therefore \(\Lambda\) is a L\'evy measure if and only if \(\tau>\rho\).
\end{proof}

\begin{lemma}[Every \(x>0\) admits a unique decomposition \(x=\rho^n r\) with \(r\in(\rho,1\rbrack\)]\label{lem:chain_representative}
Fix \(\rho\in(0,1)\). Then every \(x>0\) admits a unique decomposition
\[
x=\rho^n r,
\qquad
n\in\mathbb Z,\quad r\in(\rho,1],
\]
given by
\[
n=\left\lfloor \frac{\log x}{\log \rho}\right\rfloor,
\qquad
r=\rho^{-n}x.
\]
Equivalently, each geometric chain \(\{\rho^n x:\ n\in\mathbb Z\}\) meets the interval \((\rho,1]\) in exactly one point. We write \(n(x)\) and \(r(x)\) for these \(n\) and \(r\), and call \(r(x)\) the \emph{chain representative} of \(x\) and \(n(x)\) its \emph{chain index}.
\end{lemma}

\begin{proof}
Fix \(x>0\). For an integer \(n\), set \(r=\rho^{-n}x\), so that \(x=\rho^n r\). Then \(r\in(\rho,1]\) means
\[
\rho\ <\ \rho^{-n}x\ \le\ 1,
\]
and multiplying by \(\rho^n>0\), this is equivalent to
\[
\rho^{n+1}\ <\ x\ \le\ \rho^{n},
\]
that is, to \(x\) lying in the interval \((\rho^{n+1},\rho^n]\). For \(n\in\mathbb Z\), these intervals cover \((0,\infty)\) without overlapping,
\[
\cdots,\qquad(\rho^{2},\rho],\qquad(\rho,1],\qquad(1,\rho^{-1}],\qquad(\rho^{-1},\rho^{-2}],\qquad\cdots
\]
To find the one that contains \(x\), we take logarithms and divide by \(\log\rho<0\), which reverses the inequalities. The condition becomes
\[
n\ \le\ \frac{\log x}{\log\rho}\ <\ n+1.
\]
Exactly one integer satisfies it, namely \(n=\lfloor \log x/\log\rho\rfloor\). Hence \(x\) has exactly one decomposition \(x=\rho^n r\) with \(n\in\mathbb Z\) and \(r\in(\rho,1]\), given by \(n(x)=\lfloor \log x/\log\rho\rfloor\) and \(r(x)=\rho^{-n(x)}x\).

Finally, multiplying by \(\rho\) moves a point from one interval to the next one on its left, so the chain \(\{\rho^n x:\ n\in\mathbb Z\}\) has exactly one point in each interval, and in particular exactly one in \((\rho,1]\), namely \(r(x)\).
\end{proof}

\begin{proposition}[The measure domination condition rules out atoms of the L\'evy measure or forces them onto infinite geometric chains]\label{prop:bernstein-levy}
Let \(\Lambda\) be a L\'evy measure on \((0,\infty)\), that is,
\[
\int_{(0,\infty)}\min\{1,s\}\,\Lambda(ds)<\infty.
\]
Assume \(\Lambda\) has no singular continuous part and admits the decomposition
\[
\Lambda(ds)=m(s)\,ds+\sum_{j\ge1} c_j\,\delta_{s_j},
\qquad
m(s)\ge 0\ \text{for almost every }s>0,
\quad
c_j\ge 0.
\]
Fix \(\lambda\in(0,1)\), set \(\rho=\lambda^2\), and let \(\tau\in(0,1)\). Define, for every Borel set \(B\subset(0,\infty)\),
\[
\Lambda_{\lambda}(B)\;:=\;\Lambda\bigl(B/\rho\bigr),
\qquad
B/\rho=\{\,s>0:\rho s\in B\,\}.
\]
Then the measure domination condition
\[
\tau\,\Lambda(B)\ \ge\ \Lambda_{\lambda}(B)\quad\text{for every Borel set }B\subset(0,\infty)
\]
holds if and only if
\[
\tau\rho\,m(s)\ \ge\ m(s/\rho)\ \ \text{for almost every }s>0,
\qquad
\tau\,\Lambda(\{x\})\ \ge\ \Lambda(\{x/\rho\})\ \ \text{for every }x>0.
\]

Moreover, two cases occur, depending on whether \(\tau\le\rho\) or \(\tau>\rho\).
\begin{enumerate}
\item If \(\tau\le \rho\) and the measure domination condition holds, then \(c_j=0\) for all \(j\), so \(\Lambda\) has no atoms.

\item If \(\tau>\rho\), then \(\Lambda\) can have atoms and still satisfy the measure domination condition, but its atoms must then follow the chains of Lemma~\ref{lem:chain_representative}. Write every \(x>0\) as
\[
x=\rho^n r,
\qquad
n\in\mathbb Z,\quad r\in(\rho,1],
\]
as in Lemma~\ref{lem:chain_representative}, so that \(x\) lies on the chain \(\{\rho^n r:\ n\in\mathbb Z\}\) of its representative \(r\). If the condition holds and \(\rho^n r\) is an atom, then every point of that chain closer to \(0\), namely \(\rho^{n+1}r,\rho^{n+2}r,\dots\), is also an atom. Hence, on a chain that carries an atom, the atoms are exactly the points \(\rho^n r\) with \(n\ge n_r\), where \(n_r\) is the index of the first atom, or \(n_r=-\infty\) if every point of the chain is an atom. The set of atoms of \(\Lambda\) is therefore a union of disjoint tails of geometric chains, of the form
\[
\{\rho^n r:\ n\ge n_r\},
\qquad r\in(\rho,1],
\]
one for each chain that carries an atom. In particular, \(\Lambda\) has either no atoms or infinitely many.

Such atoms can occur on the tail of a single geometric chain, by Lemma~\ref{lem:single_chain_example}, and also on the tails of finitely or countably many geometric chains at once, provided \(\int_{(0,\infty)}\min\{1,s\}\,\Lambda(ds)<\infty\).
\end{enumerate}
\end{proposition}

\begin{proof}
\noindent\paragraph{From the measure domination condition to the inequalities on the density and on the atoms.}
By assumption, \(\Lambda\) has no singular continuous part. Hence Lemma~\ref{lem:tau-domination} gives the equivalence
\[
\begin{gathered}
\tau\,\Lambda(B)\ \ge\ \Lambda_\lambda(B)\quad\text{for every Borel }B\subset(0,\infty)
\\
\Longleftrightarrow
\\
\begin{cases}
\tau\rho\,m(s)\ge m(s/\rho)\ \text{for almost every }s>0,\\
\tau\,\Lambda(\{x\})\ge \Lambda(\{x/\rho\})\ \ \forall x>0.
\end{cases}
\end{gathered}
\]
In addition, since \(\min\{1,s\}=s\) for \(s\le 1\) and \(\min\{1,s\}=1\) for \(s>1\), the decomposition of \(\Lambda\) splits the L\'evy condition into two parts,
\[
\int_{(0,\infty)}\min\{1,s\}\,\Lambda(ds)
=
\Bigl(\int_0^1 s\,m(s)\,ds+\sum_{j:\,s_j\le 1} s_j c_j\Bigr)
+
\Bigl(\int_1^\infty m(s)\,ds+\sum_{j:\,s_j>1} c_j\Bigr)
<\infty.
\]
Both parts are nonnegative, so each of them is finite,
\[
\int_0^1 s\,m(s)\,ds+\sum_{j:\,s_j\le 1} s_j c_j<\infty,
\qquad
\int_1^\infty m(s)\,ds+\sum_{j:\,s_j>1} c_j<\infty.
\]

We now turn to the two regimes for \(\tau\).

\medskip
\noindent\textbf{(1) If \(\tau\le\rho\), then \(\Lambda\) has no atoms.}
Assume for contradiction that there is an atom at \(s_0>0\) with
\(\Lambda(\{s_0\})=:c_0>0\).
To use the measure domination condition, as in Lemma~\ref{lem:finite-atoms}, we evaluate it on the singleton \(B=\{\rho s_0\}\).
\[
\tau\,\Lambda(\{\rho s_0\}) \;\ge\; \Lambda_\lambda(\{\rho s_0\}) \;=\; \Lambda(\{s_0\}) \;=\; c_0.
\]
Since the absolutely continuous part gives no mass to singletons, only a possible atom at \(\rho s_0\) can contribute there. Let
\[
c_1\ :=\ \Lambda(\{\rho s_0\}),
\]
which is the mass carried by \(\Lambda\) at \(\rho s_0\) and is \(0\) if there is no atom at that point. The inequality above therefore gives
\[
c_1 \;\ge\; c_0/\tau \;>\; 0.
\]
There are two cases.

\smallskip
\noindent\textbf{Case 1 (no atom at \(\rho s_0\)).}
Then \(\Lambda(\{\rho s_0\})=0\), contradicting \(c_1\ge c_0/\tau>0\).
Hence this case cannot occur.

\smallskip
\noindent\textbf{Case 2 (an atom at \(\rho s_0\)).}
Then \(c_1=\Lambda(\{\rho s_0\})>0\). Testing again on \(B=\{\rho^2 s_0\}\) yields
\[
\tau\,\Lambda(\{\rho^2 s_0\})\;\ge\;\Lambda(\{\rho s_0\})=c_1\quad\Rightarrow\quad
\Lambda(\{\rho^2 s_0\})=:c_2\;\ge\;\tau^{-1}c_1\;\ge\;\tau^{-2}c_0>0.
\]
Iterating, for all \(n\ge1\),
\[
c_n\ :=\ \Lambda(\{\rho^n s_0\})\ \ge\ \tau^{-n}c_0\ >\ 0.
\]
Since \(\rho\in(0,1)\), we have \(\rho^n s_0\le 1\) for all sufficiently large \(n\).
Therefore the contribution of these atoms to \(\int_{(0,1]} s\,\Lambda(ds)\) satisfies
\[
\int_{(0,1]} s\,\Lambda(ds)
\;\ge\; \sum_{n:\,\rho^n s_0\le 1} (\rho^n s_0)\,c_n
\;\ge\; s_0 c_0 \sum_{n:\,\rho^n s_0\le 1}\Bigl(\frac{\rho}{\tau}\Bigr)^n.
\]
If \(\tau\le\rho\), then \(\rho/\tau\ge 1\) and the series on the right diverges,
contradicting the L\'evy integrability \(\int_{(0,1]} s\,\Lambda(ds)<\infty\).
Thus \(\Lambda\) has no atoms.

\medskip
\noindent\textbf{(2) If \(\tau>\rho\), then the set of atoms is a union of disjoint tails of geometric chains.}
By Lemma~\ref{lem:chain_representative}, every \(x>0\) admits a unique decomposition
\[
x=\rho^n r,
\qquad
n\in\mathbb Z,\quad r\in(\rho,1].
\]
Fix \(r\in(\rho,1]\), and let
\[
A_r:=\{\,n\in\mathbb Z:\Lambda(\{\rho^n r\})>0\,\}
\]
denote the set of indices along that chain at which \(\Lambda\) has an atom. Then, once a level \(n\) on that chain carries an atom, the next level \(n+1\) must also carry an atom. Indeed, if \(n\in A_r\), then \(\Lambda(\{\rho^n r\})>0\), and the inequality on the atoms, (ii.2) in Lemma~\ref{lem:tau-domination}, applied with \(x=\rho^{n+1}r\), gives
\[
\tau\,\Lambda(\{\rho^{n+1}r\})\ \ge\ \Lambda(\{\rho^n r\})\ >\ 0,
\]
hence \(\Lambda(\{\rho^{n+1}r\})>0\), that is,
\[
n\in A_r\quad\Longrightarrow\quad n+1\in A_r.
\]

Equivalently, for each chain representative \(r\in(\rho,1]\), there is a threshold index \(n_r\in\mathbb Z\cup\{-\infty,+\infty\}\) that describes where the atoms begin on that chain, in the sense that
\[
A_r=\{\,n\in\mathbb Z:\ n\ge n_r\,\}.
\]
Thus, if \(n_r\in\mathbb Z\), then the first atom on the chain represented by \(r\) occurs at level \(n_r\), and all later levels \(n\ge n_r\) also carry atoms. The conventions \(n_r=+\infty\) and \(n_r=-\infty\) correspond respectively to \(A_r=\varnothing\) and \(A_r=\mathbb Z\).

Therefore, for each chain representative \(r\in(\rho,1]\), the atoms on that chain are exactly
\[
\{\rho^n r:\ n\in A_r\}
=
\{\rho^n r:\ n\ge n_r\}.
\]
Since every atom belongs to a unique chain represented by some \(r\in(\rho,1]\), the set of all atoms is obtained by taking the union over all such chains. Hence the set of atoms is a union of disjoint tails of geometric chains, of the form
\[
\{\rho^n r:\ n\ge n_r\},
\qquad r\in(\rho,1].
\]
In particular, \(\Lambda\) has either no atoms or infinitely many, in agreement with Lemma~\ref{lem:finite-atoms}.

Conversely, when \(\tau>\rho\), a single geometric chain can occur by Lemma~\ref{lem:single_chain_example}. More generally, let \(I\) be finite or countably infinite, let \((r_i)_{i\in I}\subset(\rho,1]\) be distinct chain representatives, and for each \(i\in I\) let
\[
\Lambda^{(i)}=\sum_{n=0}^\infty a_{i,n}\,\delta_{\rho^n r_i}
\]
be a chain measure satisfying the measure domination condition. Assume that
\[
\sum_{i\in I}\int_{(0,\infty)} \min\{1,s\}\,\Lambda^{(i)}(ds)<\infty.
\]
In particular, every term of this sum is finite, so each \(\Lambda^{(i)}\) is a L\'evy measure. Then
\[
\Lambda:=\sum_{i\in I}\Lambda^{(i)}
\]
is a L\'evy measure whose set of atoms is a union of disjoint tails of these chains, and it still satisfies the measure domination condition. Indeed,
\[
\int_{(0,\infty)} \min\{1,s\}\,\Lambda(ds)
=
\sum_{i\in I}\int_{(0,\infty)} \min\{1,s\}\,\Lambda^{(i)}(ds)
<\infty,
\]
so \(\Lambda\) is a L\'evy measure. Moreover, for every Borel set \(B\subset(0,\infty)\),
\[
\tau\,\Lambda(B)
=
\sum_{i\in I}\tau\,\Lambda^{(i)}(B)
\;\ge\;
\sum_{i\in I}\Lambda^{(i)}_\lambda(B)
=
\Lambda_\lambda(B),
\]
where the last identity follows from
\[
\Lambda_\lambda(B)=\Lambda(B/\rho)=\sum_{i\in I}\Lambda^{(i)}(B/\rho)=\sum_{i\in I}\Lambda^{(i)}_\lambda(B).
\]
Thus, when \(\tau>\rho\), finitely or countably many disjoint geometric chains can indeed occur, provided the resulting measure is still a L\'evy measure (that is, \(\int_{(0,\infty)}\min\{1,s\}\,\Lambda(ds)<\infty\)).
\end{proof}

\newpage

\subsection{From Bernstein reduction to kernels inducing contraction}
\label{app:bernstein_reduction_to_contraction}
\begin{theorem}[Contraction and smallest contraction factor across dimensions reduce to a L\'evy--Khintchine criterion]\label{thm:contraction-drift-density-atoms}
Adopt the setting of Theorem~\ref{thm:bernstein_reduction}. For each \(d\ge 1\) and \(q\in\R^d\), consider the affine map
\[
f(x)=q+\lambda x,
\qquad x\in\R^d,
\]
together with the radial kernel
\[
k(u,v)=g(\|u-v\|),
\qquad u,v\in\R^d.
\]
Let \(h(r)=g(0)-g(r)\), and let \(\phi\) be the Bernstein function with \(h(r)=\phi(r^2)\).
Fix \(\lambda\in(0,1)\), \(\tau\in(0,1)\), and define the difference profile \(\psi_\tau\) and the kernel \(\Gamma_\tau\) built from it by
\[
\psi_\tau(t)=\tau\,\phi(t)-\phi(\lambda^2 t),
\qquad
\Gamma_\tau(x,y)=\psi_\tau(\|x-y\|^2).
\]

\begin{mdframed}[linewidth=0.4pt,leftmargin=1.5em,rightmargin=1.5em,innerleftmargin=8pt,innerrightmargin=8pt,innertopmargin=6pt,innerbottommargin=6pt,skipabove=8pt,skipbelow=8pt]
The contraction inequality
\[
\MMD^2_{k}(f\#\nu,f\#\mu)\le \tau\,\MMD^2_{k}(\nu,\mu)
\]
holds for every \(d\ge 1\) and every \(\mu,\nu\in\mathcal P(\R^d)\) if and only if
\[
\Gamma_\tau\ \text{is CND on }\R^d\ \text{for every }d\ge 1.
\]
\end{mdframed}

Write the L\'evy--Khintchine representation of \(\phi\) as
\[
\phi(t)=b\,t+\int_{(0,\infty)}(1-e^{-st})\,\Lambda(ds),\qquad b\ge0.
\]
Assume \(\Lambda\) has no singular continuous part and decompose
\[
\Lambda(ds)=m(s)\,ds+\sum_{j\ge1}c_j\,\delta_{s_j}.
\]

\begin{mdframed}[linewidth=0.4pt,leftmargin=1.5em,rightmargin=1.5em,innerleftmargin=8pt,innerrightmargin=8pt,innertopmargin=6pt,innerbottommargin=6pt,skipabove=8pt,skipbelow=8pt]
Then \(\Gamma_\tau\) is CND on \(\R^d\) for every \(d\ge 1\) if and only if
\begin{enumerate}
\item \(b(\tau-\lambda^2)\ge 0\),
\item \(\tau\,\lambda^2\,m(s)\ge m(s/\lambda^2)\) for almost every \(s>0\), and
\item \(\tau\,\Lambda(\{x\})\ge \Lambda(\{x/\lambda^2\})\) for every \(x>0\).
\end{enumerate}
Assume these conditions hold. If \(\Lambda(\{s_0\})>0\), then \(\Lambda(\{\lambda^{2n}s_0\})\ge \tau^{-n}\Lambda(\{s_0\})>0\) for every \(n\ge 0\). If \(\tau\le\lambda^2\), then \(\Lambda\) has no atoms. If \(\tau>\lambda^2\), the set of atoms of \(\Lambda\) is a union of disjoint tails of geometric chains, so \(\Lambda\) has either no atoms or infinitely many.
\end{mdframed}

To summarize the contraction side, define
\[
\boxed{%
\begin{gathered}
\tau^\star(\lambda) \;:=\; \inf\{\tau>0:\ \Gamma_\tau\ \text{is CND on }\R^d\ \text{for every }d\ge 1\}
\\
=\; \inf\{\tau>0:\ \psi_\tau\ \text{is Bernstein}\}.
\\[4pt]
\textbf{Contraction across dimensions holds}\ \text{if and only if}\ \tau^\star(\lambda)<1.
\\[4pt]
\textbf{Smallest contraction factor across dimensions:}\ 
\tau^\star(\lambda)\ \text{for }\MMD^2_{k},\ \sqrt{\tau^\star(\lambda)}\ \text{for }\MMD_{k}.
\end{gathered}
}
\]

\noindent\emph{Drift cases.}
The drift is \(b=\lim_{t\to\infty}\phi(t)/t\).
If \(b>0\), then necessarily \(\tau\ge\lambda^2\).
If \(b=0\), no inequality on \(\tau\) follows from the drift term.
\end{theorem}
\noindent The factor \(\tau^\star(\lambda)\) asks for contraction in every dimension at once. For \(\lambda=\gamma\), a factor that gives contraction in every dimension gives it in particular for returns in \(\R^d\), so by Theorem~\ref{thm:cpd_contraction} the smallest contraction factor of the main text satisfies \(\tau_k^\star(\mathcal T^\pi)\le\tau^\star(\gamma)\).

\begin{proof}

\noindent\paragraph{Step 1: Reduction to a CND test.}
The first boxed equivalence is exactly Theorem~\ref{thm:bernstein_reduction}.

\noindent\paragraph{Step 2: L\'evy--Khintchine drift and the measure domination condition.}
Write the L\'evy--Khintchine representation
\[
\phi(t)=b\,t+\int_{(0,\infty)}(1-e^{-st})\,\Lambda(ds),\qquad b\ge 0,
\]
with \(\Lambda\) a nonnegative Borel measure. By Theorem~\ref{thm:levy_scaled_difference},
\[
\begin{gathered}
\Gamma_\tau\ \text{is CND on }\R^d\ \text{for every }d\ge 1
\\
\Longleftrightarrow
\\
b(\tau-\lambda^2)\ge 0
\ \text{ and }\ 
\tau\,\Lambda(B)\ge \Lambda(B/\lambda^2)\ \ \forall B\subset(0,\infty)\ \text{Borel}.
\end{gathered}
\]

\noindent\paragraph{Step 3: The inequalities on the density and on the atoms.}
By assumption, \(\Lambda\) has no singular continuous part and decomposes as
\[
\Lambda(ds)=m(s)\,ds+\sum_{j\ge 1}c_j\,\delta_{s_j}.
\]
Applying Lemma~\ref{lem:tau-domination} with \(\rho=\lambda^2\) rewrites the measure domination condition as
\[
\tau\,\Lambda(B)\ge \Lambda(B/\lambda^2)\ \ \forall\,\text{Borel }B
\quad\Longleftrightarrow\quad
\begin{cases}
\tau\lambda^2\,m(s)\ge m(s/\lambda^2) & \text{for almost every }s>0,\\[2pt]
\tau\,\Lambda(\{x\})\ge \Lambda(\{x/\lambda^2\}) & \forall\,x>0,
\end{cases}
\]
which, together with \(b(\tau-\lambda^2)\ge 0\), gives the three conditions of the statement.

\noindent\paragraph{Step 4: Consequences for the atoms.}
Under condition 3 of the statement, if \(\Lambda(\{s_0\})>0\), then
\[
\Lambda(\{\lambda^{2n}s_0\})\ge \tau^{-n}\Lambda(\{s_0\})
\qquad\text{for all }n\ge 0.
\]
Thus, once an atom appears at a point, every smaller point of the same geometric chain must also carry an atom.

Moreover, Proposition~\ref{prop:bernstein-levy} with \(\rho=\lambda^2\) shows that if \(\tau\le \lambda^2\), then
\[
\Lambda(\{x\})=0\qquad\forall x>0.
\]
If instead \(\tau>\lambda^2\), then the set of atoms of \(\Lambda\) must be a union of disjoint tails of geometric chains, of the form
\[
\{\lambda^{2n}r:\ n\ge n_r\},
\qquad r\in(\lambda^2,1],
\]
and in particular \(\Lambda\) has either no atoms or infinitely many. Conversely, such tails can indeed occur, by Proposition~\ref{prop:bernstein-levy}.

\noindent\paragraph{Step 5: Smallest contraction factor.}
By Theorem~\ref{thm:bernstein_reduction},
\[
\Gamma_\tau\ \text{is CND on }\R^d\ \text{for every }d\ge 1
\quad\Longleftrightarrow\quad
\psi_\tau\ \text{is Bernstein}.
\]
Therefore
\[
\tau^\star(\lambda):=\inf\{\tau>0:\ \Gamma_\tau\ \text{is CND on }\R^d\ \text{for every }d\ge 1\}
=\inf\{\tau>0:\ \psi_\tau\ \text{is Bernstein}\}.
\]
Combining this with Step~1, contraction across dimensions holds if and only if \(\tau^\star(\lambda)<1\), and then \(\tau^\star(\lambda)\) is the smallest contraction factor across dimensions in \(\MMD^2_{k}\),
and \(\sqrt{\tau^\star(\lambda)}\) in \(\MMD_{k}\).

\noindent\paragraph{Step 6: Drift cases.}
The drift condition \(b(\tau-\lambda^2)\ge 0\) of Step~2 gives \(\tau\ge \lambda^2\) if \(b>0\),
and no restriction on \(\tau\) if \(b=0\).

\end{proof}

\subsection{Contraction from complete Bernstein functions}
\label{app:contraction_complete_bernstein}
\begin{theorem}[For complete Bernstein functions, the smallest contraction factor across dimensions reduces to a single supremum]
\label{thm:cbf_contraction_sup}
Adopt the setting of Theorem~\ref{thm:contraction-drift-density-atoms}. For each \(d\ge 1\) and \(q\in\R^d\), consider the affine map
\[
f(x)=q+\lambda x,
\qquad x\in\R^d.
\]
Assume moreover that \(\phi\) is a \emph{complete} Bernstein function with \(\phi(0)=0\). Then \(\phi\) admits a L\'evy--Khintchine representation
\[
\phi(t)=b\,t+\int_{0}^{\infty}\bigl(1-e^{-s t}\bigr)\,m(s)\,ds,
\qquad t\ge 0,
\]
with \(b\ge 0\) and a L\'evy density \(m:(0,\infty)\to[0,\infty)\) such that, by Definition~\ref{def:complete-bernstein},
\[
m\in C^\infty((0,\infty))
\quad\text{and}\quad
m\ \text{is completely monotone}.
\]

Fix \(\lambda\in(0,1)\) and \(\tau>0\), and define the difference profile
\[
\psi_\tau(t)=\tau\,\phi(t)-\phi(\lambda^2 t),
\qquad t\ge 0.
\]
Then \(\psi_\tau\) is a Bernstein function if and only if
\[
b(\tau-\lambda^2)\ge 0
\qquad\text{and}\qquad
\tau\,\lambda^2\,m(s)\ \ge\ m\!\left(\frac{s}{\lambda^2}\right)
\ \ \text{for every }s>0.
\]

If \(m\equiv 0\), then
\[
\tau^\star(\lambda)=\lambda^2\,\mathbf{1}_{\{b>0\}}.
\]
If \(m\not\equiv 0\), then the smallest \(\tau\) for which \(\psi_\tau\) is a Bernstein function is
\[
\boxed{%
\tau^\star(\lambda)
=
\max\!\left\{
\lambda^2\,\mathbf{1}_{\{b>0\}},
\ \ 
\sup_{s>0}\frac{m\!\left(\frac{s}{\lambda^2}\right)}{\lambda^2\,m(s)}
\right\}.}
\]

In particular, for every \(d\ge 1\) and all \(\mu,\nu\in\mathcal P(\R^d)\),
\[
\MMD^2_{k}(f\#\nu,f\#\mu)\le \tau\,\MMD^2_{k}(\nu,\mu)
\]
holds if and only if \(\tau\ge \tau^\star(\lambda)\). Consequently, contraction across dimensions holds if and only if
\[
\tau^\star(\lambda)<1,
\]
in which case \(\tau^\star(\lambda)\) is the smallest contraction factor across dimensions in \(\MMD_k^2\).
\end{theorem}

\begin{proof}
By Theorem~\ref{thm:contraction-drift-density-atoms}, \(\psi_\tau\) is Bernstein if and only if
\[
b(\tau-\lambda^2)\ge 0
\qquad\text{and}\qquad
\tau\,\lambda^2\,m(s)\ge m(s/\lambda^2)
\quad\text{for almost every }s>0.
\]
By Definition~\ref{def:complete-bernstein}, one has \(m\in C^\infty((0,\infty))\). Hence both sides of the inequality are continuous in \(s\), so the inequality, which holds almost everywhere, extends to every \(s>0\). This proves the first claim.

If \(m\equiv 0\), only the drift condition remains, so
\[
\tau^\star(\lambda)=\lambda^2\,\mathbf{1}_{\{b>0\}}.
\]

Assume now that \(m\not\equiv 0\). Since \(m\) is completely monotone and nonzero, it is strictly positive on \((0,\infty)\). Therefore \(\lambda^2 m(s)>0\) for every \(s>0\), so we may divide both sides by \(\lambda^2 m(s)\). Thus
\[
\tau\,\lambda^2\,m(s)\ge m(s/\lambda^2)
\qquad\forall s>0
\]
is equivalent to
\[
\tau\ge \frac{m(s/\lambda^2)}{\lambda^2\,m(s)}
\qquad\forall s>0.
\]
Combining this with \(b(\tau-\lambda^2)\ge 0\), we obtain that \(\psi_\tau\) is Bernstein if and only if
\[
\tau\ge \lambda^2\,\mathbf{1}_{\{b>0\}}
\qquad\text{and}\qquad
\tau\ge \frac{m(s/\lambda^2)}{\lambda^2\,m(s)}
\quad\forall s>0.
\]
Therefore
\[
\tau^\star(\lambda)
=
\max\!\left\{
\lambda^2\,\mathbf{1}_{\{b>0\}},
\ \ 
\sup_{s>0}\frac{m(s/\lambda^2)}{\lambda^2\,m(s)}
\right\}.
\]

The final equivalence with the \(\MMD_k^2\) contraction statement follows directly from Theorem~\ref{thm:contraction-drift-density-atoms}.
\end{proof}

\begin{corollaryT}[Limit at $0^+$ and consequences for the metric property and contraction]\label{corr:m0_limit_metric}
Adopt the setting of Theorem~\ref{thm:cbf_contraction_sup}. Assume there exists \(\tau<1\) such that
\[
\psi_\tau(t)=\tau\phi(t)-\phi(\lambda^2 t)
\]
is a Bernstein function. Since \(m\) is completely monotone, it is nonincreasing, so its limit at \(0^+\),
\[
m(0^+):=\lim_{s\to 0^+} m(s)\in[0,\infty],
\]
exists. Then \(m(0^+)\) is either \(0\) or \(\infty\).
\begin{enumerate}
\item If \(m(0^+)=0\), then \(m\equiv 0\), so \(\phi(t)=b\,t\) and \(k(x,y)=g(0)-b\|x-y\|^2\). In this case, \(\MMD_k\) is not a metric by Theorem~\ref{thm:bernstein_mmd_metric}. By Theorem~\ref{thm:cbf_contraction_sup}, the smallest contraction factor across dimensions is \(\tau^\star(\lambda)=\lambda^2\) if \(b>0\) and \(\tau^\star(\lambda)=0\) if \(b=0\).
\item If \(\MMD_k\) is a metric, then \(m(0^+)=\infty\).
\end{enumerate}
\end{corollaryT}

\begin{proof}
\noindent\paragraph{Step 1: Existence of $m(0^+)$.}
Since $m$ is completely monotone, it is nonincreasing on $(0,\infty)$.
Hence its boundary value at $0^+$ is given by the supremum, and we may write
\[
m(0^+)=\sup_{s>0} m(s)\ \in\ [0,\infty].
\]
\noindent\paragraph{Step 2: A strict contraction excludes $m(0^+)\in(0,\infty)$.}
Assume for contradiction that $m(0^+)\in(0,\infty)$.
By Theorem~\ref{thm:cbf_contraction_sup}, the assumption that $\psi_\tau$ is a Bernstein function implies that
\[
\tau\,\lambda^2\,m(s)\ \ge\ m\!\left(\frac{s}{\lambda^2}\right)
\qquad\text{for every }s>0.
\]
Letting $s\to 0^+$ and using $m(s)\to m(0^+)$ and $m(s/\lambda^2)\to m(0^+)$ gives
\[
\tau\,\lambda^2\,m(0^+)\ \ge\ m(0^+).
\]
Since $m(0^+)>0$, we may divide by $m(0^+)$ to obtain $\tau\,\lambda^2\ge 1$.
This contradicts $\tau<1$ and $\lambda\in(0,1)$, which together imply $\tau\,\lambda^2<1$.
Hence $m(0^+)\notin(0,\infty)$, and therefore $m(0^+)\in\{0,\infty\}$.

\noindent\paragraph{Step 3: If $m(0^+)=0$ then $m\equiv 0$, MMD is not a metric, and contraction depends only on $b$.}
If $m(0^+)=0$ and $m$ is nonincreasing and nonnegative, then for all $s>0$,
\[
0\ \le\ m(s)\ \le\ m(0^+)=0,
\]
hence $m\equiv 0$ and $\phi(t)=b\,t$.
By Theorem~\ref{thm:bernstein_mmd_metric}, $\MMD_k$ is a metric if and only if $\Lambda\not\equiv 0$,
equivalently $m\not\equiv 0$. Thus $\MMD_k$ is not a metric in this case.
Finally, when $m\equiv 0$, the supremum term in Theorem~\ref{thm:cbf_contraction_sup} vanishes and
$\tau^\star(\lambda)=\lambda^2\,\mathbf{1}_{\{b>0\}}$.
\noindent\paragraph{Step 4: If $\MMD_k$ is a metric then $m(0^+)=\infty$.}
If $\MMD_k$ is a metric, then Theorem~\ref{thm:bernstein_mmd_metric} implies $\Lambda\not\equiv 0$, hence $m\not\equiv 0$.
Step~3 shows that $m(0^+)=0$ would force $m\equiv 0$, a contradiction. Therefore $m(0^+)=\infty$.
\end{proof}

\subsection{Chaining as a general mechanism}\label{app:chaining_general_mechanism}
\paragraph{Preamble.}
Section~\ref{app:measure_domination}, and in particular Lemma~\ref{lem:single_chain_example} and Proposition~\ref{prop:bernstein-levy}, introduced a geometric chaining mechanism. When \(\Lambda\) has atoms and satisfies the measure domination condition, its atoms cannot lie anywhere. By Proposition~\ref{prop:bernstein-levy}, the set of atoms is a union of disjoint tails of geometric chains.

In this subsection, we show that this mechanism is more general. The geometric chain of atoms is the special case of a broader construction in which the base Bernstein function is \(t\mapsto 1-e^{-st}\).

As a first step, we isolate the basic building block already present in Section~\ref{app:measure_domination}, namely a single geometric chain of atoms. This is the simplest instance of the general construction that follows.

\begin{figure}[H]
  \centering
  \includegraphics[width=\linewidth]{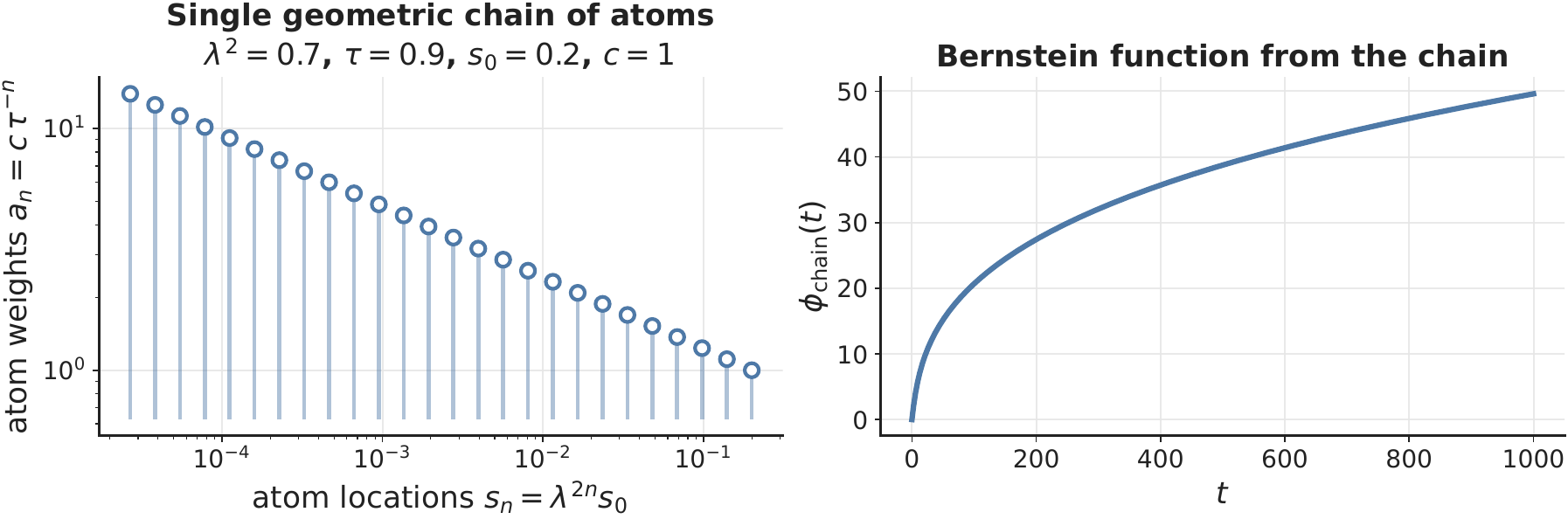}
  \caption{Single geometric chain of atoms. Left, atoms at $s_n=\lambda^{2n}s_0$ with weights $a_n=c\,\tau^{-n}$. The chain is geometric in both location and weight, since $s_{n+1}/s_n=\lambda^2$ and $a_{n+1}/a_n=\tau^{-1}$ are constant. The left panel is truncated after a finite number of atoms for readability. Right, the corresponding Bernstein function $\phi_{\mathrm{chain}}(t)=b t+\sum_{n\ge 0} c\,\tau^{-n}\bigl(1-e^{-s_n t}\bigr)$, evaluated with truncation after $250$ terms.}

  \label{fig:single_chain}
\end{figure}

\begin{proposition}[Single geometric chain of atoms]\label{prop:single_atomic_chain}
Fix \(\lambda\in(0,1)\) and set \(\rho:=\lambda^2\in(0,1)\).
Let \(\tau\in(0,1)\), let \(s_0>0\), and let \(c>0\).
Define the measure
\[
\Lambda(ds)\ :=\ \sum_{n=0}^{\infty} c\,\tau^{-n}\,\delta_{\rho^n s_0}(ds).
\]
If \(\tau>\rho\), then
\[
\int_{(0,\infty)}\min\{1,s\}\,\Lambda(ds)
\le
s_0 c\sum_{n=0}^{\infty}\Bigl(\frac{\rho}{\tau}\Bigr)^n
<\infty,
\]
so \(\Lambda\) is a Lévy measure. In addition, Lemma~\ref{lem:single_chain_example} shows that this single geometric chain satisfies the measure domination condition. Its Bernstein profile is therefore
\[
\phi_{\mathrm{chain}}(t)
=
b t+\sum_{n=0}^{\infty} c\,\tau^{-n}\bigl(1-e^{-\rho^n s_0 t}\bigr),
\qquad t\ge 0,\quad b\ge 0.
\]
Since \(\tau>\rho=\lambda^2\), the drift condition in Theorem~\ref{thm:contraction-drift-density-atoms} is also satisfied. Hence, under the radial kernel generated by \(\phi_{\mathrm{chain}}\), the affine map \(x\mapsto q+\lambda x\) is a \(\tau\)-contraction in \(\MMD^2\).

\end{proposition}
\noindent See Figure~\ref{fig:single_chain} for a visualization of the atoms and the resulting \(\phi_{\mathrm{chain}}\).

\begin{proposition}[Geometric chaining gives identities for the L\'evy measure and the difference profile]
\label{prop:exact_geometric_chaining}
Fix \(\rho\in(0,1)\) and \(\tau\in(0,1)\). Let \(\phi_0\) be a \emph{base} Bernstein function with \(\phi_0(0)=0\), and write its L\'evy--Khintchine representation as
\[
\phi_0(t)
=
b_0\,t+\int_{(0,\infty)}(1-e^{-st})\,\Lambda_0(ds),
\qquad t\ge 0,
\]
where \(b_0\ge 0\) and \(\Lambda_0\) is a L\'evy measure on \((0,\infty)\).

Consider the chained profile
\[
\phi(t):=\sum_{n=0}^\infty \tau^{-n}\phi_0(\rho^n t),
\qquad t\ge 0.
\]

For each \(n\ge 0\), define the rescaled measure
\[
\Lambda_0^{(n)}(B):=\Lambda_0(B/\rho^n),
\qquad
B/\rho^n:=\{s>0:\rho^n s\in B\}.
\]
Define
\[
b:=\sum_{n=0}^\infty \tau^{-n}\rho^n b_0,
\qquad
\Lambda(B):=\sum_{n=0}^\infty \tau^{-n}\Lambda_0(B/\rho^n),
\]
and assume that
\[
b<\infty
\qquad\text{and}\qquad
\int_{(0,\infty)}\min\{1,s\}\,\Lambda(ds)<\infty.
\]
Then \(\phi\) is a Bernstein function and admits the L\'evy--Khintchine representation
\[
\phi(t)=b\,t+\int_{(0,\infty)}(1-e^{-st})\,\Lambda(ds),
\qquad t\ge 0.
\]

Moreover, if we define
\[
\Lambda_\rho(B):=\Lambda(B/\rho),
\]
then
\[
\boxed{\tau\,\Lambda=\Lambda_\rho+\tau\,\Lambda_0}
\]
as nonnegative Borel measures on \((0,\infty)\), and
\[
\boxed{\tau\,\phi(t)-\phi(\rho t)=\tau\,\phi_0(t)}
\]
for all \(t\ge 0\).
\end{proposition}

\begin{proof}
\noindent\paragraph{L\'evy--Khintchine representation of \(\phi\).}
By definition,
\[
\phi(t)=\sum_{n=0}^\infty \tau^{-n}\phi_0(\rho^n t).
\]
Using the L\'evy--Khintchine representation of \(\phi_0\), this becomes
\[
\phi(t)
=
\sum_{n=0}^\infty \tau^{-n}
\left[
\rho^n b_0\,t+\int_{(0,\infty)}(1-e^{-u\rho^n t})\,\Lambda_0(du)
\right].
\]
Since all terms are nonnegative, Tonelli's theorem yields
\[
\phi(t)
=
\left(\sum_{n=0}^\infty \tau^{-n}\rho^n b_0\right)t
+
\int_{(0,\infty)}\sum_{n=0}^\infty \tau^{-n}(1-e^{-u\rho^n t})\,\Lambda_0(du).
\]
By the definition of the rescaled measures \(\Lambda_0^{(n)}(B)=\Lambda_0(B/\rho^n)\), this is exactly
\[
\phi(t)=b\,t+\int_{(0,\infty)}(1-e^{-st})\,\Lambda(ds).
\]

\noindent\paragraph{Exact identities for \(\Lambda\) and \(\phi\).}
By definition,
\[
\Lambda_\rho(B)=\Lambda(B/\rho)
=
\sum_{n=0}^\infty \tau^{-n}\Lambda_0(B/\rho^{n+1}).
\]
Reindexing with \(j=n+1\) gives
\[
\Lambda_\rho(B)
=
\sum_{j=1}^\infty \tau^{-(j-1)}\Lambda_0(B/\rho^j)
=
\tau\sum_{j=1}^\infty \tau^{-j}\Lambda_0(B/\rho^j).
\]
On the other hand,
\[
\tau\,\Lambda(B)
=
\tau\sum_{j=0}^\infty \tau^{-j}\Lambda_0(B/\rho^j)
=
\tau\,\Lambda_0(B)
+
\tau\sum_{j=1}^\infty \tau^{-j}\Lambda_0(B/\rho^j).
\]
Comparing the two expressions, we obtain
\[
\tau\,\Lambda(B)=\Lambda_\rho(B)+\tau\,\Lambda_0(B)
\qquad\text{for every Borel set }B\subset(0,\infty).
\]
That is,
\[
\tau\,\Lambda=\Lambda_\rho+\tau\,\Lambda_0.
\]

Recall that
\[
\phi(t)=\sum_{n=0}^\infty \tau^{-n}\phi_0(\rho^n t).
\]
Therefore
\[
\tau\,\phi(t)
=
\tau\sum_{n=0}^\infty \tau^{-n}\phi_0(\rho^n t)
=
\tau\,\phi_0(t)+\tau\sum_{j=1}^\infty \tau^{-j}\phi_0(\rho^j t).
\]
On the other hand,
\[
\phi(\rho t)
=
\sum_{n=0}^\infty \tau^{-n}\phi_0(\rho^{n+1} t).
\]
Reindexing with \(j=n+1\) gives
\[
\phi(\rho t)
=
\sum_{j=1}^\infty \tau^{-(j-1)}\phi_0(\rho^j t)
=
\tau\sum_{j=1}^\infty \tau^{-j}\phi_0(\rho^j t).
\]
Comparing the two expressions, we obtain
\[
\tau\,\phi(t)-\phi(\rho t)=\tau\,\phi_0(t),
\]
which is exactly the claimed identity for the difference profile.
\end{proof}

\begin{proposition}[When geometric chaining gives a Bernstein function]
\label{prop:geometric_chaining_admissibility}
Fix \(\rho\in(0,1)\) and \(\tau\in(0,1)\). Let \(\phi_0\) be a base Bernstein function with \(\phi_0(0)=0\), and write its L\'evy--Khintchine representation as
\[
\phi_0(t)=b_0\,t+\int_{(0,\infty)}(1-e^{-st})\,\Lambda_0(ds),
\qquad t\ge 0,
\]
where \(b_0\ge 0\) and \(\Lambda_0\) is a L\'evy measure on \((0,\infty)\). Consider the chained profile
\[
\phi(t):=\sum_{n=0}^\infty \tau^{-n}\phi_0(\rho^n t),
\qquad t\ge 0.
\]
For each \(n\ge 0\), define the rescaled measure
\[
\Lambda_0^{(n)}(A):=\Lambda_0(A/\rho^n),
\qquad
A/\rho^n:=\{s>0:\rho^n s\in A\},
\]
and define, as in Proposition~\ref{prop:exact_geometric_chaining},
\[
b:=\sum_{n=0}^\infty \tau^{-n}\rho^n b_0,
\qquad
\Lambda(A):=\sum_{n=0}^\infty \tau^{-n}\Lambda_0(A/\rho^n).
\]
Introduce also
\[
W_{\rho,\tau}(s):=\sum_{n=0}^\infty \tau^{-n}\min\{1,\rho^n s\},
\qquad s>0.
\]
\begin{mdframed}[linewidth=0.4pt,leftmargin=1.5em,rightmargin=1.5em,innerleftmargin=8pt,innerrightmargin=8pt,innertopmargin=6pt,innerbottommargin=6pt,skipabove=8pt,skipbelow=8pt]
Then the conditions
\[
b<\infty
\quad\text{and}\quad
\int_{(0,\infty)}\min\{1,s\}\,\Lambda(ds)<\infty
\]
hold if and only if
\[
\sum_{n=0}^\infty \tau^{-n}\rho^n b_0<\infty
\quad\text{and}\quad
\int_{(0,\infty)}W_{\rho,\tau}(s)\,\Lambda_0(ds)<\infty.
\]
\end{mdframed}
Whenever these equivalent conditions hold, the chained profile is a Bernstein function and admits the representation
\[
\phi(t)=b\,t+\int_{(0,\infty)}(1-e^{-st})\,\Lambda(ds),
\qquad t\ge 0.
\]
In particular, if \(\phi_0\not\equiv 0\), these conditions can hold only if \(\tau>\rho\).
\end{proposition}

\begin{proof}
\noindent\paragraph{Equivalence of the integrability conditions.}
By definition,
\[
b=\sum_{n=0}^\infty \tau^{-n}\rho^n b_0,
\]
so the drift part is well defined if and only if
\[
\sum_{n=0}^\infty \tau^{-n}\rho^n b_0<\infty.
\]

For the measure part, Tonelli's theorem applies because the measures
\(\tau^{-n}\Lambda_0^{(n)}\) are nonnegative and the integrand \(\min\{1,u\}\) is nonnegative. Therefore
\begin{align*}
\int_{(0,\infty)}\min\{1,u\}\,\Lambda(du)
&=
\int_{(0,\infty)}\min\{1,u\}\sum_{n=0}^\infty \tau^{-n}\Lambda_0^{(n)}(du)
\\
&=
\sum_{n=0}^\infty \tau^{-n}\int_{(0,\infty)}\min\{1,u\}\,\Lambda_0^{(n)}(du).
\end{align*}
By the definition of \(\Lambda_0^{(n)}\),
\[
\int_{(0,\infty)}\min\{1,u\}\,\Lambda_0^{(n)}(du)
=
\int_{(0,\infty)}\min\{1,\rho^n s\}\,\Lambda_0(ds).
\]
Hence
\begin{align*}
\int_{(0,\infty)}\min\{1,u\}\,\Lambda(du)
&=
\sum_{n=0}^\infty \tau^{-n}\int_{(0,\infty)}\min\{1,\rho^n s\}\,\Lambda_0(ds)
\\
&=
\int_{(0,\infty)}\sum_{n=0}^\infty \tau^{-n}\min\{1,\rho^n s\}\,\Lambda_0(ds)
\\
&=
\int_{(0,\infty)}W_{\rho,\tau}(s)\,\Lambda_0(ds).
\end{align*}
Thus \(\Lambda\) satisfies the L\'evy integrability condition if and only if
\[
\int_{(0,\infty)}W_{\rho,\tau}(s)\,\Lambda_0(ds)<\infty.
\]
This proves the equivalence.

Under these equivalent conditions, Proposition~\ref{prop:exact_geometric_chaining}
applies to
\[
\phi(t)=\sum_{n=0}^\infty \tau^{-n}\phi_0(\rho^n t)
\]
and yields
\[
\phi(t)=b\,t+\int_{(0,\infty)}(1-e^{-st})\,\Lambda(ds).
\]
In particular, \(\phi\) is then a Bernstein function.

\noindent\paragraph{The regime \(\tau>\rho\) is unavoidable.}
Assume \(\phi_0\not\equiv 0\). If \(b_0>0\), then
\[
b=b_0\sum_{n=0}^\infty \left(\frac{\rho}{\tau}\right)^n,
\]
and therefore \(b<\infty\) implies \(\rho/\tau<1\), that is, \(\tau>\rho\).

Assume now \(b_0=0\). Since \(\phi_0\not\equiv 0\), the measure \(\Lambda_0\) is nonzero, so there exists an interval \([u,v]\subset(0,\infty)\) such that
\[
\int_{[u,v]} s\,\Lambda_0(ds)>0.
\]
Choose \(K\) such that \(\rho^n v\le 1\) for every \(n\ge K\). Then, for every \(n\ge K\) and every \(s\in [u,v]\), one has \(\rho^n s\le \rho^n v\le 1\), hence \(\min\{1,\rho^n s\}=\rho^n s\). Therefore,
\begin{align*}
\int_{(0,\infty)}\min\{1,\rho^n s\}\,\Lambda_0(ds)
&\ge
\int_{[u,v]}\min\{1,\rho^n s\}\,\Lambda_0(ds)
\\
&=
\int_{[u,v]}\rho^n s\,\Lambda_0(ds)
\\
&=
\rho^n\int_{[u,v]} s\,\Lambda_0(ds).
\end{align*}
It follows, by Tonelli's theorem as in the preceding part, that
\begin{align*}
\int_{(0,\infty)}W_{\rho,\tau}(s)\,\Lambda_0(ds)
&=
\int_{(0,\infty)}\sum_{n=0}^\infty \tau^{-n}\min\{1,\rho^n s\}\,\Lambda_0(ds)
\\
&=
\sum_{n=0}^\infty \tau^{-n}\int_{(0,\infty)}\min\{1,\rho^n s\}\,\Lambda_0(ds)
\\
&\ge
\sum_{n=K}^\infty \tau^{-n}\int_{(0,\infty)}\min\{1,\rho^n s\}\,\Lambda_0(ds)
\\
&\ge
\sum_{n=K}^\infty \tau^{-n}\rho^n\int_{[u,v]} s\,\Lambda_0(ds)
\\
&=
\left(\int_{[u,v]} s\,\Lambda_0(ds)\right)\sum_{n=K}^\infty \left(\frac{\rho}{\tau}\right)^n.
\end{align*}
If \(\tau\le\rho\), then \(\rho/\tau\ge 1\), so the series on the right diverges. Therefore
\[
\int_{(0,\infty)}W_{\rho,\tau}(s)\,\Lambda_0(ds)=\infty,
\]
which contradicts the integrability criterion proved above. Thus any such geometric chaining construction must satisfy \(\tau>\rho\).
\end{proof}

\begin{theorem}[Geometric chaining gives contraction with factor \(\tau\)]\label{thm:general_geometric_chaining_factory}
Fix \(\lambda\in(0,1)\) and set \(\rho:=\lambda^2\). Let \(\tau\in(0,1)\), and let \(\phi_0\) be a base Bernstein function with \(\phi_0(0)=0\). Adopt the setting of Proposition~\ref{prop:exact_geometric_chaining}, and assume that \(b<\infty\) and that \(\Lambda\) is a L\'evy measure, as in Proposition~\ref{prop:geometric_chaining_admissibility}. Let
\[
\phi(t):=b\,t+\int_{(0,\infty)}(1-e^{-st})\,\Lambda(ds),
\qquad t\ge 0.
\]
For each \(d\ge 1\) and \(q\in\R^d\), consider the affine map
\[
f(x)=q+\lambda x,
\qquad x\in\R^d,
\]
and define the radial profile and kernel by
\[
h(r):=\phi(r^2),
\qquad
k(x,y):=C-\phi(\|x-y\|^2),
\]
where \(C\in\R\) is arbitrary. Define moreover the difference profile
\[
\psi_\tau(t):=\tau\,\phi(t)-\phi(\rho t),
\qquad t\ge 0.
\]

Then
\[
\boxed{%
\begin{gathered}
\psi_\tau(t)=\tau\,\phi_0(t)
\qquad\text{for every }t\ge 0,
\\[4pt]
\text{hence }\psi_\tau\text{ is a Bernstein function,}
\\[4pt]
\text{and therefore }
\MMD_k^2(f\#\nu,f\#\mu)\le \tau\,\MMD_k^2(\nu,\mu)
\\
\text{for every }d\ge 1\text{ and every }\mu,\nu\in\mathcal P(\R^d).
\end{gathered}
}
\]
In particular, under the kernel generated by the geometrically chained profile \(\phi\), every such affine map \(f\) is a \(\tau\)-contraction in \(\MMD^2\).
\end{theorem}

\begin{proof}
\noindent\paragraph{Step 1: Exact difference identity.}
By Proposition~\ref{prop:exact_geometric_chaining},
\[
\tau\,\phi(t)-\phi(\rho t)=\tau\,\phi_0(t)
\qquad\text{for every }t\ge 0.
\]
That is,
\[
\psi_\tau(t)=\tau\,\phi_0(t).
\]
\noindent\paragraph{Step 2: Bernstein property and contraction.}
Since \(\phi_0\) is a Bernstein function and the class of Bernstein functions is closed under multiplication by a nonnegative scalar~\cite[Corollary~3.7(i)]{schilling2009bernstein}, it follows from Step~1 that \(\psi_\tau=\tau\,\phi_0\) is Bernstein. Applying Theorem~\ref{thm:bernstein_reduction} with the present notation and \(\rho=\lambda^2\), we obtain
\[
\MMD_k^2(f\#\nu,f\#\mu)\le \tau\,\MMD_k^2(\nu,\mu)
\]
for every \(d\ge 1\) and every \(\mu,\nu\in\mathcal P(\R^d)\). This is exactly the claimed \(\tau\)-contraction in \(\MMD^2\).
\end{proof}

\newpage
\section{Explaining known results}
\paragraph{Preamble.} In this section, we show that our framework explains three known results on MMD contraction for radial kernels. As in Appendix~\ref{app:contraction_radial_kernels_via_Bernstein}, we study contraction for a fixed affine map \(f(x)=q+\lambda x\), where \(\lambda\) plays the role of the discount \(\gamma\). These results can then be extended to the setting of distributional RL and its Bellman operators using the results from Appendix~\ref{app:general_contraction}.

The first positive result was established in \citep{nguyen2021distributional} for power-distance kernels. Proposition~\ref{prop:power_distance_contraction} shows that these kernels are induced by a complete Bernstein function and satisfy the criterion of Theorem~\ref{thm:cbf_contraction_sup}. Hence they induce contraction under a fixed map. The corresponding negative result was also established in \citep{nguyen2021distributional}, where it was shown that the RBF kernel does not induce contraction in general. Proposition~\ref{prop:rbf_no_contraction} shows that the RBF kernel is likewise induced by a Bernstein function, and that its non-contraction can be explained through its L\'evy--Khintchine representation together with the obstruction for saturating profiles from Proposition~\ref{prop:no_contraction_saturating_BF}. The second positive result was established in \citep{killingberg2023the} and \citep[Lemma~10]{debes2026distributional} for the multiquadric kernel. Proposition~\ref{prop:gmq_contraction} extends this result to the family of \emph{generalized multiquadric kernels} and shows that these kernels are also induced by a complete Bernstein function.

\begin{proposition}[Contraction under power-distance kernels {\cite{nguyen2021distributional}}]\label{prop:power_distance_contraction}
Let
\[
k_\alpha(x,y) \;=\; -\|x-y\|^\alpha,
\qquad 0<\alpha\le 2,
\]
and let
\[
f(x)=q+\lambda x,
\qquad \lambda\in(0,1).
\]
Then
\[
h(r)=r^\alpha=\phi(r^2),
\qquad
\phi(t)=t^{\alpha/2},
\]
and the smallest contraction factor across dimensions in \(\MMD_{k_\alpha}^2\) is \(\tau^\star(\lambda)=\lambda^\alpha\). For \(\alpha=2\), \(\MMD_{k_2}\) is not a metric on probability measures.
\end{proposition}

\begin{proof}
\leavevmode\par
\noindent\textbf{L\'evy--Khintchine representation.} By \cite[Ch.~15.2, Identity~1]{schilling2009bernstein},
\begin{itemize}
\item for \(0<\alpha<2\) (so \(\alpha/2\in(0,1)\)),
\[
\phi(t)=\frac{\alpha/2}{\Gamma(1-\alpha/2)}\int_0^\infty (1-e^{-st})\,s^{-1-\alpha/2}\,ds,
\]
so the drift is \(b=0\) and the L\'evy measure is absolutely continuous,
\[
\Lambda(ds)=c_\alpha\,s^{-1-\alpha/2}\,ds,
\qquad
c_\alpha=\frac{\alpha/2}{\Gamma(1-\alpha/2)}.
\]
\item for \(\alpha=2\), \(\phi(t)=t\), so the drift is \(b=1\) and \(\Lambda=0\).
\end{itemize}

\noindent\textbf{Bernstein test for the difference profile.}
By Theorem~\ref{thm:bernstein_reduction}, in the present radial setting, contraction in \(\MMD^2\) reduces to checking whether the difference profile
\[
\psi_\tau(t):=\tau\,\phi(t)-\phi(\lambda^2 t)
\]
is Bernstein. Here
\[
\psi_\tau(t)=(\tau-\lambda^\alpha)\,t^{\alpha/2}.
\]
Since \(\phi(t)=t^{\alpha/2}\) is Bernstein and the class of Bernstein functions is closed under multiplication by a nonnegative scalar~\cite[Corollary~3.7(i)]{schilling2009bernstein}, it follows that \(\psi_\tau\) is Bernstein if and only if \(\tau-\lambda^\alpha\ge 0\). Therefore
\[
\boxed{%
\begin{gathered}
\MMD_{k_\alpha}^2(f\#\nu,f\#\mu)\le \tau\,\MMD_{k_\alpha}^2(\nu,\mu)
\\
\text{for every }d\ge 1\text{ and every }\mu,\nu\in\mathcal P(\R^d)
\\[1pt]
\Longleftrightarrow
\\[1pt]
\tau\ge \lambda^\alpha.
\end{gathered}
}
\]

\noindent\emph{Via Theorem~\ref{thm:cbf_contraction_sup}.}
For \(0<\alpha<2\), the function \(\phi(t)=t^{\alpha/2}\) is a complete Bernstein function by \cite[Ch.~15.2, Identity~1]{schilling2009bernstein}, with \(b=0\) and
\[
m(s)=c_\alpha\,s^{-1-\alpha/2}.
\]
Hence Theorem~\ref{thm:cbf_contraction_sup} applies directly and gives
\[
\tau^\star(\lambda)
=
\sup_{s>0}\frac{m(s/\lambda^2)}{\lambda^2\,m(s)}
=
\sup_{s>0}\frac{c_\alpha (s/\lambda^2)^{-1-\alpha/2}}{\lambda^2 c_\alpha s^{-1-\alpha/2}}
=
\lambda^\alpha.
\]
For \(\alpha=2\), one has \(\phi(t)=t\), hence \(b=1\) and \(m\equiv 0\), so Theorem~\ref{thm:cbf_contraction_sup} gives \(\tau^\star(\lambda)=\lambda^2\). Thus in all cases \(\tau^\star(\lambda)=\lambda^\alpha\), which agrees with the Bernstein test above.

\noindent\textbf{Contraction factor.}
The smallest contraction factor across dimensions in \(\MMD_{k_\alpha}^2\) is
\[
\tau^\star(\lambda)=\lambda^\alpha\ (<1),
\]
so
\[
\MMD_{\!k_\alpha}^2(f\#\nu,f\#\mu)
\;\le\;\lambda^\alpha\,\MMD_{\!k_\alpha}^2(\nu,\mu).
\]
Equivalently,
\[
\MMD_{\!k_\alpha}(f\#\nu,f\#\mu)
\;\le\;\lambda^{\alpha/2}\,\MMD_{\!k_\alpha}(\nu,\mu).
\]

\medskip
\noindent\emph{Special case \(\alpha=2\).}
Since \(k_2(x,y)=-\|x-y\|^2\) can be written in the Bernstein form
\[
k_2(x,y)=C-\phi(\|x-y\|^2)
\quad\text{with}\quad
C=0,\ \phi(t)=t,
\]
Theorem~\ref{thm:bernstein_mmd_metric} applies and gives the criterion
\[
\MMD_{k_2}\ \text{is a metric}
\quad\Longleftrightarrow\quad
\Lambda\not\equiv 0.
\]
Here \(\phi(t)=t\) is linear, so its L\'evy measure satisfies \(\Lambda\equiv 0\).
Hence \(\MMD_{k_2}\) is not a metric on probability measures.
\end{proof}

\begin{proposition}[Contraction under generalized multiquadric kernels, generalizing {\cite{killingberg2023the}} and {\cite[Lemma~10]{debes2026distributional}}]\label{prop:gmq_contraction}
Fix \(h>0\) and \(\beta\in(0,1]\). Consider
\[
k_{\beta,h}(x,y)\;=\;-\bigl(1+h^2\|x-y\|^2\bigr)^{\beta},
\qquad
\phi_{\beta}(t)\;=\;\bigl(1+h^2 t\bigr)^{\beta}-1.
\]
Assume \(f(x)=q+\lambda x\) with \(\lambda\in(0,1)\). Then the smallest contraction factor across dimensions in \(\MMD_{k_{\beta,h}}^2\) is \(\tau^\star(\lambda)=\lambda^{2\beta}\).
\end{proposition}

\begin{proof}
Define
\[
\psi_\tau(t)=\tau\,\phi_{\beta}(t)-\phi_{\beta}(\lambda^2 t),
\qquad
\Gamma_\tau(x,y)=\psi_\tau(\|x-y\|^2).
\]

\noindent\textbf{L\'evy--Khintchine representation.} By \cite[Ch.~15.2, Identity~2]{schilling2009bernstein}, for \(0<\beta<1\),
\[
\phi_{\beta}(t)\;=\;\frac{\beta\,h^{2\beta}}{\Gamma(1-\beta)}
\int_{0}^{\infty}\bigl(1-e^{-s t}\bigr)\,e^{-s/h^{2}}\,s^{-\beta-1}\,ds,
\]
so \(b=0\) and the L\'evy measure is absolutely continuous with density
\[
m_{\beta,h}(s)\;=\;\frac{\beta\,h^{2\beta}}{\Gamma(1-\beta)}\,e^{-s/h^{2}}\,s^{-\beta-1},
\qquad s>0.
\]
For \(\beta=1\), \(\phi_{1}(t)=h^{2}t\) is linear, hence \(b=h^{2}\) and \(\Lambda\equiv0\).
In particular, \(\phi_{\beta}\) is a complete Bernstein function.

\noindent\textbf{Bernstein test for the difference profile.}
By Theorem~\ref{thm:bernstein_reduction}, in the present radial setting, contraction in \(\MMD^2\) reduces to checking whether the difference profile
\[
\psi_\tau(t)=\tau\,\phi_{\beta}(t)-\phi_{\beta}(\lambda^2 t)
\]
is Bernstein. Thus
\[
\MMD_{k_{\beta,h}}^2(f\#\nu,f\#\mu)\le \tau\,\MMD_{k_{\beta,h}}^2(\nu,\mu)
\qquad
\text{for every }d\ge 1\text{ and every }\mu,\nu\in\mathcal P(\R^d)
\]
holds if and only if \(\psi_\tau\) is Bernstein.

\noindent\emph{Via Theorem~\ref{thm:cbf_contraction_sup}.}
For \(0<\beta<1\), the representation above shows that \(\phi_\beta\) is a complete Bernstein function with \(b=0\) and
\[
m_{\beta,h}(s)=\frac{\beta\,h^{2\beta}}{\Gamma(1-\beta)}\,e^{-s/h^{2}}\,s^{-\beta-1}.
\]
Hence Theorem~\ref{thm:cbf_contraction_sup} applies directly, so the smallest contraction factor across dimensions is
\[
\tau^\star(\lambda)
=
\sup_{s>0}\frac{m_{\beta,h}(s/\lambda^{2})}{\lambda^{2}\,m_{\beta,h}(s)}.
\]
Using the formula for \(m_{\beta,h}\), we obtain for all \(s>0\),
\begin{align*}
\frac{m_{\beta,h}(s/\lambda^{2})}{\lambda^{2}\,m_{\beta,h}(s)}
&=
\frac{1}{\lambda^{2}}\,
\frac{
\frac{\beta\,h^{2\beta}}{\Gamma(1-\beta)}\,
e^{-\frac{s}{\lambda^{2}h^{2}}}\,
\left(\frac{s}{\lambda^{2}}\right)^{-\beta-1}
}{
\frac{\beta\,h^{2\beta}}{\Gamma(1-\beta)}\,
e^{-\frac{s}{h^{2}}}\,
s^{-\beta-1}
}
\\
&=
\frac{1}{\lambda^{2}}\,
\exp\Bigl(-\frac{s}{\lambda^{2}h^{2}}+\frac{s}{h^{2}}\Bigr)\,
\left(\frac{s}{\lambda^{2}}\right)^{-\beta-1}\,
s^{\beta+1}
\\
&=
\frac{1}{\lambda^{2}}\,
\exp\Bigl(-\frac{1-\lambda^{2}}{h^{2}\lambda^{2}}\,s\Bigr)\,
\lambda^{2(\beta+1)}
\\
&=
\lambda^{2\beta}\,
\exp\Bigl(-\frac{1-\lambda^{2}}{h^{2}\lambda^{2}}\,s\Bigr)
\ \le\ \lambda^{2\beta}.
\end{align*}
Since \(\lambda\in(0,1)\), the coefficient \(\frac{1-\lambda^{2}}{h^{2}\lambda^{2}}\) is positive, so the exponential factor converges to \(1\) as \(s\to 0^{+}\). Therefore the ratio converges to \(\lambda^{2\beta}\) as \(s\to 0^{+}\), and the supremum is exactly \(\lambda^{2\beta}\). Thus \(\tau^\star(\lambda)=\lambda^{2\beta}\) for \(0<\beta<1\).
For \(\beta=1\), one has \(\phi_1(t)=h^2 t\), hence \(b=h^2\) and \(m\equiv 0\), so Theorem~\ref{thm:cbf_contraction_sup} gives \(\tau^\star(\lambda)=\lambda^2\).
Therefore \(\tau^\star(\lambda)=\lambda^{2\beta}\) for all \(\beta\in(0,1]\).
Equivalently,
\[
\Gamma_\tau\ \text{is CND on }\R^d\ \text{for every }d\ge 1
\quad\Longleftrightarrow\quad
\tau\ge \lambda^{2\beta},
\]
or, by Theorem~\ref{thm:bernstein_reduction},
\[
\MMD_{k_{\beta,h}}^2(f\#\nu,f\#\mu)\le \tau\,\MMD_{k_{\beta,h}}^2(\nu,\mu)
\quad\Longleftrightarrow\quad
\tau\ge \lambda^{2\beta}.
\]

\noindent\textbf{Contraction factor.}
The smallest contraction factor across dimensions in \(\MMD^{2}\) is \(\tau^\star(\lambda)=\lambda^{2\beta}\), and the corresponding smallest contraction factor in \(\MMD\) is \(\sqrt{\tau^\star(\lambda)}=\lambda^{\beta}\), so
\[
\MMD_{k_{\beta,h}}(f\#\nu,f\#\mu)\ \le\ \lambda^{\beta}\,\MMD_{k_{\beta,h}}(\nu,\mu).
\]
In particular, for \(\lambda\in(0,1)\) one has \(\tau^\star(\lambda)<1\), so \(f\) is a contraction in \(\MMD^2\) with factor \(\tau^\star(\lambda)\).
\end{proof}

\begin{proposition}[No contraction under RBF kernels {\cite{nguyen2021distributional}}]\label{prop:rbf_no_contraction}
Fix \(h>0\) and consider the RBF kernel
\[
k_{\mathrm{rbf}}(x,y)=\exp\bigl(-h^{2}\|x-y\|^{2}\bigr),
\qquad
g(r)=e^{-h^{2}r^{2}}.
\]
Then
\[
g(0)-g(r)=1-e^{-h^{2}r^{2}}=\phi(r^{2}),
\qquad
\phi(t)=1-e^{-h^{2}t}.
\]
Assume \(f(x)=q+\lambda x\) with \(\lambda\in(0,1)\). Then there is no contraction factor in \(\MMD_{k_{\mathrm{rbf}}}^{2}\), that is, \(\tau^\star(\lambda)\ge 1\).
\end{proposition}

\begin{proof}
Define
\[
\psi_\tau(t)=\tau\,\phi(t)-\phi(\lambda^{2}t),
\qquad
\Gamma_\tau(x,y)=\psi_\tau(\|x-y\|^{2}).
\]

\noindent\textbf{L\'evy--Khintchine representation.}
The function \(\phi\) is a Bernstein function with
\[
\phi(t)=\int_{(0,\infty)}(1-e^{-st})\,\Lambda(ds),
\qquad
\Lambda=\delta_{h^{2}},
\qquad
b=0.
\]
Thus the L\'evy measure is a single atom at \(s=h^{2}\).

\noindent\textbf{Bernstein test for the difference profile.}
By Theorem~\ref{thm:bernstein_reduction}, in the present radial setting, contraction in \(\MMD^2\) reduces to checking whether the difference profile \(\psi_\tau\) is Bernstein. Equivalently,
\[
\MMD_{k_{\mathrm{rbf}}}^{2}(f\#\nu,f\#\mu)\le \tau\,\MMD_{k_{\mathrm{rbf}}}^{2}(\nu,\mu)
\qquad
\text{for every }d\ge 1\text{ and every }\mu,\nu\in\mathcal P(\R^d)
\]
holds if and only if \(\psi_\tau\) is Bernstein.

\noindent\emph{Via finitely many atoms.}
Since \(\Lambda=\delta_{h^{2}}\) has a single atom, Proposition~\ref{prop:bernstein-levy} rules out the required measure domination condition. Indeed, the case \(\tau\le \lambda^{2}\) is impossible because atoms must vanish, while for \(\tau>\lambda^{2}\) the atoms of \(\Lambda\) must lie on disjoint tails of geometric chains, so there must be infinitely many of them, not a single one. Hence \(\Gamma_\tau\) cannot be CND in every ambient dimension for any \(\tau\in(0,1)\).

\noindent\emph{Via saturating profiles.}
Proposition~\ref{prop:no_contraction_saturating_BF} gives the same conclusion, since \(\lim_{t\to\infty}\phi(t)=1\), so \(\phi\) is saturating.

\noindent\textbf{Contraction factor.}
Under the fixed affine map and the RBF kernel, there is no contraction factor in \(\MMD^{2}\). Equivalently,
\[
\tau^\star(\lambda)
=
\inf\{\tau>0:\ \Gamma_\tau\ \text{is CND on }\R^d\ \text{for every }d\ge 1\}
\ge
1.
\]
Hence no contraction holds in \(\MMD_{k_{\mathrm{rbf}}}^{2}\), and therefore no contraction holds in \(\MMD_{k_{\mathrm{rbf}}}\) either.
\end{proof}

\newpage
\section{Some new kernels inducing contraction}
\paragraph{Preamble.}
This section presents kernels that induce contraction and, to the best of our knowledge, are new. First, Section~\ref{app:cbf_search} goes through the complete Bernstein functions listed in \citep{schilling2009bernstein}. We use Theorem~\ref{thm:cbf_contraction_sup}, which gives the smallest contraction factor across dimensions as a supremum over one variable, to turn the contraction test into a numerical optimization problem. Second, Section~\ref{app:geometric_chains} applies the geometric chaining of Section~\ref{app:chaining_general_mechanism} to build kernels that induce contraction with a prescribed contraction factor. This gives the proof of Theorem~\ref{thm:main_gaussian_chain} of the main text, where the kernel is built from RBF kernels.

\subsection{From complete Bernstein functions}
\label{app:cbf_search}
\paragraph{Methodology.}
We started from the tables of complete Bernstein functions in \citep{schilling2009bernstein} and retained only the entries for which an explicit L\'evy density \(m(s)\) is available, since this is the regime needed to apply Theorem~\ref{thm:cbf_contraction_sup}. We then discarded the entries for which the profile \(\phi\) or the density \(m\) involves hypergeometric functions or other special functions not listed below. The retained entries are reported in Tables~\ref{tab:cbf_screen_candidates_a}--\ref{tab:cbf_screen_candidates_d}.

\paragraph{Notation for special functions.}
In Tables~\ref{tab:cbf_screen_candidates_a}--\ref{tab:cbf_screen_candidates_d}, we use the notation for special functions of \citep[Section~15.1]{schilling2009bernstein}. The \emph{Euler gamma function} is
\[
\Gamma(z)=\int_0^\infty e^{-t} t^{z-1}\,dt.
\]
The \emph{lower and upper incomplete gamma functions} are
\[
\gamma(a,x)=\int_0^x t^{a-1} e^{-t}\,dt,
\qquad
\Gamma(a,x)=\int_x^\infty t^{a-1} e^{-t}\,dt
\]
(Entries~14, 63, 64, 85, 86). The \emph{modified Bessel functions} are
\[
I_\nu(x)=\sum_{n=0}^\infty \frac{(x/2)^{\nu+2n}}{n!\,\Gamma(\nu+n+1)},
\qquad
K_\nu(x)=\frac{\pi}{2}\,\frac{I_{-\nu}(x)-I_\nu(x)}{\sin(\nu\pi)},
\]
so that \(I_0=I_\nu|_{\nu=0}\), \(I_1=I_\nu|_{\nu=1}\), and \(K_0=K_\nu|_{\nu=0}\) (Entries~7, 57, 77, 78). The \emph{exponential integral} is
\[
\operatorname{Ei}(x)=-\int_{-x}^{\infty}\frac{e^{-t}}{t}\,dt,
\]
interpreted as a Cauchy principal value when \(x>0\) (Entries~29, 30, 33, 100--104). The \emph{error and complementary error functions} are
\[
\operatorname{Erf}(x)=\frac{2}{\sqrt{\pi}}\int_0^x e^{-t^2}\,dt,
\qquad
\operatorname{Erfc}(x)=1-\operatorname{Erf}(x)
\]
(Entries~38, 108, 109). The \emph{generalized exponential integral} is
\[
E_n(x)=\int_1^\infty \frac{e^{-xt}}{t^n}\,dt
\]
(Entry~99). The \emph{inverse hyperbolic cosine} is
\[
\operatorname{arcosh}(x)=\log\!\bigl(x+\sqrt{x^2-1}\bigr)
\]
(Entries~77, 78). Finally, \(\log\Gamma\) denotes the logarithm of the Euler gamma function, namely \(x\mapsto \log\Gamma(x)\) (Entry~88).

\newpage
\newcommand{\cbfnumyes}{\textcolor{green!50!black}{\textbf{V}}}
\newcommand{\cbfnumno}{\textcolor{red!70!black}{\textbf{X}}}
\newcommand{\cbfsource}[1]{{\small \citep[Entry~#1]{schilling2009bernstein}}}
\newcommand{\cbfrange}[1]{Valid for #1}
\newcommand{\cbfrangenone}{No free parameter.}
\newcommand{\cbfsep}{\\ \specialrule{0.25pt}{0pt}{0pt}}
\newcommand{\cbfentry}[3]{%
\begin{tabular}[t]{@{}l@{}}
{\footnotesize #1}\\[2pt]
{\small #2}\\[2pt]
{\small #3}
\end{tabular}}

\begin{center}
\footnotesize
\setlength{\tabcolsep}{4pt}
\renewcommand{\arraystretch}{1.14}
\begin{tabular}{p{0.8cm} p{10.6cm} c c}
\toprule
{\small\bfseries Entry} & {\small\bfseries Profile \(\phi\), L\'evy density \(m\), and parameter range} & {\small\bfseries Ana.} & {\small\bfseries Num.} \\
\midrule
1 & \cbfentry{$\phi(t)=t^\alpha$}{$m(s)=\dfrac{\alpha}{\Gamma(1-\alpha)}\,s^{-1-\alpha}$}{\cbfrange{$0<\alpha<1$}} & \cbfnumyes & \cbfnumyes \cbfsep
2 & \cbfentry{$\phi(t)=(t+1)^\alpha-1$}{$m(s)=\dfrac{\alpha}{\Gamma(1-\alpha)}\,e^{-s}s^{-1-\alpha}$}{\cbfrange{$0<\alpha<1$}} & \cbfnumyes & \cbfnumyes \cbfsep
3 & \cbfentry{$\phi(t)=1-(1+t)^{\alpha-1}$}{$m(s)=\dfrac{1}{\Gamma(1-\alpha)}\,e^{-s}s^{-\alpha}$}{\cbfrange{$0<\alpha<1$}} & \cbfnumno & \cbfnumno \cbfsep
4 & \cbfentry{$\phi(t)=\dfrac{t}{t+a}$}{$m(s)=a\,e^{-as}$}{\cbfrange{$a>0$}} & \cbfnumno & \cbfnumno \cbfsep
5 & \cbfentry{$\phi(t)=\dfrac{t}{\sqrt{t+a}}$}{$m(s)=\dfrac{e^{-as}(2as+1)}{2\sqrt{\pi}\,s^{3/2}}$}{\cbfrange{$a>0$}} &  & \cbfnumyes \cbfsep
6 & \cbfentry{$\phi(t)=\dfrac{t}{(t+a)^\alpha}$}{$m(s)=\dfrac{\sin(\alpha\pi)\Gamma(1-\alpha)}{\pi}\,e^{-as}s^{\alpha-2}(as+1-\alpha)$}{\cbfrange{$a>0,\ 0<\alpha<1$}} &  & \cbfnumyes \cbfsep
7 & \cbfentry{$\phi(t)=\sqrt{\dfrac{t}{t+a}}$}{$m(s)=\dfrac{a}{2}e^{-as/2}\left(I_0\left(\dfrac{as}{2}\right)-I_1\left(\dfrac{as}{2}\right)\right)$}{\cbfrange{$a>0$}} &  & \cbfnumno \cbfsep
14 & \cbfentry{$\phi(t)=\dfrac{\pi}{\sin(\alpha\pi)}\,\dfrac{t(c^\alpha-t^\alpha)}{c-t}$}{$m(s)=\Gamma(\alpha+2)c^{\alpha+1}e^{cs}\Gamma(-\alpha-1,cs)$}{\cbfrange{$c>0,\ -1<\alpha<1$}} &  & \cbfnumyes \cbfsep
18 & \cbfentry{$\phi(t)=\sqrt{t}\left(1-e^{-2a\sqrt{t}}\right)$}{$m(s)=\dfrac{(2a^2-s)e^{-a^2/s}+s}{2\sqrt{\pi}\,s^{5/2}}$}{\cbfrange{$a>0$}} &  & \cbfnumyes \cbfsep
19 & \cbfentry{$\phi(t)=\sqrt{t}\left(1+e^{-2a\sqrt{t}}\right)$}{$m(s)=\dfrac{(s-2a^2)e^{-a^2/s}+s}{2\sqrt{\pi}\,s^{5/2}}$}{\cbfrange{$a>0$}} &  & \cbfnumno \cbfsep
20 & \cbfentry{$\phi(t)=\dfrac{t\left(1-e^{-2\sqrt{t+a}}\right)}{\sqrt{t+a}}$}{$m(s)=\dfrac{e^{-1/s-as}\left(2+s(e^{1/s}-1)(1+2as)\right)}{2\sqrt{\pi}\,s^{5/2}}$}{\cbfrange{$a>0$}} &  & \cbfnumyes \\
\bottomrule
\end{tabular}
\captionof{table}{Complete Bernstein functions of \citep{schilling2009bernstein} with an explicit L\'evy density. Entry numbers refer to \citep{schilling2009bernstein}. Our table departs from \citep{schilling2009bernstein} in nine entries. For Entries~30, 104 and~108, we correct \(\phi\), and for Entries~82 and~84, we correct the constant factor in \(m\). For Entry~85, we restrict the range \(\nu\in\mathbb{R}\) to \(\nu<1\), where \(m\) is positive. For Entries~64, 131 and~132, we write \(m\) or \(\phi\) in another form. The last two columns tell whether the entry induces contraction (\cbfnumyes) or not (\cbfnumno), according to an analytic result of this paper (Ana.) and to the numerical search (Num.). An empty Ana. cell means that the paper has no analytic result for that entry.}
\label{tab:cbf_screen_candidates_a}
\end{center}

\begin{center}
\footnotesize
\setlength{\tabcolsep}{4pt}
\renewcommand{\arraystretch}{1.14}
\begin{tabular}{p{0.8cm} p{10.6cm} c c}
\toprule
{\small\bfseries Entry} & {\small\bfseries Profile \(\phi\), L\'evy density \(m\), and parameter range} & {\small\bfseries Ana.} & {\small\bfseries Num.} \\
\midrule
26 & \cbfentry{$\phi(t)=\log\left(1+\dfrac{t}{a}\right)$}{$m(s)=\dfrac{e^{-as}}{s}$}{\cbfrange{$a>0$}} & \cbfnumno & \cbfnumno \cbfsep
27 & \cbfentry{$\phi(t)=t\log\left(1+\dfrac{a}{t}\right)$}{$m(s)=\dfrac{1-e^{-as}(1+as)}{s^2}$}{\cbfrange{$a>0$}} &  & \cbfnumno \cbfsep
28 & \cbfentry{$\phi(t)=\log\left(\dfrac{b(t+a)}{a(t+b)}\right)$}{$m(s)=\dfrac{e^{-as}-e^{-bs}}{s}$}{\cbfrange{$0<a<b$}} &  & \cbfnumno \cbfsep
29 & \cbfentry{$\phi(t)=\dfrac{1}{a}-\dfrac{1}{t}\log\left(1+\dfrac{t}{a}\right)$}{$m(s)=-\operatorname{Ei}(-as)$}{\cbfrange{$a>0$}} &  & \cbfnumno \cbfsep
30 & \cbfentry{$\phi(t)=\dfrac{a+t}{at}\log\left(1+\dfrac{t}{a}\right)-\dfrac{1}{a}$}{$m(s)=\dfrac{e^{-as}}{as}+\operatorname{Ei}(-as)$}{\cbfrange{$a>0$}} &  & \cbfnumno \cbfsep
31 & \cbfentry{$\phi(t)=(t+a)\log(t+a)-t\log t-a\log a$}{$m(s)=\dfrac{1-e^{-as}}{s^2}$}{\cbfrange{$a>0$}} &  & \cbfnumno \cbfsep
32 & \cbfentry{$\phi(t)=(t+b)\log(t+b)-b\log b-(t+a)\log(t+a)+a\log a$}{$m(s)=\dfrac{e^{-as}-e^{-bs}}{s^2}$}{\cbfrange{$0<a<b$}} &  & \cbfnumno \cbfsep
33 & \cbfentry{$\phi(t)=\dfrac{t}{a-t}\log\left(\dfrac{a}{t}\right)$}{$m(s)=a e^{as}\operatorname{Ei}(-as)+s^{-1}$}{\cbfrange{$a>0$}} &  & \cbfnumno \cbfsep
38 & \cbfentry{$\phi(t)=\dfrac12\left(a^2+a\sqrt{a^2+2t}\right)-a^2+t\log\left(\sqrt{1+\dfrac{a^2}{2t}}+\dfrac{a}{\sqrt{2t}}\right)$}{$m(s)=\dfrac{1}{2s^2}\operatorname{Erf}\left(a\sqrt{\dfrac{s}{2}}\right)$}{\cbfrange{$a>0$}} &  & \cbfnumyes \cbfsep
45 & \cbfentry{$\phi(t)=t(1+t)\log\left(1+\dfrac{1}{t}\right)-t$}{$m(s)=\dfrac{e^{-s}(2+s)+s-2}{s^3}$}{\cbfrangenone} &  & \cbfnumno \cbfsep
46 & \cbfentry{$\phi(t)=t\left(1-t\log\left(1+\dfrac{1}{t}\right)\right)$}{$m(s)=\dfrac{2-e^{-s}(2+2s+s^2)}{s^3}$}{\cbfrangenone} &  & \cbfnumno \\
\bottomrule
\end{tabular}
\captionof{table}{Continuation of Table~\ref{tab:cbf_screen_candidates_a}.}
\label{tab:cbf_screen_candidates_b}
\end{center}

\begin{center}
\footnotesize
\setlength{\tabcolsep}{4pt}
\renewcommand{\arraystretch}{1.14}
\begin{tabular}{p{0.8cm} p{10.6cm} c c}
\toprule
{\small\bfseries Entry} & {\small\bfseries Profile \(\phi\), L\'evy density \(m\), and parameter range} & {\small\bfseries Ana.} & {\small\bfseries Num.} \\
\midrule
57 & \cbfentry{$\phi(t)=\log\left(1+2t+2\sqrt{t(1+t)}\right)$}{$m(s)=s^{-1}e^{-s/2}I_0\left(\dfrac{s}{2}\right)$}{\cbfrangenone} &  & \cbfnumno \cbfsep
63 & \cbfentry{$\phi(t)=\sqrt{t}\,\arctan\left(\sqrt{\dfrac{a}{t}}\right)$}{$m(s)=\dfrac{\gamma\left(\frac32,as\right)}{2s^{3/2}}$}{\cbfrange{$a>0$}} &  & \cbfnumno \cbfsep
64 & \cbfentry{$\phi(t)=\sqrt{t}\,\arctan\left(\sqrt{\dfrac{t}{a}}\right)$}{$m(s)=\dfrac{\Gamma\left(\frac32,as\right)}{2s^{3/2}}$}{\cbfrange{$a>0$}} &  & \cbfnumyes \cbfsep
77 & \cbfentry{$\phi(t)=\operatorname{arcosh}(1+t)=\log\left(1+t+\sqrt{t(t+2)}\right)$}{$m(s)=\dfrac{e^{-s}I_0(s)}{s}$}{\cbfrangenone} &  & \cbfnumno \cbfsep
78 & \cbfentry{$\phi(t)=\dfrac{1}{2}\operatorname{arcosh}^2(1+t)$}{$m(s)=\dfrac{e^{-s}K_0(s)}{s}$}{\cbfrangenone} &  & \cbfnumno \cbfsep
82 & \cbfentry{$\phi(t)=\dfrac{\Gamma\left(\frac{t+a}{2a}\right)}{\Gamma\left(\frac{t}{2a}\right)}$}{$m(s)=\dfrac{a e^{2as}}{\sqrt{\pi}\left(e^{2as}-1\right)^{3/2}}$}{\cbfrange{$a>0$}} &  & \cbfnumyes \cbfsep
83 & \cbfentry{$\phi(t)=\dfrac{\Gamma(\alpha t+1)}{\Gamma(\alpha t+1-\alpha)}$}{$m(s)=\dfrac{e^{-s/\alpha}}{\Gamma(1-\alpha)\left(1-e^{-s/\alpha}\right)^{1+\alpha}}$}{\cbfrange{$0<\alpha<1$}} &  & \cbfnumyes \cbfsep
84 & \cbfentry{$\phi(t)=\dfrac{t\,\Gamma(\alpha t+1-\alpha)}{\Gamma(\alpha t+1)}$}{$m(s)=\dfrac{(1-\alpha)e^{s/\alpha}}{\alpha\Gamma(1+\alpha)\left(e^{s/\alpha}-1\right)^{2-\alpha}}$}{\cbfrange{$0<\alpha<1$}} &  & \cbfnumyes \cbfsep
85 & \cbfentry{$\phi(t)=t^{1-\nu}e^t\Gamma(\nu,t)$}{$m(s)=\dfrac{\Gamma(2-\nu)}{\Gamma(1-\nu)}(1+s)^{\nu-2}$}{\cbfrange{$\nu<1$}} &  & \cbfnumno \cbfsep
86 & \cbfentry{$\phi(t)=t^{1-\nu}e^{at}\Gamma(\nu,at)$}{$m(s)=\dfrac{\Gamma(2-\nu)}{\Gamma(1-\nu)}(a+s)^{\nu-2}$}{\cbfrange{$a>0,\ 0<\nu<1$}} &  & \cbfnumno \cbfsep
88 & \cbfentry{$\phi(t)=\log\Gamma(t+a+b)-\log\Gamma(t+a)+\log\Gamma(a)-\log\Gamma(a+b)$}{$m(s)=\dfrac{e^{-as}\left(1-e^{-bs}\right)}{s\left(1-e^{-s}\right)}$}{\cbfrange{$a,b>0$}} &  & \cbfnumno \\
\bottomrule
\end{tabular}
\captionof{table}{Continuation of Table~\ref{tab:cbf_screen_candidates_a}.}
\label{tab:cbf_screen_candidates_c}
\end{center}

\begin{center}
\footnotesize
\setlength{\tabcolsep}{4pt}
\renewcommand{\arraystretch}{1.14}
\begin{tabular}{p{0.8cm} p{10.6cm} c c}
\toprule
{\small\bfseries Entry} & {\small\bfseries Profile \(\phi\), L\'evy density \(m\), and parameter range} & {\small\bfseries Ana.} & {\small\bfseries Num.} \\
\midrule
99 & \cbfentry{$\phi(t)=t e^t E_n(t)$}{$m(s)=\dfrac{n}{(1+s)^{n+1}}$}{\cbfrange{$n\in\mathbb{N}$}} & \cbfnumno & \cbfnumno \cbfsep
100 & \cbfentry{$\phi(t)=-t e^{at}\operatorname{Ei}(-at)$}{$m(s)=\dfrac{1}{(a+s)^2}$}{\cbfrange{$a>0$}} & \cbfnumno & \cbfnumno \cbfsep
101 & \cbfentry{$\phi(t)=t e^{at}\left(\operatorname{Ei}(-a(b+t))-\operatorname{Ei}(-at)\right)$}{$m(s)=\dfrac{1-(b(s+a)+1)e^{-(s+a)b}}{(s+a)^2}$}{\cbfrange{$a,b>0$}} &  & \cbfnumno \cbfsep
102 & \cbfentry{$\phi(t)=-t e^{at}\operatorname{Ei}(-a(b+t))$}{$m(s)=\dfrac{(b(s+a)+1)e^{-(s+a)b}}{(s+a)^2}$}{\cbfrange{$a,b>0$}} & \cbfnumno & \cbfnumno \cbfsep
103 & \cbfentry{$\phi(t)=(-1)^{n+1}t^{n+1}e^{at}\operatorname{Ei}(-at)+\sum_{k=1}^{n}(-1)^{n-k}(k-1)!a^{-k}t^{n-k+1}$}{$m(s)=\dfrac{(n+1)!}{(a+s)^{n+2}}$}{\cbfrange{$a>0,\ n\in\mathbb{N}$}} & \cbfnumno & \cbfnumno \cbfsep
104 & \cbfentry{$\phi(t)=\gamma_E+\log(at)-e^{at}\operatorname{Ei}(-at)$}{$m(s)=\dfrac{a}{s(a+s)}$}{\cbfrange{$a>0$}} &  & \cbfnumno \cbfsep
108 & \cbfentry{$\phi(t)=\sqrt{\pi t}\,e^{at}\operatorname{Erfc}\left(\sqrt{at}\right)$}{$m(s)=\dfrac{1}{2(a+s)^{3/2}}$}{\cbfrange{$a>0$}} & \cbfnumno & \cbfnumno \cbfsep
109 & \cbfentry{$\phi(t)=\sqrt{\dfrac{\pi}{a}}\,t-\pi t^{3/2}e^{at}\operatorname{Erfc}\left(\sqrt{at}\right)$}{$m(s)=\Gamma\left(\dfrac52\right)(a+s)^{-5/2}$}{\cbfrange{$a>0$}} & \cbfnumno & \cbfnumno \cbfsep
131 & \cbfentry{$\phi(t)=-(2a)^\nu\Gamma(-\nu)\,\dfrac{t\,\Gamma\left(1+\nu+\frac{t}{2a}\right)}{(2a\nu+t)\Gamma\left(1+\frac{t}{2a}\right)}$}{$m(s)=\left(\dfrac{a}{\sinh(as)}\right)^{1+\nu}e^{(1-\nu)as}$}{\cbfrange{$a>0,\ 0<\nu<1$}} &  & \cbfnumyes \cbfsep
132 & \cbfentry{$\phi(t)=(2a)^\nu\Gamma(-\nu)\left(\dfrac{\Gamma(\mu)}{\Gamma(1-\mu)}-\dfrac{\Gamma\left(\mu+\frac{t}{2a}\right)}{\Gamma\left(1-\mu+\frac{t}{2a}\right)}\right)$}{$m(s)=\left(\dfrac{a}{\sinh(as)}\right)^{1+\nu}$}{\cbfrange{$\mu=\dfrac{1+\nu}{2},\ a>0,\ 0<\nu<1$}} &  & \cbfnumyes \\
\bottomrule
\end{tabular}
\captionof{table}{Continuation of Table~\ref{tab:cbf_screen_candidates_a}.}
\label{tab:cbf_screen_candidates_d}
\end{center}

\paragraph{Reduction to a supremum.}
For \(\lambda\in(0,1)\), Theorem~\ref{thm:cbf_contraction_sup} gives the smallest contraction factor across dimensions of a complete Bernstein function as
\[
\tau^\star(\lambda)=\max\!\left\{\lambda^2\mathbf 1_{\{b>0\}},\ \sup_{s>0}R_\lambda(s)\right\},
\qquad
R_\lambda(s):=\frac{m(s/\lambda^2)}{\lambda^2\,m(s)}.
\]
Every entry of Tables~\ref{tab:cbf_screen_candidates_a}--\ref{tab:cbf_screen_candidates_d} has drift \(b=0\), so \(\tau^\star(\lambda)\) is the supremum of \(R_\lambda\) over \(s>0\). Deciding whether an entry induces contraction thus amounts to computing, for each \(\lambda\), the supremum of a function of one variable.

\paragraph{Two difficulties.}
Computing this supremum numerically raises two difficulties. First, \(s\) ranges over \((0,\infty)\), but a numerical search can only explore a bounded range. The supremum may even be approached only as \(s\to0^+\) or as \(s\to\infty\). The search mitigates this with a wide window, and reports when its best point lies at an end of the window, since the supremum may then lie further out. Second, the shape of \(R_\lambda\) is not known in advance, and a single local search may stop at a local maximum below the supremum. The search mitigates this by combining a fine grid over the whole window with local searches around its best points.

\paragraph{Logarithmic scale.}
To treat small and large values of \(s\) alike, we write \(s=e^u\) and search over \(u\in\R\). A window \([-U,U]\) in \(u\) then covers \(s\) from \(e^{-U}\) to \(e^{U}\), and a uniform grid in \(u\) puts about the same number of points in each interval \([10^k,10^{k+1}]\) of \(s\). We also maximize the logarithm of the ratio,
\[
\ell_\lambda(u):=\log R_\lambda(e^u)=\log m\left(\frac{e^u}{\lambda^2}\right)-\log m(e^u)-\log(\lambda^2),
\]
which avoids overflow and underflow when \(m\) is very large or very small. Since the logarithm is increasing,
\[
\tau^\star(\lambda)=\sup_{s>0}R_\lambda(s)=\exp\Bigl(\sup_{u\in\R}\ell_\lambda(u)\Bigr).
\]

\paragraph{The search.}
The search maximizes \(\ell_\lambda\) on a window \([-U,U]\) in two steps.
\begin{enumerate}
    \item \textbf{Grid.} It evaluates \(\ell_\lambda\) at \(n_{\mathrm{grid}}\) equally spaced points of \([-U,U]\), spaced \(\Delta u=2U/(n_{\mathrm{grid}}-1)\) apart, and selects the \(K\) points with the largest values.
    \item \textbf{Local searches.} It then solves \(K\) local maximization problems, one around each selected point \(u_0\), on the interval \([u_0-c_{\mathrm{loc}}\Delta u,\ u_0+c_{\mathrm{loc}}\Delta u]\) cut to \([-U,U]\). We use \texttt{minimize\_scalar}, the routine from SciPy \citep{2020SciPy-NMeth} for optimizing a function of one variable on an interval. The best of the \(K\) results is denoted \(u^\star\).
\end{enumerate}
The search returns \(\exp(\ell_\lambda(u^\star))\) as its estimate of \(\tau^\star(\lambda)\), together with whether \(u^\star\) is an end of the window. Algorithm~\ref{alg:cbf_sup_search} gives the details, and Table~\ref{tab:cbf_sup_hparams} lists the values of \(U\), \(n_{\mathrm{grid}}\), \(K\) and \(c_{\mathrm{loc}}\). The search does not guarantee that the supremum is found.
\begin{algorithm}[t]
\caption{Search for the supremum of the ratio \(R_\lambda\)}
\label{alg:cbf_sup_search}
\begin{algorithmic}[1]
  \STATE {\bfseries Require:} Discount \(\lambda\in(0,1)\), L\'evy density \(m:(0,\infty)\to(0,\infty)\)
  \STATE {\bfseries Require:} Parameters of Table~\ref{tab:cbf_sup_hparams} \(U,n_{\mathrm{grid}},K,c_{\mathrm{loc}}\)
  \STATE {\bfseries Output:} Estimate \(\widehat{\tau}^{\star}(\lambda)\) of \(\sup_{s>0}R_\lambda(s)\), best point \(u^\star\), \(s^\star=e^{u^\star}\), and whether \(u^\star\) is an end of the window
  \STATE Set
  \[
  \ell_\lambda(u)\gets \log m(e^u/\lambda^2)-\log m(e^u)-\log(\lambda^2)
  \]
  \STATE \textcolor{green!50!black}{\textit{Evaluate the objective \(\ell_\lambda\) on a grid over the window.}}
  \STATE Form a uniform grid \(u_1,\dots,u_{n_{\mathrm{grid}}}\) on \([-U,U]\)
  \STATE Evaluate \(\ell_\lambda(u_j)\) on the grid and select the \(K\) grid points with the largest values
  \STATE Set
  \[
  \Delta u\gets \frac{2U}{n_{\mathrm{grid}}-1}
  \]
  \FOR{each of the \(K\) selected grid points \(u_j\)}
    \STATE Define the local interval
    \[
    [a_j,b_j]\gets \bigl[\max(-U,\,u_j-c_{\mathrm{loc}}\Delta u),\ \min(U,\,u_j+c_{\mathrm{loc}}\Delta u)\bigr]
    \]
    \STATE Maximize \(\ell_\lambda\) on \([a_j,b_j]\) with \texttt{minimize\_scalar}, giving the point \(\tilde u_j\) and its value \(\tilde L_j\)
  \ENDFOR
  \STATE Let \((u^\star,L^\star)\) be the pair \((\tilde u_j,\tilde L_j)\) with the largest value \(\tilde L_j\)
  \STATE \textbf{return} \(\exp(L^\star),u^\star,e^{u^\star}\), and whether \(|u^\star|=U\)
\end{algorithmic}
\end{algorithm}

\begin{table}[t]
\centering
\small
\begin{tabular}{@{}lp{1.0cm}p{7.5cm}@{}}
\toprule
\textbf{Setting} & \textbf{Value} & \textbf{Description} \\
\midrule
Window \(U\) & \(80\) & The window is \([-U,U]\). \\
\midrule
Grid size \(n_{\mathrm{grid}}\) & \(8500\) & Number of grid points on the window. \\
\midrule
Selected grid points \(K\) & \(12\) & Number of best grid points around which \(\ell_\lambda\) is maximized locally. \\
\midrule
Local interval \(c_{\mathrm{loc}}\) & \(120\) & The local interval around a selected grid point \(u_0\) is \([u_0-c_{\mathrm{loc}}\Delta u,\ u_0+c_{\mathrm{loc}}\Delta u]\). \\
\midrule
Local tolerance & \(10^{-10}\) & Precision on \(u\) at which \texttt{minimize\_scalar} stops. \\
\bottomrule
\end{tabular}
\caption{Values of the constants of Algorithm~\ref{alg:cbf_sup_search}.}
\label{tab:cbf_sup_hparams}
\end{table}

\paragraph{Experimental setup.}
We run the search on CPUs, at \(100\) equally spaced values of \(\lambda\) from \(0.01\) to \(0.99\), with the constants of Table~\ref{tab:cbf_sup_hparams}. Entries with parameters use \(\alpha=0.6\) in Entries~1--3, \(\alpha=\nu=0.5\) in the others, \(b=3\) in Entries~28 and~32, \(c=2\) in Entry~14, and \(a=b=n=1\) otherwise. Figure~\ref{fig:cbf_parametric_regimes} varies one of these parameters.

\paragraph{Numerical precision.}
The values of the density \(m\) are computed with \texttt{mpmath} \citep{mpmath} at 80 significant digits, so that they stay accurate where they are very large or very small. Their logarithms are then rounded to double precision, in which the grid and the local searches work.

\paragraph{Checks against known results.}
Entries~1 and~2 give the power-distance and generalized multiquadric kernels, whose smallest contraction factors across dimensions are known in closed form (Propositions~\ref{prop:power_distance_contraction} and~\ref{prop:gmq_contraction}). For both entries, the search recovers these values up to rounding errors. Entry~2 shows that the search can find the supremum even when it is approached only as \(s\to0^+\). Proposition~\ref{prop:three_kernels_no_contraction} proves that Entries~3, 4 and~26 do not induce contraction, in agreement with the search. By Corollary~\ref{corr:m0_limit_metric}, an entry does not induce contraction if its L\'evy density \(m(s)\) has a finite limit as \(s\to0^+\). Entries~99, 100, 102, 103, 108 and~109 meet this condition. More entries can be confirmed with this corollary, but for brevity we only list those for which the limit is immediate.

\clearpage
\subsubsection{Entries that do not induce contraction}

\begin{figure}[H]
    \centering
    \includegraphics[width=0.95\textwidth]{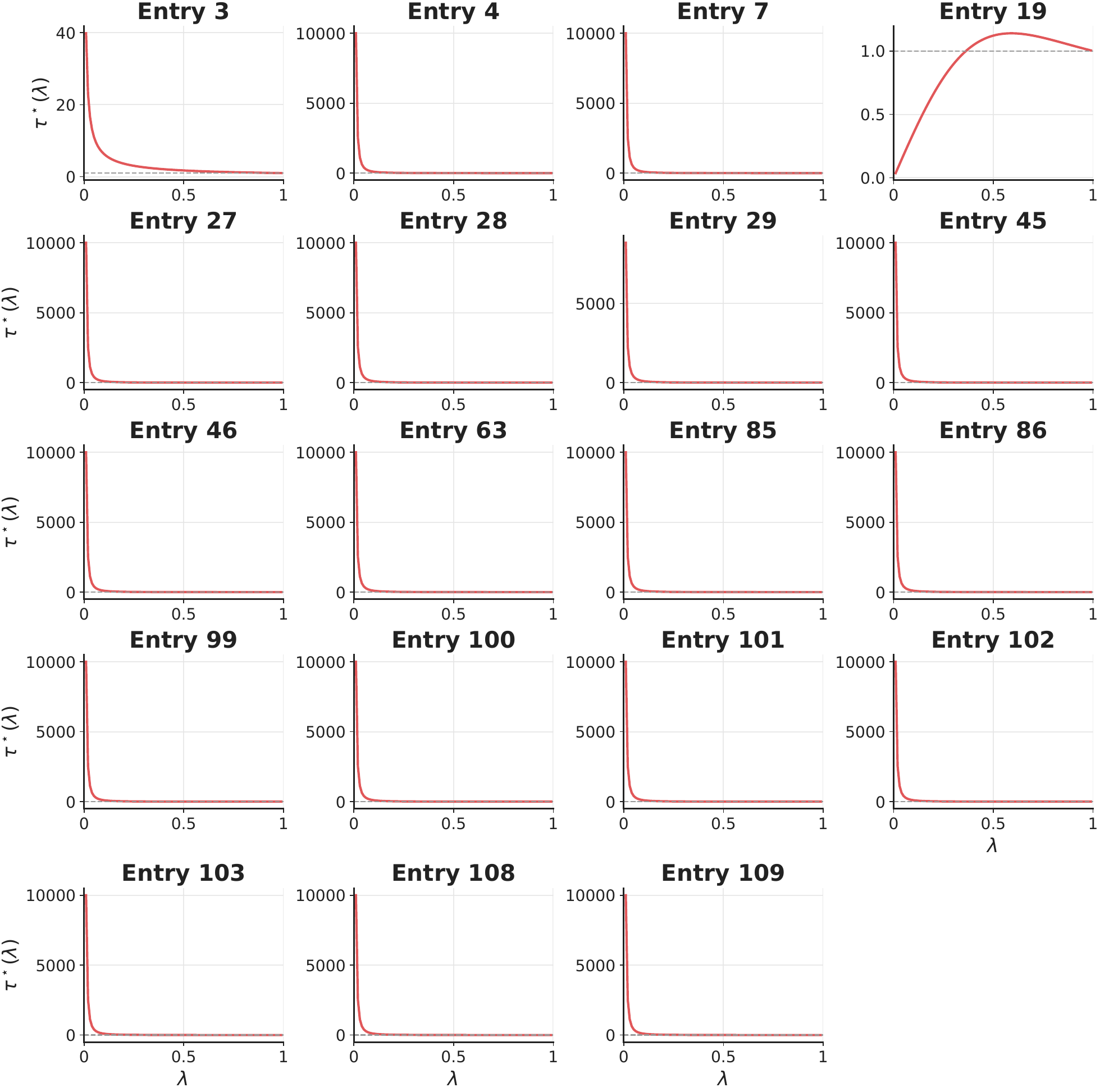}
    \caption{Entries whose estimate of \(\tau^\star(\lambda)\) exceeds \(1\) for some \(\lambda\). Entry numbers refer to \citep{schilling2009bernstein}.}
    \label{fig:cbf_rejected_tau_star_curves}
\end{figure}

\begin{figure}[t]
    \centering
    \includegraphics[width=0.95\textwidth]{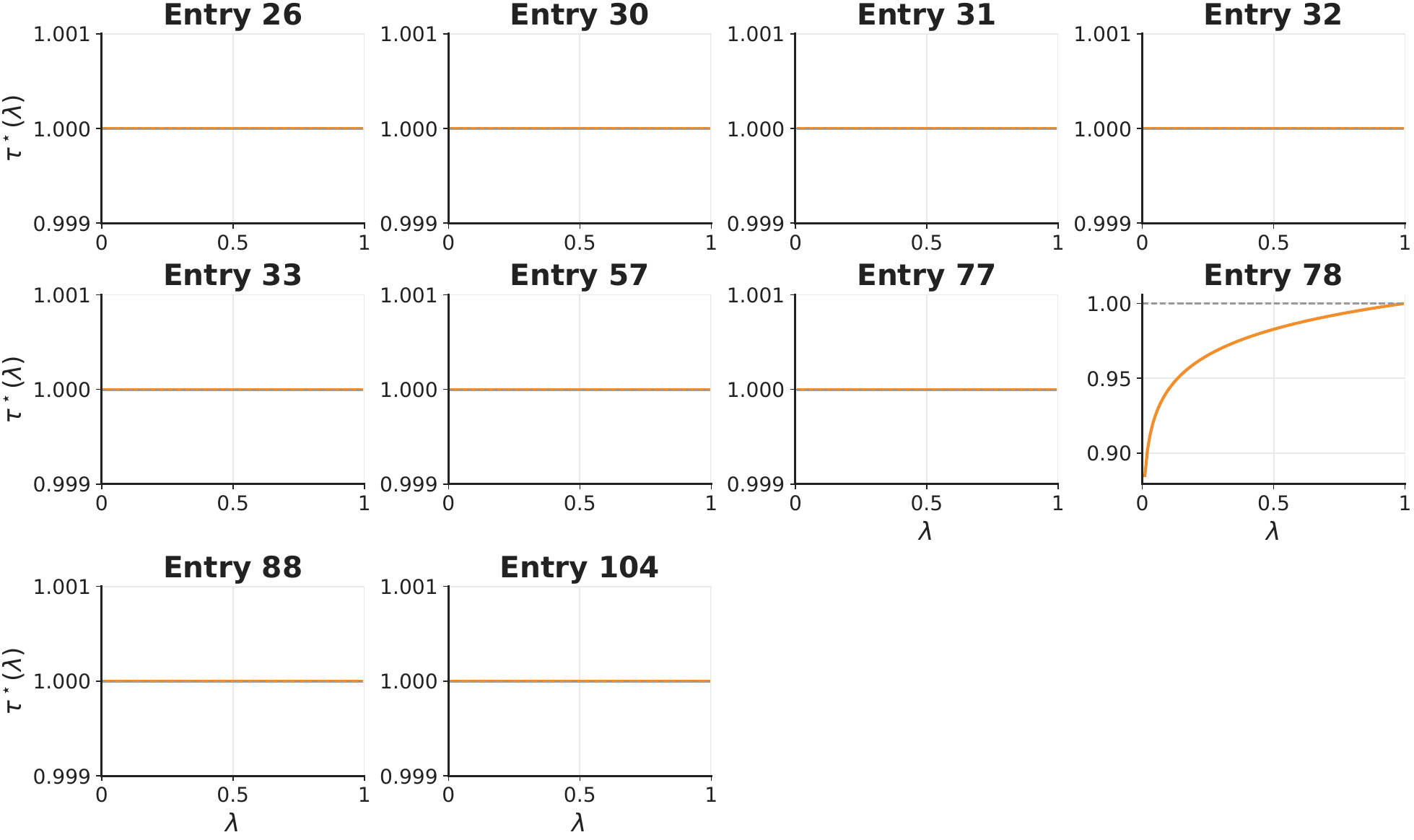}
    \caption{Entries whose estimate of \(\tau^\star(\lambda)\) equals \(1\) up to numerical precision at every \(\lambda\), and Entry~78. Entry numbers refer to \citep{schilling2009bernstein}.}
    \label{fig:cbf_possibly_non_expansive_tau_star_curves}
\end{figure}

The search finds two groups of entries that do not induce contraction. For the entries of Figure~\ref{fig:cbf_rejected_tau_star_curves}, the estimate of \(\tau^\star(\lambda)\) exceeds \(1\) for some \(\lambda\). Since the estimate never exceeds \(\tau^\star(\lambda)\), these entries do not induce contraction. For the entries of Figure~\ref{fig:cbf_possibly_non_expansive_tau_star_curves}, the estimate equals \(1\) up to numerical precision for every \(\lambda\). These entries do not induce contraction either, but they may be non-expansive.

Entry~78 also belongs to the second group, although its estimate stays below \(1\). At every \(\lambda\), its best point is the left end of the window, \(u^\star=-80\), so its estimate in Figure~\ref{fig:cbf_possibly_non_expansive_tau_star_curves} may miss a larger value further left. Further left, the ratio keeps increasing towards \(1\). We conclude that, for Entry~78, \(\tau^\star(\lambda)=1\) for every \(\lambda\in(0,1)\), although its estimate in Figure~\ref{fig:cbf_possibly_non_expansive_tau_star_curves}, limited by the window, stays below \(1\).

\subsubsection{Entries that induce contraction}
\label{app:positive_results_cmpl}
\begin{figure}[htbp]
    \centering
    \includegraphics[width=0.95\textwidth]{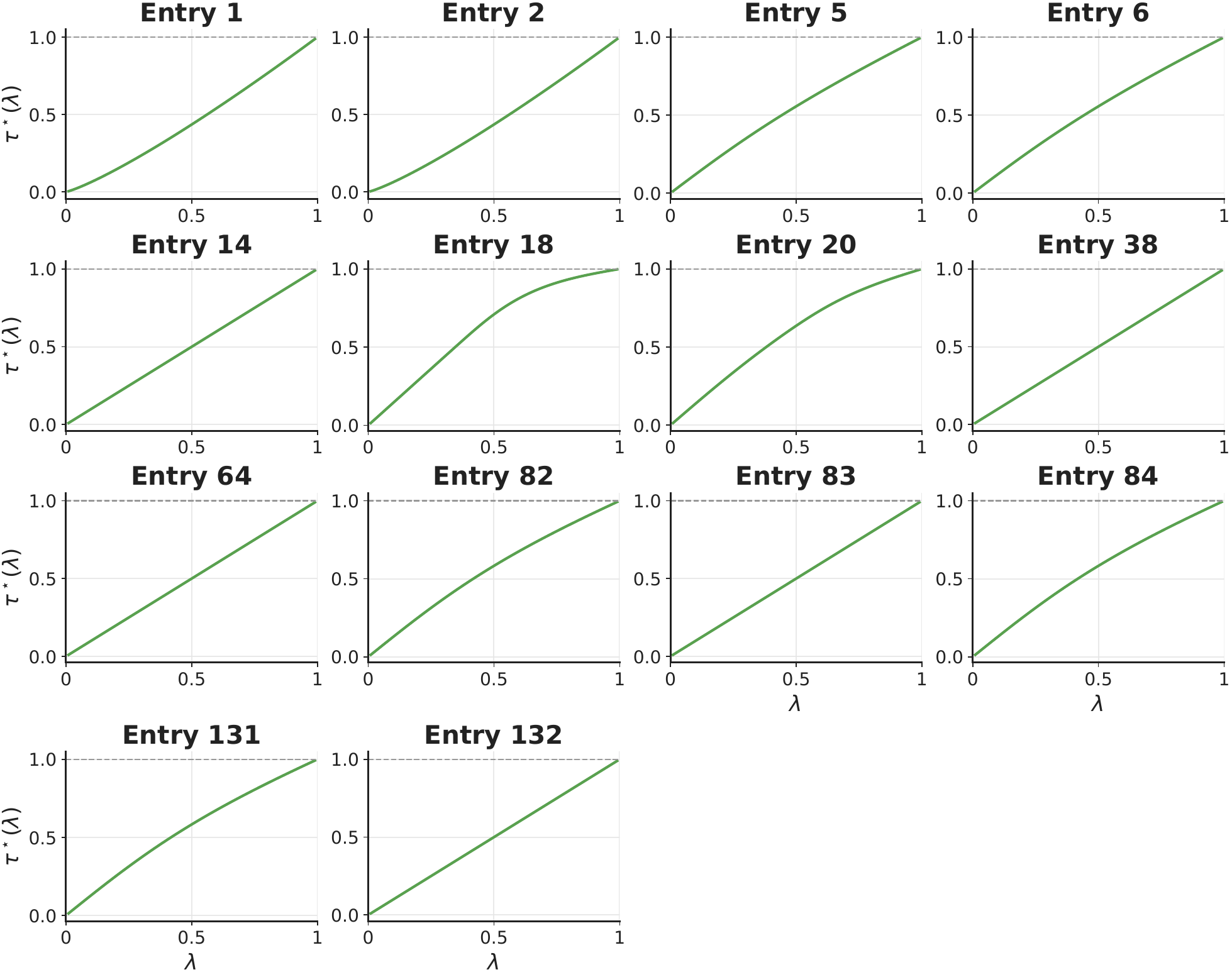}
    \caption{Estimates of \(\tau^\star(\lambda)\) for the entries of Table~\ref{tab:accepted-bernstein-functions}. Entry numbers refer to \citep{schilling2009bernstein}.}
    \label{fig:cbf_accepted_tau_star_curves}
\end{figure}

\begin{figure}[htbp]
    \centering
    \includegraphics[width=0.95\textwidth]{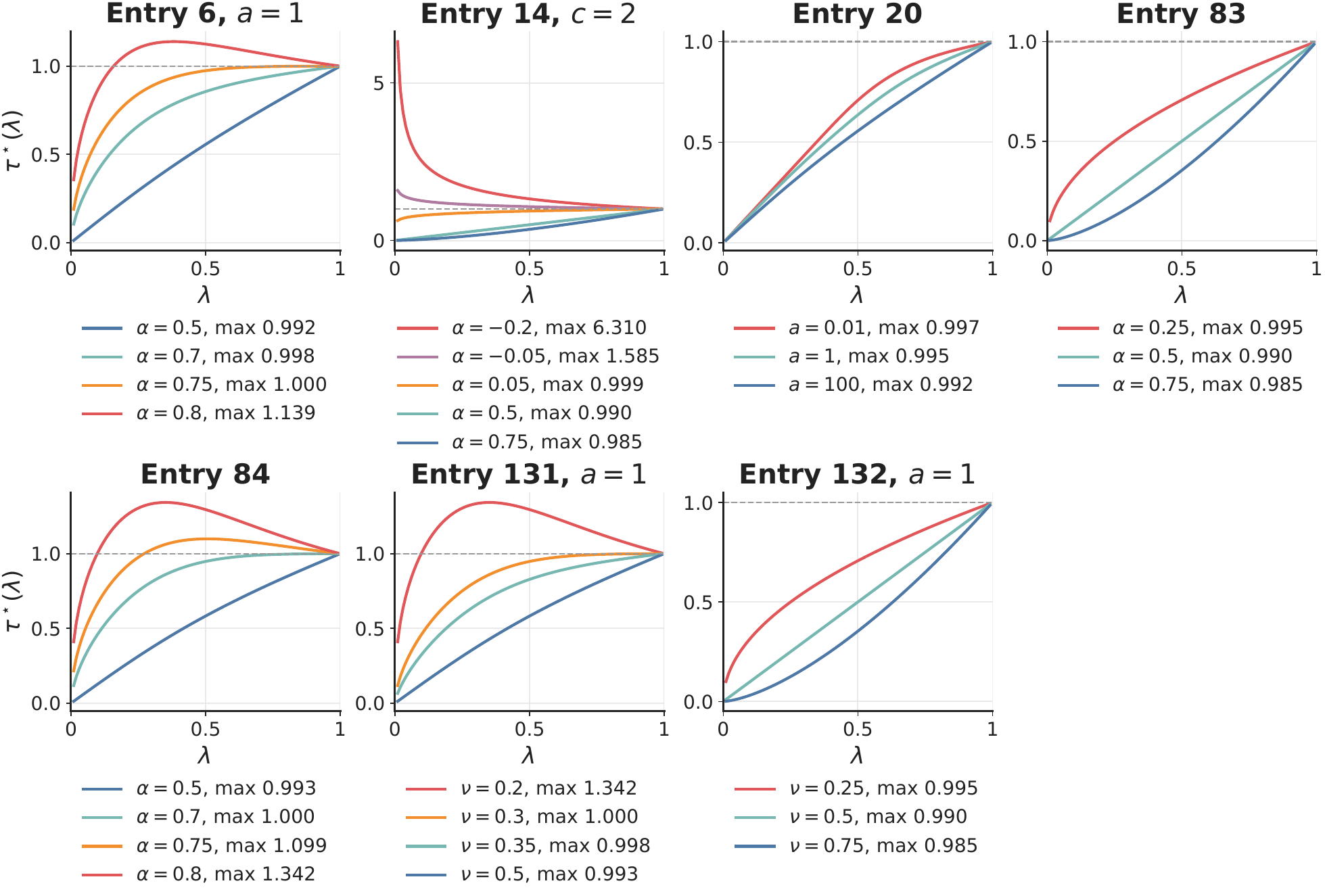}
    \caption{Estimates of \(\tau^\star(\lambda)\) for seven entries with a parameter, at several values of the parameter. An entry induces contraction at a value of its parameter when its curve stays below the dashed line \(\tau^\star(\lambda)=1\). Entry numbers refer to \citep{schilling2009bernstein}.}
    \label{fig:cbf_parametric_regimes}
\end{figure}

\begin{figure}[htbp]
    \centering
    \includegraphics[width=0.95\textwidth]{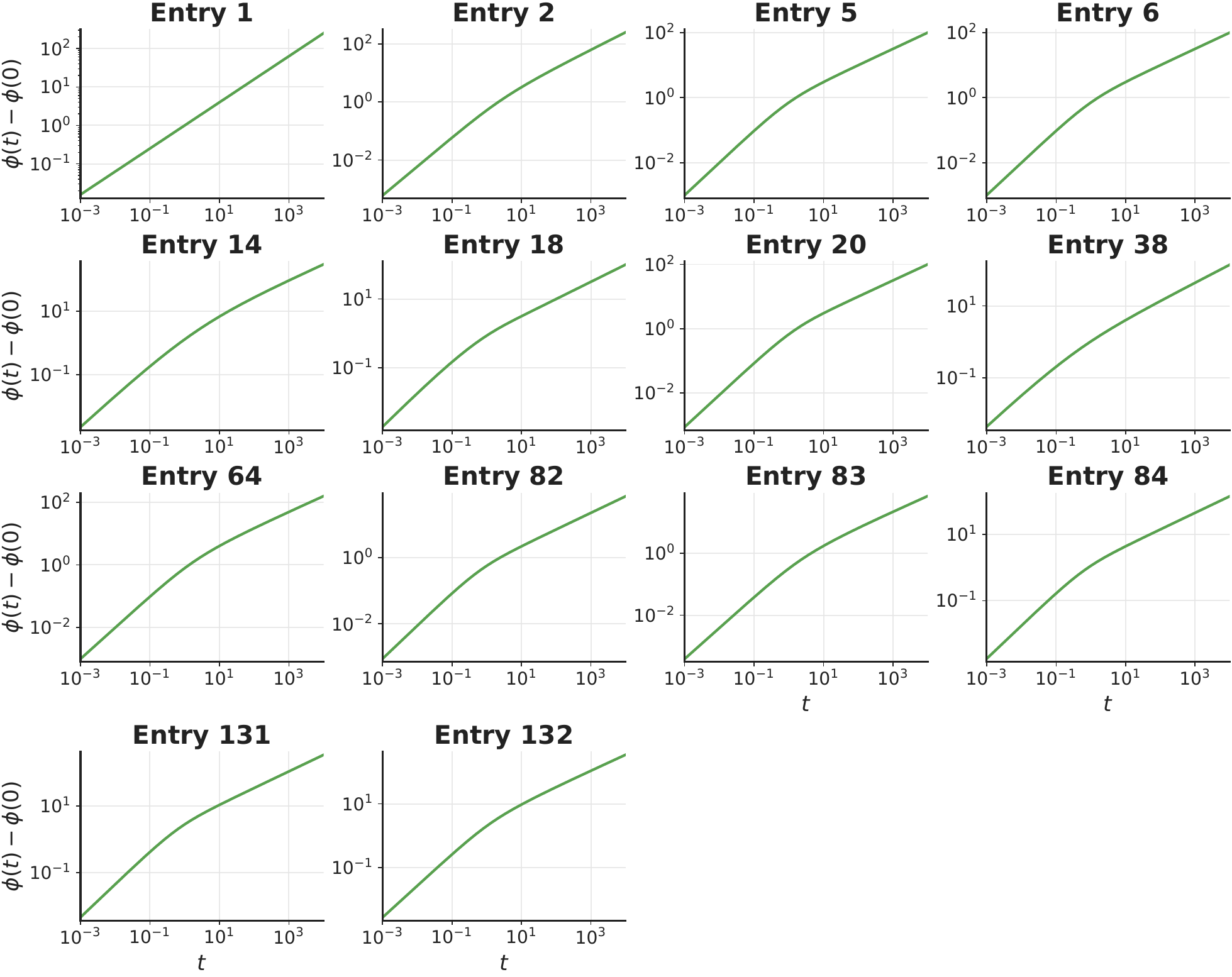}
    \caption{Profiles \(\phi\) of the entries of Table~\ref{tab:accepted-bernstein-functions}, on logarithmic axes. Entry numbers refer to \citep{schilling2009bernstein}.}
    \label{fig:cbf_accepted_profiles}
\end{figure}

We accept an entry when its estimate of \(\tau^\star(\lambda)\) stays below \(1\) for every \(\lambda\), and its best point is never at an end of the window. Table~\ref{tab:accepted-bernstein-functions} lists the accepted entries, and Figures~\ref{fig:cbf_accepted_tau_star_curves} and~\ref{fig:cbf_accepted_profiles} show their estimates of \(\tau^\star(\lambda)\) and their profiles \(\phi\). For an entry with a parameter, these results hold at the parameter values given in the experimental setup. Contraction may then hold only on part of the range allowed in \citep{schilling2009bernstein}. Figure~\ref{fig:cbf_parametric_regimes} shows, for seven such entries, how the estimate approaches or crosses \(1\) as the parameter varies.

\begin{figure}[htbp]
    \centering
    \includegraphics[width=0.95\textwidth]{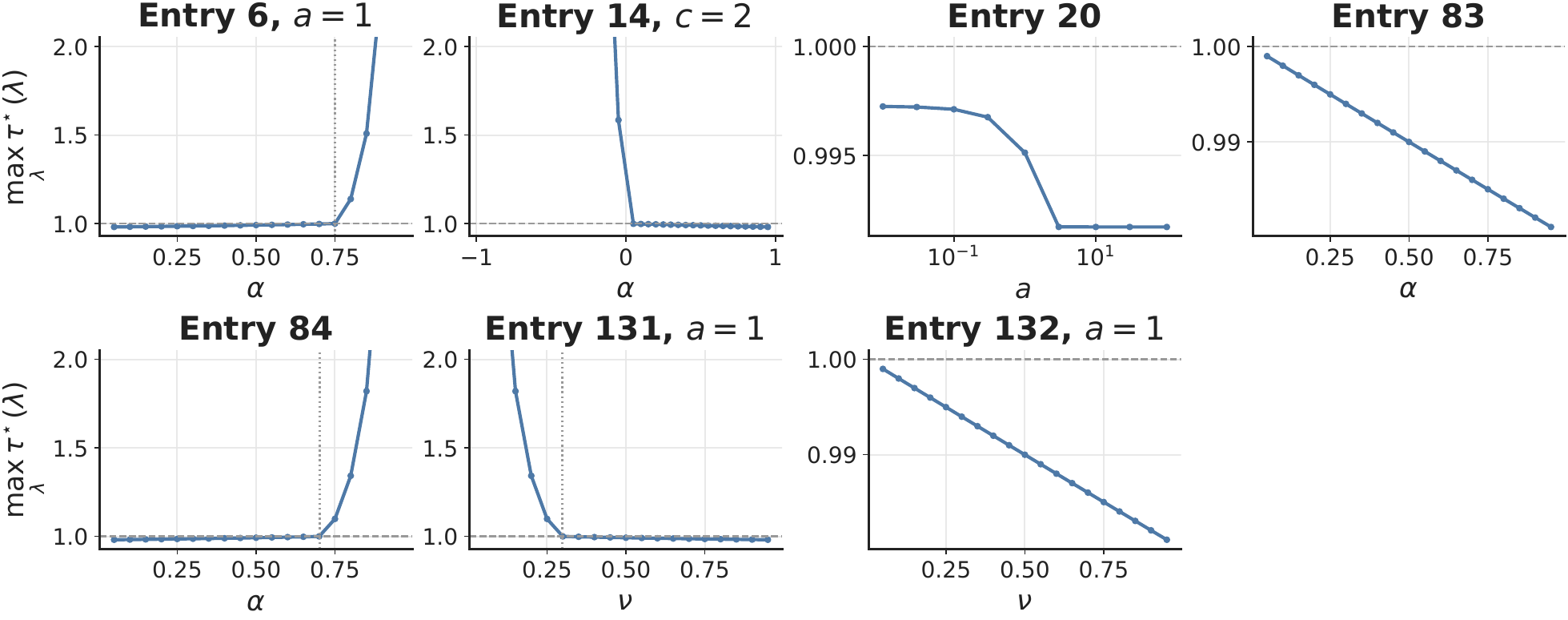}
    \caption{Each panel shows, for one entry of Table~\ref{tab:accepted-bernstein-functions}, the largest estimate of \(\tau^\star(\lambda)\) over \(\lambda\) as a function of the entry's parameter. The entry induces contraction where the curve is below the dashed line. The dotted lines mark where the curve crosses \(1\), found by a root search. Entry numbers refer to \citep{schilling2009bernstein}.}
    \label{fig:cbf_largest_tau_star_by_parameter}
\end{figure}

For seven entries of Table~\ref{tab:accepted-bernstein-functions}, the parameter values giving contraction need no computation. Propositions~\ref{prop:power_distance_contraction} and~\ref{prop:gmq_contraction} cover every value of the parameter of Entries~1 and~2. In Entries~5, 18, 38, 64 and~82, the parameter only changes the scale of \(t\), which leaves \(\tau^\star(\lambda)\) unchanged. For the other seven, Entries~6, 14, 20, 83, 84, 131 and~132, whether the entry induces contraction may depend on the value of its parameter. We therefore ran the search for many values of the parameter across its range. Figure~\ref{fig:cbf_largest_tau_star_by_parameter} shows the largest estimate over \(\lambda\) for each value. When this curve crosses \(1\), a root search on the parameter locates where it does, which gives the range in Table~\ref{tab:accepted-bernstein-functions}.

\begin{table}[htbp]
\centering
\small
\setlength{\tabcolsep}{6pt}
\renewcommand{\arraystretch}{1.2}
\begin{tabular}{r p{10.5cm}}
\hline
Entry & Parameter values giving contraction \\
\hline
1 & every value \\
2 & every value \\
5 & every value \\
6 & \(\alpha\le 3/4\) \\
14 & \(\alpha>0\) \\
18 & every value \\
20 & every value \\
38 & every value \\
64 & every value \\
82 & every value \\
83 & every value \\
84 & \(\alpha\) up to about \(0.70\) \\
131 & \(\nu\) from about \(0.30\) \\
132 & every value \\
\hline
\end{tabular}
\caption{Entries that induce contraction. Entry numbers refer to \citep{schilling2009bernstein}. The second column gives the parameter values for which the entry induces contraction, among those allowed in \citep{schilling2009bernstein}. Entry~1 gives the power-distance kernel studied in \citep{nguyen2021distributional}, and Entry~2 with \(\alpha=1/2\) gives the multiquadric kernel studied in \citep{killingberg2023the}. Entry~5 is Entry~6 with \(\alpha=1/2\).}
\label{tab:accepted-bernstein-functions}
\end{table}

\clearpage

\subsubsection{Entries 3, 4 and 26 in closed form}

\begin{proposition}[No contraction under inverse multiquadric, rational and logarithmic kernels]
\label{prop:three_kernels_no_contraction}
Assume \(f(x)=q+\lambda x\) with \(\lambda\in(0,1)\). For each kernel \(k\) below, with its profile \(\phi\), the stated \(\tau^\star(\lambda)\) is the smallest \(\tau\) such that \(\MMD_k^2(f\#\nu,f\#\mu)\le \tau\,\MMD_k^2(\nu,\mu)\) for every \(d\ge 1\) and every \(\mu,\nu\in\mathcal P(\R^d)\).
\begin{enumerate}
\item[(i)] Inverse multiquadric kernels, \(\alpha\in(0,1)\):
\[
k_\alpha(x,y)=\bigl(1+\|x-y\|^2\bigr)^{\alpha-1},
\qquad
\phi_\alpha(t)=1-(1+t)^{\alpha-1},
\qquad
\tau^\star(\lambda)=\lambda^{2\alpha-2}.
\]
\item[(ii)] Rational kernels, \(a>0\):
\[
k_a(x,y)=\frac{a}{\|x-y\|^2+a},
\qquad
\phi_a(t)=\frac{t}{t+a},
\qquad
\tau^\star(\lambda)=\lambda^{-2}.
\]
\item[(iii)] Logarithmic kernels, \(a>0\):
\[
k_a(x,y)=-\log\!\left(1+\frac{\|x-y\|^2}{a}\right),
\qquad
\phi_a(t)=\log\!\left(1+\frac{t}{a}\right),
\qquad
\tau^\star(\lambda)=1.
\]
\end{enumerate}
For (i) and (ii), \(\tau^\star(\lambda)>1\), so for some \(d\), \(\mu\) and \(\nu\) the map \(f\) increases \(\MMD_k^2\). For (iii), \(\tau^\star(\lambda)=1\), so \(f\) is non-expansive in \(\MMD_{k_a}^2\), that is,
\[
\MMD_{k_a}^2(f\#\nu,f\#\mu)\le \MMD_{k_a}^2(\nu,\mu)
\qquad
\text{for every }d\ge 1\text{ and every }\mu,\nu\in\mathcal P(\R^d),
\]
but \(f\) is not a contraction in \(\MMD_{k_a}^2\).
\end{proposition}

\begin{proof}
\noindent\textbf{L\'evy--Khintchine representations.}
For (i), (ii) and (iii), Entries~3, 4 and~26 of Tables~\ref{tab:cbf_screen_candidates_a}--\ref{tab:cbf_screen_candidates_b} give \(b=0\) and the L\'evy densities
\[
\text{(i)}\ \ m(s)=\frac{e^{-s}s^{-\alpha}}{\Gamma(1-\alpha)},
\qquad
\text{(ii)}\ \ m(s)=a\,e^{-as},
\qquad
\text{(iii)}\ \ m(s)=\frac{e^{-as}}{s},
\qquad s>0.
\]

\noindent\textbf{Application of Theorem~\ref{thm:cbf_contraction_sup}.}
Set \(\rho=\lambda^2\in(0,1)\). Since \(b=0\), Theorem~\ref{thm:cbf_contraction_sup} gives, in each case,
\[
\tau^\star(\lambda)=\sup_{s>0}R_\lambda(s),
\qquad
R_\lambda(s):=\frac{m(s/\rho)}{\rho\,m(s)}.
\]
Using the formula for \(m\) in each case, we obtain
\begin{align*}
\text{(i)}\quad R_\lambda(s)
&=
\frac{1}{\rho}\,
\frac{\Gamma(1-\alpha)^{-1}e^{-s/\rho}(s/\rho)^{-\alpha}}{\Gamma(1-\alpha)^{-1}e^{-s}s^{-\alpha}}
=
\rho^{\alpha-1}\exp\!\left(-s\left(\frac{1}{\rho}-1\right)\right),
\\
\text{(ii)}\quad R_\lambda(s)
&=
\frac{1}{\rho}\,
\frac{a e^{-as/\rho}}{a e^{-as}}
=
\rho^{-1}\exp\!\left(-as\left(\frac{1}{\rho}-1\right)\right),
\\
\text{(iii)}\quad R_\lambda(s)
&=
\frac{1}{\rho}\,
\frac{e^{-as/\rho}}{s/\rho}\,
\frac{s}{e^{-as}}
=
\exp\!\left(-as\left(\frac{1}{\rho}-1\right)\right).
\end{align*}

\noindent\textbf{Suprema of the ratios.}
For (i), (ii) and (iii), we show that the ratios \(R_\lambda\) stay below \(\rho^{\alpha-1}\), \(\rho^{-1}\) and \(1\), respectively, and approach these values as \(s\to0^+\), which makes these values their suprema.

Since \(0<\rho<1\), the factor \(1/\rho-1\) in each exponent above is positive. Each exponent is therefore a negative constant times \(s\), so it is negative for every \(s>0\) and tends to \(0\) as \(s\to0^+\). Each exponential thus lies strictly between \(0\) and \(1\) for every \(s>0\), and tends to \(1\) as \(s\to0^+\).

In the expressions above, the ratios \(R_\lambda(s)\) are these exponentials multiplied by \(\rho^{\alpha-1}\) for (i), by \(\rho^{-1}\) for (ii) and by \(1\) for (iii). Hence, for every \(s>0\),
\[
\text{(i)}\ \ 0<R_\lambda(s)<\rho^{\alpha-1},
\qquad
\text{(ii)}\ \ 0<R_\lambda(s)<\rho^{-1},
\qquad
\text{(iii)}\ \ 0<R_\lambda(s)<1.
\]
The ratios also tend to their upper bounds as \(s\to0^+\). Therefore, these upper bounds are their suprema, and Theorem~\ref{thm:cbf_contraction_sup} yields
\[
\text{(i)}\ \ \tau^\star(\lambda)=\rho^{\alpha-1}=\lambda^{2\alpha-2},
\qquad
\text{(ii)}\ \ \tau^\star(\lambda)=\rho^{-1}=\lambda^{-2},
\qquad
\text{(iii)}\ \ \tau^\star(\lambda)=1.
\]

\noindent\textbf{Conclusions.}
For (i), since \(\alpha\in(0,1)\), we have \(2\alpha-2<0\), hence \(\tau^\star(\lambda)=\lambda^{2\alpha-2}>1\) for every \(0<\lambda<1\). For (ii), \(\tau^\star(\lambda)=\lambda^{-2}>1\) for every \(0<\lambda<1\).
\end{proof}

\subsection{From geometric chaining}\label{app:geometric_chains}
\begin{proposition}[Geometric chaining builds kernels with prescribed contraction factor]
\label{prop:kernel_geometric_chaining_factory}
Fix \(\lambda\in(0,1)\) and set \(\rho:=\lambda^2\in(0,1)\). Let \(\tau\in(0,1)\), let \(\phi_0\) be a Bernstein function on \([0,\infty)\) with \(\phi_0(0)=0\), and define
\[
h_0(r):=\phi_0(r^2),
\qquad r\ge 0.
\]
Assume that the chained profile
\[
\phi(t):=\sum_{n=0}^\infty \tau^{-n}\phi_0(\rho^n t),
\qquad t\ge 0,
\]
is well defined in the sense of Proposition~\ref{prop:geometric_chaining_admissibility}. Define its radial profile
\[
h_{\mathrm{chain}}(r):=\phi(r^2),
\qquad r\ge 0,
\]
and, for an arbitrary constant \(C\in\R\),
\[
k_{\mathrm{chain}}(x,y):=C-h_{\mathrm{chain}}(\|x-y\|),
\qquad x,y\in\R^d.
\]
Then, for every \(d\ge 1\), every \(q\in\R^d\) and every affine map \(f(x)=q+\lambda x\), one has
\[
\boxed{%
\MMD_{k_{\mathrm{chain}}}^2(f\#\nu,f\#\mu)
\le
\tau\,\MMD_{k_{\mathrm{chain}}}^2(\nu,\mu)
\qquad
\forall \mu,\nu\in\mathcal P(\R^d).
}
\]
Moreover, the chained radial profile is given by
\[
\boxed{%
h_{\mathrm{chain}}(r)
=
\sum_{n=0}^\infty \tau^{-n}h_0(\lambda^n r),
\qquad r\ge 0,
}
\]
and satisfies the exact identity
\[
\boxed{%
\tau\,h_{\mathrm{chain}}(r)-h_{\mathrm{chain}}(\lambda r)=\tau\,h_0(r),
\qquad r\ge 0.
}
\]
Thus, whenever the chained profile is well defined, every such affine map \(f\) is a \(\tau\)-contraction in \(\MMD^2\) under the radial kernel it produces.
\end{proposition}

\begin{proof}
By Proposition~\ref{prop:geometric_chaining_admissibility}, the chained profile \(\phi\) defines a Bernstein function. Therefore Theorem~\ref{thm:general_geometric_chaining_factory}
applies and yields
\[
\MMD_{k_{\mathrm{chain}}}^2(f\#\nu,f\#\mu)
\le
\tau\,\MMD_{k_{\mathrm{chain}}}^2(\nu,\mu)
\qquad
\forall d\ge 1,\ \forall q\in\R^d,\ \forall \mu,\nu\in\mathcal P(\R^d),
\]
for \(f(x)=q+\lambda x\).

It remains to identify the chained radial profile. Since \(\rho=\lambda^2\),
\[
h_{\mathrm{chain}}(r)
=
\phi(r^2)
=
\sum_{n=0}^\infty \tau^{-n}\phi_0(\rho^n r^2)
=
\sum_{n=0}^\infty \tau^{-n}\phi_0((\lambda^n r)^2)
=
\sum_{n=0}^\infty \tau^{-n}h_0(\lambda^n r).
\]
Finally, Proposition~\ref{prop:exact_geometric_chaining} gives
\[
\tau\,\phi(t)-\phi(\rho t)=\tau\,\phi_0(t)
\qquad\forall t\ge 0.
\]
Substituting \(t=r^2\) and using \(\rho r^2=(\lambda r)^2\) yields
\[
\tau\,h_{\mathrm{chain}}(r)-h_{\mathrm{chain}}(\lambda r)=\tau\,h_0(r),
\qquad r\ge 0.
\]
\end{proof}

\begin{proposition}[The chained RBF kernel realizes any prescribed smallest contraction factor across dimensions]
\label{prop:gaussian_geometric_chaining_series}
Fix \(\lambda\in(0,1)\) and set \(\rho:=\lambda^2\in(0,1)\). Let \(\tau\in(\rho,1)\), \(a>0\), and \(c>0\), and define
\[
\phi_{\mathrm{chain}}(t)
:=
\sum_{n=0}^{\infty} c\,\tau^{-n}\bigl(1-e^{-a\rho^n t}\bigr),
\qquad
h_{\mathrm{chain}}(r):=\phi_{\mathrm{chain}}(r^2).
\]
For an arbitrary constant \(C\in\R\), let
\[
k_{\mathrm{chain}}(x,y):=C-h_{\mathrm{chain}}(\|x-y\|),
\qquad x,y\in\R^d.
\]

Then, for every \(d\ge 1\), every \(q\in\R^d\), and every affine map
\[
f(x)=q+\lambda x,
\]
one has
\[
\MMD_{k_{\mathrm{chain}}}^2(f\#\nu,f\#\mu)
\le
\tau\,\MMD_{k_{\mathrm{chain}}}^2(\nu,\mu)
\qquad
\forall \mu,\nu\in\mathcal P(\R^d).
\]
Moreover, \(\tau\) is the smallest contraction factor for \(k_{\mathrm{chain}}\), namely
\[
\tau^\star(\lambda)=\tau.
\]
Thus, for every prescribed factor \(\tau\in(\lambda^2,1)\), this construction yields a kernel whose smallest contraction factor across dimensions is exactly \(\tau\). For the chained RBF kernel, smaller choices of \(\tau\) lead to faster contraction, and
\[
\inf_{\tau>\lambda^2}\tau^\star(\lambda)=\lambda^2,
\]
although this value is not attained since the construction requires \(\tau>\lambda^2\).

If we introduce the lengthscales
\[
\ell_n:=\frac{1}{\sqrt{a\rho^n}}
=
\frac{1}{\sqrt a}\,\lambda^{-n},
\qquad n\ge 0,
\]
then \(a\rho^n=1/\ell_n^2\), and
\[
h_{\mathrm{chain}}(r)
=
\sum_{n=0}^{\infty} c\,\tau^{-n}
\Bigl(1-\exp\bigl(-r^2/\ell_n^2\bigr)\Bigr),
\qquad r\ge 0.
\]
Thus the geometric rescaling \(r\mapsto \lambda^n r\) can be viewed as a series of RBF kernels with geometrically increasing lengthscales \((\ell_n)_{n\ge 0}\).
\end{proposition}

\begin{proof}
\noindent\paragraph{Smallest contraction factor.}
The base Bernstein function
\[
\phi_0(t):=c\bigl(1-e^{-at}\bigr)
\]
has zero drift and a L\'evy measure with a single atom, \(c\,\delta_a\). Thus the present construction is exactly the single geometric chain of atoms from Proposition~\ref{prop:single_atomic_chain}, with \(s_0=a\). Since \(\tau>\rho=\lambda^2\), that proposition shows that \(\phi_{\mathrm{chain}}\) is a Bernstein function and that, under the induced radial kernel, the affine map \(x\mapsto q+\lambda x\) is a \(\tau\)-contraction in \(\MMD^2\). In particular, for the kernel designed with parameter \(\tau\), this yields
\[
\tau^\star(\lambda)\le \tau.
\]
This is also the RBF case of the general chaining results, namely Propositions~\ref{prop:exact_geometric_chaining} and~\ref{prop:geometric_chaining_admissibility} together with Proposition~\ref{prop:kernel_geometric_chaining_factory}.

It remains to prove
\[
\tau^\star(\lambda)\ge \tau,
\]
since together with the already established \(\tau^\star(\lambda)\le \tau\), this yields
\[
\tau^\star(\lambda)=\tau.
\]
Let \(\bar\tau>0\) be any contraction factor in \(\MMD^2_{k_{\mathrm{chain}}}\). Write
\[
s_n:=\rho^n a,
\qquad
\Lambda(ds)=\sum_{n=0}^{\infty} c\,\tau^{-n}\,\delta_{s_n}(ds).
\]
By condition 3 of Theorem~\ref{thm:contraction-drift-density-atoms}, one must have
\[
\bar\tau\,\Lambda(\{x\})\ge \Lambda(\{x/\rho\})
\qquad\text{for every }x>0.
\]
Applying this with \(x=s_n\) for \(n\ge 1\) gives
\[
\bar\tau\,\Lambda(\{s_n\})\ge \Lambda(\{s_{n-1}\}).
\]
Since
\[
\Lambda(\{s_n\})=c\,\tau^{-n},
\qquad
\Lambda(\{s_{n-1}\})=c\,\tau^{-(n-1)},
\]
it follows that
\[
\bar\tau\,c\,\tau^{-n}\ge c\,\tau^{-(n-1)},
\]
hence
\[
\bar\tau\ge \tau.
\]
Therefore \(\tau^\star(\lambda)\ge \tau\), and so \(\tau^\star(\lambda)=\tau\). Taking the infimum over \(\tau>\lambda^2\) gives
\[
\inf_{\tau>\lambda^2}\tau^\star(\lambda)=\lambda^2.
\]

\noindent\paragraph{Explicit form as a series of RBF kernels.}
By definition,
\[
\phi_{\mathrm{chain}}(t)
=
\sum_{n=0}^{\infty} c\,\tau^{-n}\bigl(1-e^{-a\rho^n t}\bigr),
\]
and therefore
\[
h_{\mathrm{chain}}(r)
=
\phi_{\mathrm{chain}}(r^2)
=
\sum_{n=0}^{\infty} c\,\tau^{-n}\Bigl(1-e^{-a\rho^n r^2}\Bigr).
\]
Introducing
\[
\ell_n=\frac{1}{\sqrt{a\rho^n}}=\frac{1}{\sqrt a}\lambda^{-n}
\]
gives \(a\rho^n=1/\ell_n^2\), so
\[
h_{\mathrm{chain}}(r)
=
\sum_{n=0}^{\infty} c\,\tau^{-n}
\Bigl(1-\exp\bigl(-r^2/\ell_n^2\bigr)\Bigr).
\]
Finally,
\[
k_{\mathrm{chain}}(x,y)
=
C-
\sum_{n=0}^{\infty} c\,\tau^{-n}
\Bigl(1-\exp\bigl(-\|x-y\|^2/\ell_n^2\bigr)\Bigr),
\]
which is the form used in Theorem~\ref{thm:main_gaussian_chain}.
\end{proof}

\begin{figure}[H]
\centering
\begin{minipage}[t]{0.48\linewidth}
  \centering
  \includegraphics[width=\linewidth]{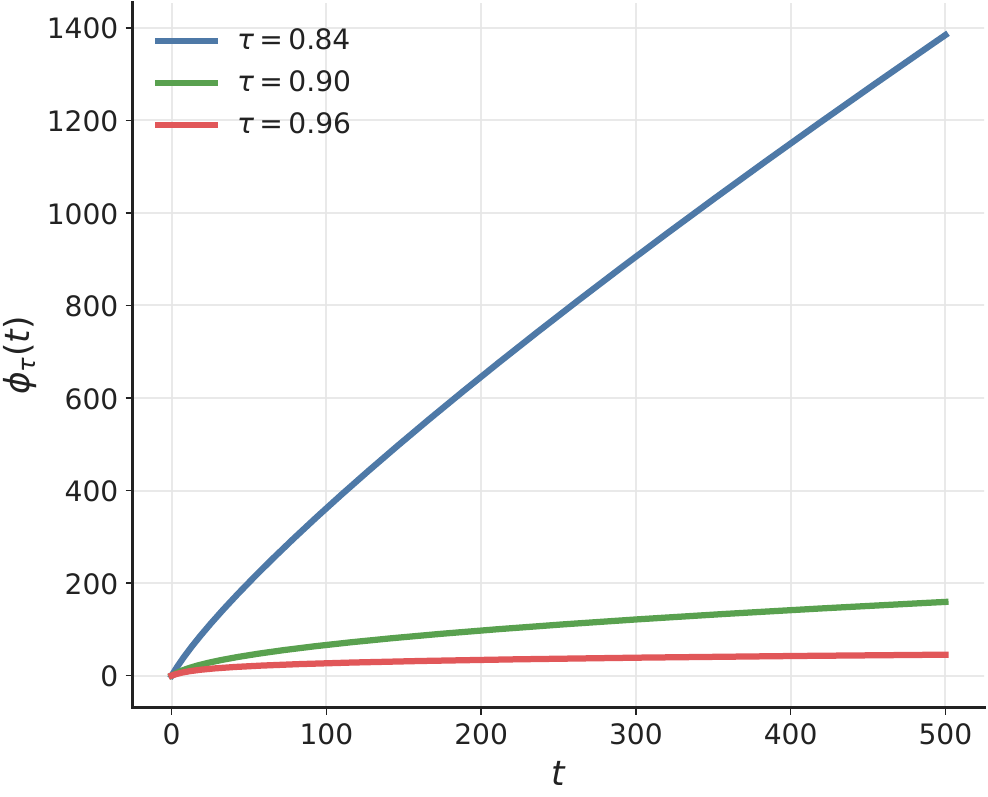}
  \caption{Bernstein functions \(\phi_\tau(t)=\sum_{n=0}^{K}c\,\tau^{-n}(1-e^{-a\gamma^{2n}t})\) for \(\tau\in\{0.84,0.90,0.96\}\), truncated at \(K=700\) with \(\gamma=0.90\), \(a=0.20\).}
  \label{fig:gaussian_chain_tau_bernstein}
\end{minipage}
\hfill
\begin{minipage}[t]{0.48\linewidth}
  \centering
  \includegraphics[width=\linewidth]{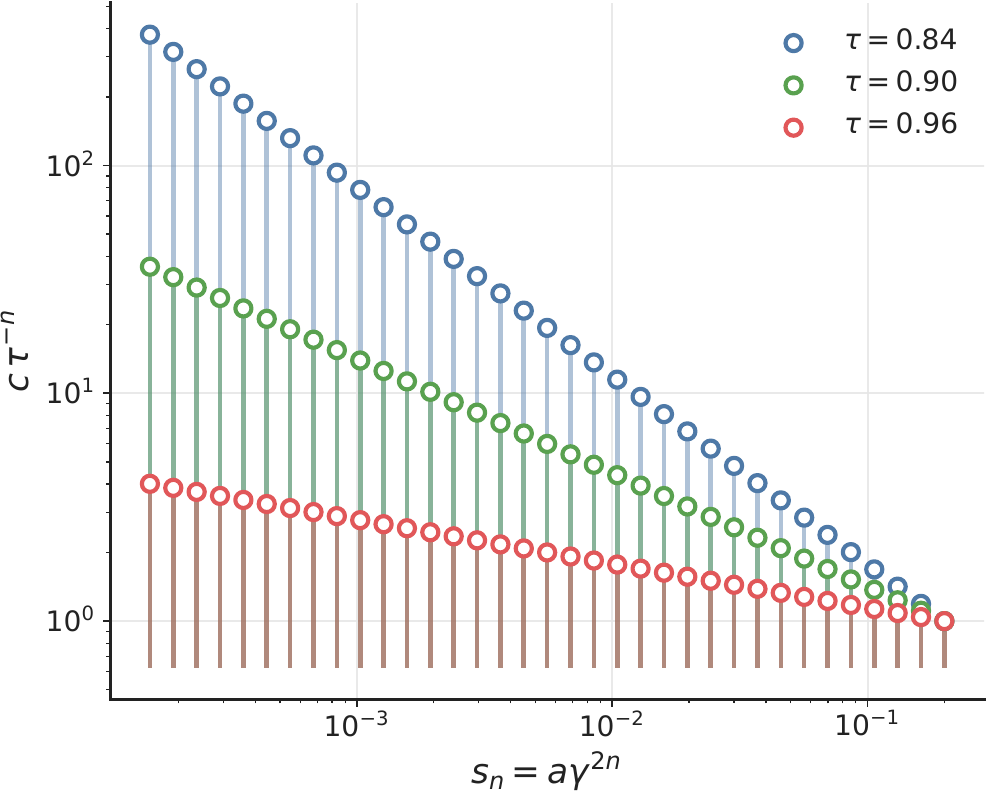}
  \caption{Lévy measure atoms \((s_n, c\,\tau^{-n})\) with \(s_n=a\gamma^{2n}\), first \(35\) atoms shown on a log-log scale.}
  \label{fig:gaussian_chain_tau_levy}
\end{minipage}
\end{figure}

\begin{figure}[t]
    \centering

    \begin{minipage}[t]{0.48\textwidth}
        \centering
        \includegraphics[width=\linewidth]{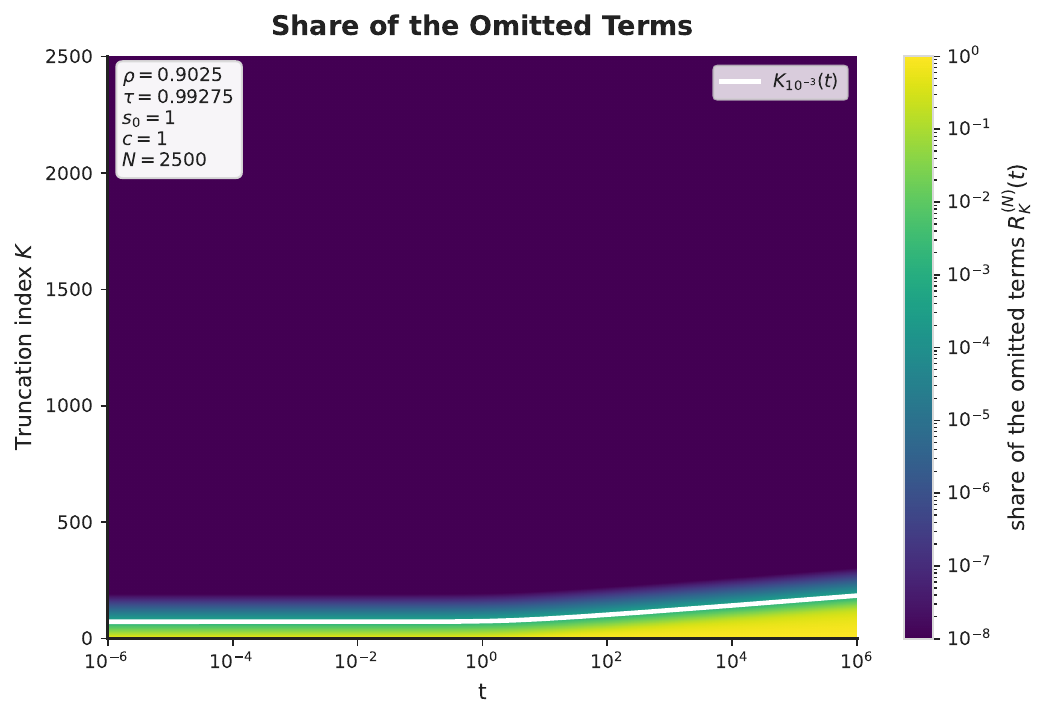}
        \caption{Share of the omitted terms \(R_K^{(N)}(t)\) for the chained RBF kernel. The white curve shows the truncation threshold \(K_{10^{-3}}(t)\), beyond which the omitted terms account for at most \(10^{-3}\) of the approximated radial profile.}
        \label{fig:chain_tail_share}
    \end{minipage}
    \hfill
    \begin{minipage}[t]{0.48\textwidth}
        \centering
        \includegraphics[width=\linewidth]{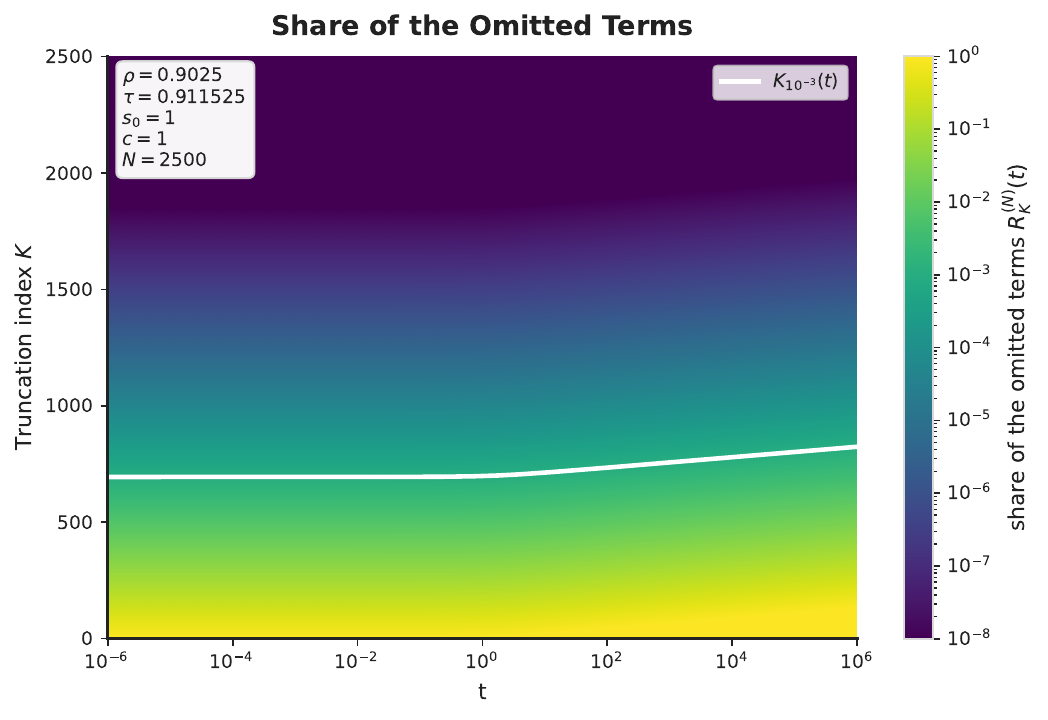}
        \caption{The same share of the omitted terms when \(\rho/\tau\) is closer to one. The white curve shifts upward, showing that more terms are needed before the omitted terms account for less than \(10^{-3}\) of the approximated radial profile.}
        \label{fig:chain_tail_close}
    \end{minipage}
\end{figure}
\begin{table}[t]
\centering
\caption{Contraction of truncated chains on a ball, for the two chains of Figures~\ref{fig:chain_tail_share} and~\ref{fig:chain_tail_close}, with \(\rho=0.9025\), that is \(\gamma=0.95\), and \(s_0=c=1\). The test of Section~\ref{sec:bounded_support} uses \(10\) sets of \(512\) points drawn at random in the ball of radius \(1\), in dimension~\(1\), the same for every number of terms. The estimate is the largest contraction threshold over these sets. The full chain is computed with \(2500\) terms. We compute \(R_K^{(N)}(t)\) with \(N=5000\), for every number of terms and squared distances up to \(4\). The test is run with every number of terms up to \(40\), then with \(50\), \(60\), \(80\), \(100\), \(150\), \(200\), \(300\), \(400\), \(500\), \(700\), \(1000\) and \(1500\) terms.}
\label{tab:truncated_chain_contraction}
\begin{tabular}{lcc}
\toprule
 & \(\tau=0.99275\) & \(\tau=0.911525\) \\
\midrule
Estimated smallest contraction factor of the full chain & \(0.964\) & \(0.910\) \\
Terms needed to pass the test & \(13\) & \(12\) \\
Terms needed to be within \(10^{-3}\) of the full chain & \(60\) & \(300\) \\
Terms needed for \(R_K^{(N)}(t)\le 10^{-3}\) on the ball & \(80\) & \(706\) \\
\bottomrule
\end{tabular}
\end{table}

\paragraph{Numerical estimation.}\label{app:chaining_numerical}

To use the geometric chaining construction as an MMD kernel, the infinite series has to be approximated numerically and therefore truncated. Consider the Bernstein function of the chained RBF kernel
\[
\phi(t)
=
\sum_{n\ge 0}
c\,\tau^{-n}
\bigl(1-\exp(-s_0\rho^n t)\bigr),
\qquad
0<\rho<\tau<1,
\]
and write its \(n\)-th term as
\[
\phi_n(t)
:=
c\,\tau^{-n}
\bigl(1-\exp(-s_0\rho^n t)\bigr).
\]
For a truncation index \(K\), we measure the relative contribution of the omitted terms by
\[
R_K(t)
:=
\frac{
\sum_{n\ge K} \phi_n(t)
}{
\sum_{n\ge 0} \phi_n(t)
}.
\]
In practice, we approximate this quantity by replacing the infinite denominator and numerator with a sufficiently large finite cutoff \(N\),
\[
R_K^{(N)}(t)
:=
\frac{
\sum_{K\le n\le N} 
c\,\tau^{-n}
\bigl(1-\exp(-s_0\rho^n t)\bigr)
}{
\sum_{0\le n\le N}
c\,\tau^{-n}
\bigl(1-\exp(-s_0\rho^n t)\bigr)
},
\qquad N\gg 1.
\]
Given a tolerance level \(\alpha>0\), we define the truncation threshold
\[
K_\alpha(t)
:=
\min\left\{
K\in\mathbb N
\mid
R_K^{(N)}(t)\le \alpha
\right\}.
\]
In the numerical results below, we use \(\alpha=10^{-3}\) and report the corresponding threshold \(K_{10^{-3}}(t)\).

As shown in Figure~\ref{fig:chain_tail_share}, the terms of the chain decay rapidly. For each fixed \(t\), there is a finite index \(K_{10^{-3}}(t)\) beyond which the omitted terms account for less than \(0.1\%\) of the approximated radial profile. Figure~\ref{fig:chain_tail_close} shows the same diagnostic when \(\rho/\tau\) is closer to one, where the share of the omitted terms decreases more slowly as \(K\) grows and a larger truncation index is needed to reach the same \(0.1\%\) threshold.

\paragraph{Contraction of truncated chains.}
A truncated chain is a finite sum of RBF kernels, so by Proposition~\ref{prop:main_finite_atomic_obstruction} it does not induce contraction on all of \(\R^d\). As discussed in Section~\ref{sec:bounded_support}, it can still induce contraction on a bounded set, which we check with the test detailed in Appendix~\ref{sec:bounded_support_numerical_experiments} for the two chains of Figures~\ref{fig:chain_tail_share} and~\ref{fig:chain_tail_close} (Table~\ref{tab:truncated_chain_contraction}). About a dozen terms are enough for the truncated chain to pass the test. With more terms, its estimated smallest contraction factor on the ball decreases towards that of the full chain. This limit is below \(\tau\), since contraction on a ball is a weaker requirement than on all of \(\R^d\). Truncating with the rule above, which makes the profile accurate to \(10^{-3}\) on the ball, already brings the estimated contraction factor of the truncated chain within \(10^{-3}\) of that of the full chain. Both take many more terms when \(\rho/\tau\) is closer to one.

\clearpage
\section{Contraction rate and discriminability}\label{app:tradeoff_detectability}
\paragraph{Preamble.}
A kernel with a smaller contraction factor is not always the better choice, because changing the kernel also changes which differences between two distributions the MMD can detect. This section studies this tension, which Section~\ref{sec:tradeoff} introduces. It starts with two results on the kernels \(k(x,y)=C-b\|x-y\|^2\), whose Bernstein function \(\phi(t)=bt\) has a zero L\'evy measure. First, Proposition~\ref{prop:pure_drift_identities} shows that such a kernel only compares the means of two distributions. Second, Proposition~\ref{prop:nonzero_levy_strictly_above_drift} shows that only these kernels can reach the smallest possible contraction factor \(\gamma^2\). The rest of the section describes the experiments. Section~\ref{app:permutation_test_power} presents the permutation test we use to measure the power of a kernel, that is, how often it detects that two distributions differ. Section~\ref{app:tradeoff_sweep} then explains how we estimate the tradeoff between this power and the contraction factor. Finally, Section~\ref{app:tradeoff_dimension} repeats these experiments in higher dimensions.

\begin{proposition}[The kernel \(C-b\|x-y\|^2\) compares only means]
\label{prop:pure_drift_identities}
Let \(b\ge 0\) and \(C\in\R\), and consider the kernel
\[
k(x,y):=C-b\|x-y\|^2,
\qquad x,y\in\R^d.
\]

\textbf{Empirical level.}
Given samples \(x_{1:n}=(x_1,\dots,x_n)\) and \(y_{1:m}=(y_1,\dots,y_m)\), define the empirical means
\[
\bar x_n:=\frac{1}{n}\sum_{i=1}^n x_i,
\qquad
\bar y_m:=\frac{1}{m}\sum_{j=1}^m y_j.
\]
Then the biased empirical estimator satisfies
\[
\boxed{
\widehat{\MMD}_{k,\mathrm{b}}^2(x_{1:n},y_{1:m})
=
2b\|\bar x_n-\bar y_m\|^2
}.
\]
In particular, the biased empirical \(\MMD^2\) depends only on the empirical means of the two samples.

\textbf{Population level.}
Similarly, at the population level, let \(P,Q\in\mathcal P(\R^d)\) have finite second moments. Then
\[
\boxed{
\MMD_k^2(P,Q)
=
2b\|\E_P[X]-\E_Q[X]\|^2
}.
\]
In particular, \(\MMD_k^2(P,Q)=0\) whenever \(P\) and \(Q\) have the same first moment.
\end{proposition}

\begin{proof}
\textbf{Empirical identity.}
Let \(N:=n+m\), define the aggregate sample
\[
z_i:=x_i \quad (1\le i\le n),
\qquad
z_{n+j}:=y_j \quad (1\le j\le m),
\]
and let
\[
\ell
:=
\begin{pmatrix}
\frac{1}{n}\mathbf 1_n\\[2pt]
-\frac{1}{m}\mathbf 1_m
\end{pmatrix}
\in\R^N,
\qquad\text{so that }\mathbf 1_N^\top \ell=0.
\]
Let \(K\in\R^{N\times N}\) be the Gram matrix of the aggregate sample, \(K_{ij}:=k(z_i,z_j)\). By Section~\ref{app:mmd_gram_matrices},
\[
\widehat{\MMD}_{k,\mathrm{b}}^2(x_{1:n},y_{1:m})
=
\ell^\top K\ell.
\]
Using \(k(z_i,z_j)=C-b\|z_i-z_j\|^2\), we obtain
\[
\ell^\top K\ell
=
C(\mathbf 1_N^\top \ell)^2
-
b\sum_{i,j=1}^N \ell_i\ell_j\|z_i-z_j\|^2
=
-b\sum_{i,j=1}^N \ell_i\ell_j\|z_i-z_j\|^2,
\]
since \(\mathbf 1_N^\top \ell=0\). Applying Lemma~\ref{lem:euclidean_squared_distance_identity} then yields
\[
\widehat{\MMD}_{k,\mathrm{b}}^2(x_{1:n},y_{1:m})
=
2b\Bigl\|\sum_{i=1}^N \ell_i z_i\Bigr\|^2.
\]
Finally,
\[
\sum_{i=1}^N \ell_i z_i
=
\frac{1}{n}\sum_{i=1}^n x_i-\frac{1}{m}\sum_{j=1}^m y_j
=
\bar x_n-\bar y_m,
\]
which gives the result.

\medskip
\textbf{Population identity.}
Let \(\sigma:=P-Q\), so that \(\sigma(\R^d)=0\). By the definition of \(\MMD_k^2\) as a quadratic form,
\[
\MMD_k^2(P,Q)
=
\iint_{\R^d\times\R^d} k(x,y)\,d\sigma(x)\,d\sigma(y).
\]
Expanding the kernel gives
\[
k(x,y)=C-b\|x\|^2-b\|y\|^2+2b\,x^\top y.
\]
Since \(\sigma(\R^d)=0\), the constant term and the terms that depend on only one of \(x\) and \(y\) vanish in the quadratic form,
\[
\iint C\,d\sigma(x)\,d\sigma(y)=C\,\sigma(\R^d)^2=0,
\]
and
\[
\iint \|x\|^2\,d\sigma(x)\,d\sigma(y)
=
\Bigl(\int \|x\|^2\,d\sigma(x)\Bigr)\sigma(\R^d)
=
0,
\]
with the same identity for the \(\|y\|^2\) term. Therefore,
\[
\MMD_k^2(P,Q)
=
2b\iint x^\top y\,d\sigma(x)\,d\sigma(y)
=
2b\Bigl\|\int x\,d\sigma(x)\Bigr\|^2.
\]
Since \(\int x\,d\sigma(x)=\E_P[X]-\E_Q[X]\), the claim follows.
\end{proof}

\begin{proposition}[Only the kernels \(C-b\|x-y\|^2\) reach the factor \(\gamma^2\)]\label{prop:nonzero_levy_strictly_above_drift}
Let \(k(x,y)=C-\phi(\|x-y\|^2)\) for \(x,y\in\R^d\), where \(\phi\) is a Bernstein function with Lévy measure \(\Lambda\). If \(\Lambda\not\equiv 0\), then \(\tau_k^\star(\mathcal T^\pi)>\gamma^2\). Equivalently, the value \(\gamma^2\) can only be reached by a kernel of the form \(k(x,y)=C-b\|x-y\|^2\).
\end{proposition}
\begin{proof}
\leavevmode\par
\textbf{A lower bound from two point masses.}
We first derive a lower bound on every contraction factor. The second part then shows that this bound exceeds \(\gamma^2\) when \(\Lambda\not\equiv 0\).

A constant in \(\phi\) only shifts \(C\), so we take \(\phi(0)=0\). For two point masses, this gives \(\MMD_k^2(\delta_a,\delta_b)=2\phi(\|a-b\|^2)\). Take the one-state MDP of Lemma~\ref{lem:deterministic_bellman_embedding} with reward \(0\), whose Bellman operator multiplies every return by \(\gamma\). It has a single state-action pair, so the supremum in \(\overline{\MMD}_k^2\) is over that pair alone.

Fix \(x\in\R^d\) with \(x\neq 0\), let \(t=\|x\|^2\), and let \(\eta_1\) and \(\eta_2\) put the point masses \(\delta_x\) and \(\delta_0\) on that pair. Before a Bellman step,
\[
\overline{\MMD}_k^2(\eta_1,\eta_2)=\MMD_k^2(\delta_x,\delta_0)=2\phi(t).
\]
After the step, \(\mathcal T^\pi\eta_1\) and \(\mathcal T^\pi\eta_2\) put \(\delta_{\gamma x}\) and \(\delta_0\) on the pair, so
\[
\overline{\MMD}_k^2(\mathcal T^\pi\eta_1,\mathcal T^\pi\eta_2)=\MMD_k^2(\delta_{\gamma x},\delta_0)=2\phi(\gamma^2 t).
\]
If \(\mathcal T^\pi\) is a \(\tau\)-contraction for this MDP, the definition of contraction in Section~\ref{sec:background} applied to \(\eta_1\) and \(\eta_2\) gives
\[
2\phi(\gamma^2 t)=\overline{\MMD}_k^2(\mathcal T^\pi\eta_1,\mathcal T^\pi\eta_2)\le\tau\,\overline{\MMD}_k^2(\eta_1,\eta_2)=2\tau\,\phi(t),
\]
so \(\phi(\gamma^2 t)\le\tau\,\phi(t)\).
The factor \(\tau_k^\star(\mathcal T^\pi)\) is the smallest one valid for every MDP, so it is valid for this one as well. Hence, whenever \(\phi(t)>0\),
\[
\tau_k^\star(\mathcal T^\pi)\ge\frac{\phi(\gamma^2 t)}{\phi(t)}.
\]

\textbf{The bound exceeds \(\gamma^2\) when \(\Lambda\not\equiv 0\).}
In the Lévy--Khintchine representation of \(\phi\), the terms in \(b\) cancel,
\begin{align*}
\phi(\gamma^2 t)-\gamma^2\phi(t)
&=\bigl(b\gamma^2 t-\gamma^2 bt\bigr)
+\int_{(0,\infty)}\bigl(1-e^{-\gamma^2 st}\bigr)\,\Lambda(ds)-\gamma^2\int_{(0,\infty)}\bigl(1-e^{-st}\bigr)\,\Lambda(ds)\\
&=\int_{(0,\infty)}\Bigl[\bigl(1-e^{-\gamma^2 st}\bigr)-\gamma^2\bigl(1-e^{-st}\bigr)\Bigr]\,\Lambda(ds).
\end{align*}
The bracket has the form \(g(\gamma^2 u)-\gamma^2 g(u)\) with \(g(u):=1-e^{-u}\) and \(u=st\), and we show that it is positive for every \(u>0\). First, \(g\) is strictly concave on \([0,\infty)\), since \(g''(u)=-e^{-u}<0\), and \(g(0)=0\). Second, since \(\gamma^2\in(0,1)\), the point \(\gamma^2 u\) lies between \(0\) and \(u\), as the convex combination \(\gamma^2 u=\gamma^2\cdot u+(1-\gamma^2)\cdot 0\). Strict concavity at this point gives
\[
g(\gamma^2 u)=g\bigl(\gamma^2 u+(1-\gamma^2)\cdot 0\bigr)>\gamma^2 g(u)+(1-\gamma^2)\,g(0)=\gamma^2 g(u).
\]
So the bracket is positive for every \(s>0\). When \(\Lambda\not\equiv 0\), the integral of this positive function against \(\Lambda\) is positive, and \(\phi(t)\ge\int_{(0,\infty)}g(st)\,\Lambda(ds)>0\). Hence
\[
\tau_k^\star(\mathcal T^\pi)\ge\frac{\phi(\gamma^2 t)}{\phi(t)}>\gamma^2.
\]
Since \(\tau^\star(\gamma)\ge\tau_k^\star(\mathcal T^\pi)\), as noted after Theorem~\ref{thm:contraction-drift-density-atoms}, the smallest contraction factor across dimensions also exceeds \(\gamma^2\).

\textbf{When \(\Lambda\equiv 0\).}
Then \(\phi(t)=bt\), and \(k\) is the kernel
\[
k(x,y)=C-b\,\|x-y\|^2.
\]
For this kernel, \(\MMD_k\) only compares means (Proposition~\ref{prop:pure_drift_identities}).

With \(\Lambda\equiv 0\), the criterion of Theorem~\ref{thm:contraction-drift-density-atoms} reduces to \(b(\tau-\gamma^2)\ge 0\), which holds for every \(\tau\in[\gamma^2,1)\). Contraction then holds across dimensions, so
\[
\tau_k^\star(\mathcal T^\pi)\le\tau^\star(\gamma)\le\gamma^2.
\]
Conversely, when \(b>0\), the lower bound of the first part applies with \(\phi(t)=bt>0\), and its ratio is exactly
\[
\frac{\phi(\gamma^2 t)}{\phi(t)}=\frac{b\gamma^2 t}{bt}=\gamma^2.
\]
Hence \(\tau_k^\star(\mathcal T^\pi)=\gamma^2\), and this kernel reaches the value \(\gamma^2\). When \(b=0\) as well, the kernel is constant and \(\MMD_k\) is identically zero.
\end{proof}

\subsection{Permutation test and power estimation}\label{app:permutation_test_power}
\(\MMD^2\) values are not comparable across kernels. For example, if \(k'=c\,k\) with \(c>0\), then \(\MMD_{k'}^2=c\,\MMD_k^2\), so a larger value can reflect only a change of units. Following \citet{borgwardt2006integrating,gretton2006kernel,JMLR:v13:gretton12a}, we therefore compare kernels through the power of a level-\(\alpha\) permutation test built from the biased empirical statistic.

\paragraph{Level and power of a test.} A hypothesis test of \(H_0\colon P=Q\) against 
\(H_1\colon P\neq Q\) is a binary rule \(\Delta_k\in\{0,1\}\), with \(\Delta_k=1\) 
meaning rejection. It has level \(\alpha\) if \(\sup_{P}\,\mathbb{P}_{P\times P}
(\Delta_k=1)\le\alpha\), controlling the false-positive rate. Its power against 
\((P,Q)\) with \(P\neq Q\) is \(\mathbb{P}_{P\times Q}(\Delta_k=1)\), 
the probability of correctly detecting that \(P\neq Q\).

\paragraph{Test statistic.} Given samples \(X_{1:n}\sim P^n\) and \(Y_{1:m}\sim Q^m\), let \(N:=n+m\), form the aggregate sample \(Z_{1:N}=(X_{1:n},Y_{1:m})\), and let \(K\in\R^{N\times N}\) be its Gram matrix, \(K_{ij}:=k(Z_i,Z_j)\). Define the two-sample weight vector \(\ell\in\R^N\) as in Section~\ref{app:mmd_gram_matrices}, \[ \ell_i := \begin{cases} \tfrac{1}{n}, & 1\le i\le n,\\[3pt] -\tfrac{1}{m}, & n+1\le i\le N, \end{cases} \] so that the test statistic is \[ T_k := \ell^\top K\ell = \widehat{\MMD}_{k,\mathrm{b}}^2(X_{1:n},Y_{1:m}). \]

\paragraph{Null hypothesis and permutation test.} Under \(H_0\), the labels are exchangeable, so permuting \(\{1,\dots,N\}\) leaves the joint distribution of the aggregate sample unchanged. Given \(B_{\mathrm{perm}}\) uniformly drawn permutations \(\pi_1,\dots,\pi_{B_{\mathrm{perm}}}\) of \(\{1,\dots,N\}\), each induces a permuted statistic \(T_{k,\pi_b}:=\ell^\top K_{\pi_b}\ell\), where \((K_{\pi_b})_{ij}:=k(Z_{\pi_b(i)},Z_{\pi_b(j)})\). The permutation \(p\)-value is \[ \hat p := \frac{1 + \#\bigl\{b\in\{1,\dots,B_{\mathrm{perm}}\} : T_{k,\pi_b}\ge T_k\bigr\}}{B_{\mathrm{perm}}+1}, \] and \(H_0\) is rejected if and only if \(\hat p\le\alpha\).

\begin{figure}[t]
    \centering
    \includegraphics[width=\linewidth]{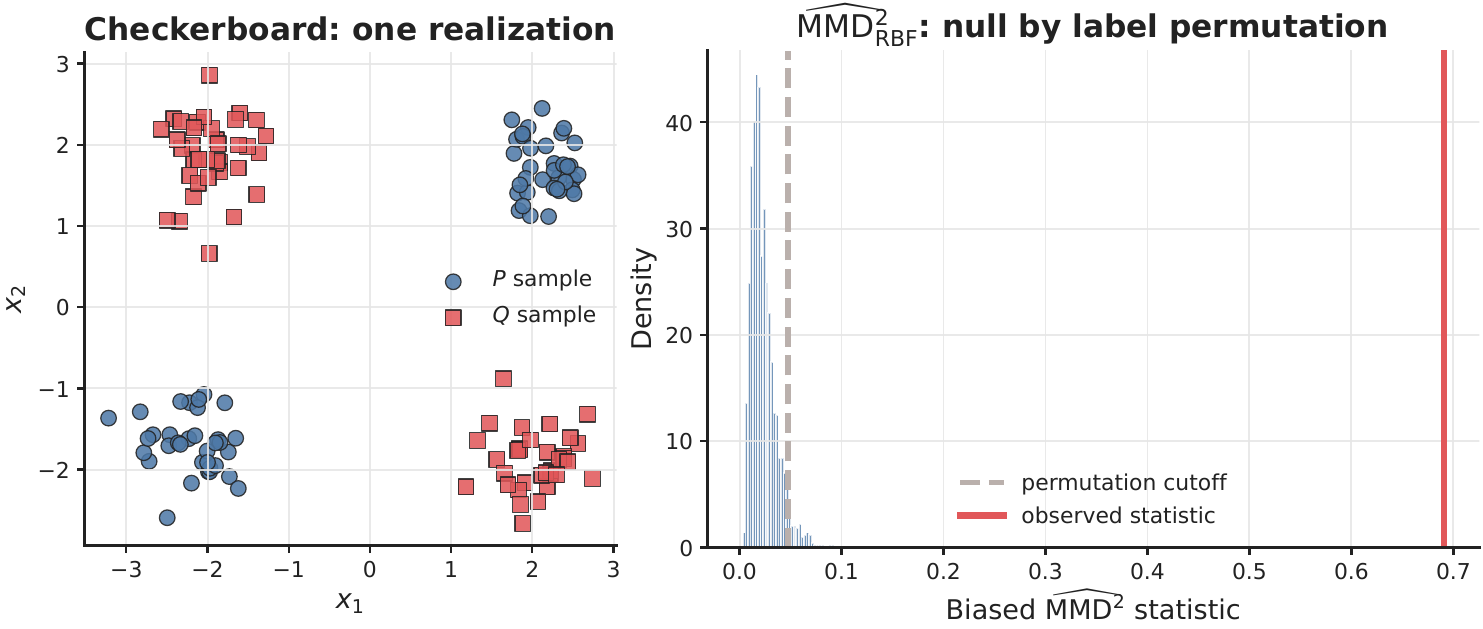}
    \caption{
    One example of the permutation test used for power estimation, using the RBF kernel
    with lengthscale \(\ell=1.05\) at level \(\alpha=0.05\).
    Left, one realization of the checkerboard two-sample problem (\(n=m=70\)).
    Right, the permutation null for the biased \(\widehat{\MMD}^2_{\mathrm{RBF}}\)
    statistic, obtained from \(2{,}000\) label permutations on the aggregate sample.
    The dashed line is the level-\(\alpha\) cutoff and the red line is the observed statistic.
    }
    \label{fig:permutation_null_example}
\end{figure}

Figure~\ref{fig:permutation_null_example} illustrates one instance of this test. The aggregate sample is fixed, the labels are permuted to form the null distribution, and the observed statistic is compared with the resulting level-\(\alpha\) cutoff.

\paragraph{Power estimation.} The power of the level-\(\alpha\) test for a given kernel \(k\) is \(\mathbb{P}_{H_1}(\hat{p} \le \alpha)\). It is estimated by repeating the following procedure \(B\) times independently. Each run draws a fresh dataset pair \((X_{1:n}^{(b)}, Y_{1:m}^{(b)})\), computes the observed statistic \(T_k^{(b)}\), runs \(B_{\mathrm{perm}}\) permutations to obtain the p-value \(\hat{p}^{(b)}\), and records whether \(\hat{p}^{(b)} \le \alpha\). The empirical power \(\widehat{D}_{k,n,m,\alpha}(P,Q)\) is the fraction of rejections across the \(B\) repetitions. Algorithm~\ref{alg:perm_detectability_protocol} summarizes this procedure.

\begin{algorithm}[t]
\caption{Estimating the power of the permutation test for one kernel configuration}
\label{alg:perm_detectability_protocol}
\begin{algorithmic}[1]
\STATE Fix \((P,Q)\), sample sizes \((n,m)\), a level \(\alpha\), an outer repetition count \(B\), and a permutation budget \(B_{\mathrm{perm}}\).
\FOR{\(b=1,\dots,B\)}
\STATE Draw \(X_{1:n}^{(b)}\sim P^n\) and \(Y_{1:m}^{(b)}\sim Q^m\).
\STATE Build the Gram matrix \(K^{(b)}\) of the aggregate sample and compute \(T_k^{(b)}=\widehat{\MMD}_{k,\mathrm{b}}^2\).
\STATE Permute the labels \(B_{\mathrm{perm}}\) times on the aggregate sample and recompute the corresponding statistics \(T_{k,\pi}^{(b)}\).
\STATE Compute the permutation $p$-value \[ \hat{p}^{(b)} \;=\; \frac{1+\#\bigl\{T_{k,\pi}^{(b)}\ge T_k^{(b)}\bigr\}}{B_{\mathrm{perm}}+1}. \]
\ENDFOR
\STATE Return \(\widehat D_{k,n,m,\alpha}(P,Q) := \frac{1}{B}\sum_{b=1}^B \mathbf{1}\bigl[\hat{p}^{(b)}\le\alpha\bigr]\).
\end{algorithmic}
\end{algorithm}

\subsection{Estimating the tradeoff between power and contraction}\label{app:tradeoff_sweep}
For a given distribution pair $(P,Q)$ and kernel family $\{k_\theta\}$, lowering the smallest contraction factor on a bounded set and maximizing discriminative power pull in opposite directions. To make this tradeoff visible, we fix a contraction budget $\bar\tau$ and report, over all $\theta$ with $\widehat\tau^\star_{k_\theta,E}(\gamma) \leq \bar\tau$, the best achievable power \[ \widehat{D}^{\,\star}(\gamma, \bar\tau) \;:=\; \max\bigl\{\,\widehat{D}_{k_\theta,n,m,\alpha}(P,Q) : \widehat{\tau}^\star_{k_\theta,E}(\gamma) \leq \bar\tau\,\bigr\}, \] leaving cells with no feasible $\theta$ undefined. Sweeping over $(\gamma, \bar\tau)$ produces the tradeoff heatmap of Algorithm~\ref{alg:tradeoff_sweep}.

\paragraph{Estimation procedure.}
For each kernel family and parameter grid in Table~\ref{tab:tradeoff_kernel_grids}, and for each $\gamma$, the smallest contraction factor $\tau^\star_{k_\theta,E}(\gamma)$ on the ball $E = B(0, R)$ is estimated by Algorithm~\ref{alg:tau-star-mc}. For each dataset in Table~\ref{tab:tradeoff_datasets}, \(R\) is chosen to contain most of the mass and is used to clip the sampled data. The numerical settings are given in Table~\ref{tab:tradeoff_hparams}. In particular, we use support size $N$ and $J$ independent random supports, retaining the maximum over the $J$ estimates. For each $\theta$, the power $\widehat{D}_{k_\theta, n, m, \alpha}(P,Q)$ is estimated by Algorithm~\ref{alg:perm_detectability_protocol}. The full procedure is summarized in Algorithm~\ref{alg:tradeoff_sweep}.

\begin{algorithm}[t]
\caption{Measuring the tradeoff between power and contraction}
\label{alg:tradeoff_sweep}
\begin{algorithmic}[1]
\STATE \textbf{Require:} Distribution pair $(P,Q)$, sample sizes $(n,m)$, level $\alpha$, discount grid $\{\gamma_l\}$, grid of contraction budgets $\bar\tau$, parameter grid $\{\theta_i\}$, ball radius $R$, support size $N$, trial count $J$, outer repetition count $B$, permutation budget $B_{\mathrm{perm}}$.
\medskip
\STATE \textcolor{green!50!black}{\textit{Step 1: estimate power for each kernel parameter.}}
\FOR{each $\theta_i$}
  \STATE Run Algorithm~\ref{alg:perm_detectability_protocol} with kernel $k_{\theta_i}$. Set $\widehat{D}_i := \widehat{D}_{k_{\theta_i},n,m,\alpha}(P,Q)$.
\ENDFOR
\medskip
\STATE \textcolor{green!50!black}{\textit{Step 2: estimate the smallest contraction factor on \(E\) for each kernel parameter and discount.}}
\FOR{each $\theta_i$ and each $\gamma_l$}
  \STATE Run Algorithm~\ref{alg:tau-star-mc} with kernel $k_{\theta_i}$, scaling parameter $\lambda = \gamma_l$, ball $B(0,R)$, support size $N$, and $J$ trials. Set $\widehat{\tau}_{i,l} := \max_{1 \leq j \leq J} \widehat{\tau}^\star_{S_j}(\gamma_l)$.
\ENDFOR
\medskip
\STATE \textcolor{green!50!black}{\textit{Step 3: build the tradeoff heatmap.}}
\FOR{each $\gamma_l$ and each budget $\bar\tau$}
  \STATE Record $\max_{i\,:\,\widehat{\tau}_{i,l} \leq \bar\tau} \widehat{D}_i$, or mark the cell as empty if no $\theta_i$ satisfies the constraint.
\ENDFOR
\end{algorithmic}
\end{algorithm}

\begin{table}[t]
\centering
\small
\begin{tabular}{@{}lp{1.0cm}p{7.5cm}@{}}
\toprule
\textbf{Setting} & \textbf{Value} & \textbf{Description} \\
\midrule
Support size \(N\) & \(512\) & Number of points drawn uniformly in \(B(0,R)\) for each trial support. \\
\midrule
Trial count \(J\) & \(1\) & Number of independent trial supports used to estimate \(\tau^\star_{k_\theta,E}(\gamma)\). \\
\midrule
Eigenvalue tolerance \(\varepsilon_{\mathrm{eig}}\) & \(10^{-9}\) & Relative tolerance below which eigenvalues of \(A\) are discarded before forming the reduced problem. \\
\midrule
Outer repetitions \(B\) & \(1000\) & Number of independent dataset draws used to estimate power. \\
\midrule
Permutation budget \(B_{\mathrm{perm}}\) & \(1000\) & Number of label permutations used to compute the $p$-value on each dataset draw. \\
\midrule
Level \(\alpha\) & \(0.05\) & Significance level of the permutation test. \\
\bottomrule
\end{tabular}
\caption{Numerical settings used in Algorithm~\ref{alg:tradeoff_sweep}.}
\label{tab:tradeoff_hparams}
\end{table}

\begin{table}[H]
\centering
\small
\begin{tabular}{@{}p{6.4cm}p{2.6cm}p{3.4cm}@{}}
\toprule
\textbf{Kernel family and formula} & \textbf{Parameter \(\theta\)} & \textbf{Grid over \(\Theta\)} \\
\midrule
\begin{tabular}[t]{@{}l@{}}
RBF \\
\(k_h(x,y)=\exp\!\left(-h^2\|x-y\|^2\right)\)
\end{tabular}
&
Scale \(h\)
&
\(\operatorname{geom}(10^{-3},10^3,100)\)
\\
\midrule
\begin{tabular}[t]{@{}l@{}}
Generalized multiquadric \\
\(k_{\beta,h}(x,y)=-\bigl(1+h^2\|x-y\|^2\bigr)^\beta\)
\end{tabular}
&
Power \(\beta\)
&
\(\operatorname{lin}(0.05,1.00,50)\)
\\
\midrule
\begin{tabular}[t]{@{}l@{}}
Power distance \\
\(k_\alpha(x,y)=-\|x-y\|^\alpha\)
\end{tabular}
&
Power \(\alpha\)
&
\(\operatorname{lin}(0.20,2.00,50)\)
\\
\midrule
\begin{tabular}[t]{@{}l@{}}
Logarithmic kernel \\
\(k_a(x,y)=-\log\!\left(1+\|x-y\|^2/a\right)\)
\end{tabular}
&
Scale \(a\)
&
\(\operatorname{geom}(10^{-3},10^3,100)\)
\\
\bottomrule
\end{tabular}
\caption{Kernel families and the values of their parameter used to study the tradeoff between power and contraction. Here \(\operatorname{geom}(a,b,n)\) denotes \(n\) geometrically spaced values from \(a\) to \(b\), and \(\operatorname{lin}(a,b,n)\) denotes \(n\) linearly spaced values from \(a\) to \(b\).}
\label{tab:tradeoff_kernel_grids}
\end{table}

\begin{table}[t]
\centering
\small
\setlength{\tabcolsep}{4pt}
\begin{tabular}{@{}p{3.2cm}p{8.2cm}p{1.2cm}@{}}
\toprule
\textbf{Dataset} & \textbf{Distribution} & \textbf{Radius \(R\)} \\
\midrule
1D Mean Shift
&
\(\begin{aligned}
P &= \mathcal{N}(-0.40,0.18^2),\\
Q &= \mathcal{N}(0.40,0.18^2)
\end{aligned}\)
&
\(1.40\)
\\
\midrule
1D Same Mean, Different Std
&
\(\begin{aligned}
P &= \mathcal{N}(0,0.25^2),\\
Q &= \mathcal{N}(0,0.55^2)
\end{aligned}\)
&
\(2.00\)
\\
\midrule
1D Bimodal vs Unimodal
&
\(\begin{aligned}
P &= \tfrac12\mathcal{N}(-0.65,0.15^2)
   + \tfrac12\mathcal{N}(0.65,0.15^2),\\
Q &= \mathcal{N}(0,0.65^2+0.15^2)
\end{aligned}\)
&
\(2.00\)
\\
\midrule
Two-Gaussian Location
&
\(\begin{aligned}
P &= \mathcal{N}\!\left((0.55,0),0.18^2 I_2\right),\\
Q &= \mathcal{N}\!\left((-0.55,0),0.18^2 I_2\right)
\end{aligned}\)
&
\(1.35\)
\\
\midrule
Covariance Swap
&
\(\begin{aligned}
P &= \mathcal{N}\!\left(0,
\begin{bmatrix}
0.28 & 0.18\\
0.18 & 0.28
\end{bmatrix}\right),\\
Q &= \mathcal{N}\!\left(0,
\begin{bmatrix}
0.28 & -0.18\\
-0.18 & 0.28
\end{bmatrix}\right)
\end{aligned}\)
&
\(2.40\)
\\
\midrule
Bimodal vs Unimodal
&
\(\begin{aligned}
P &= \tfrac12\mathcal{N}\!\left((-0.75,0),0.18^2 I_2\right)
   + \tfrac12\mathcal{N}\!\left((0.75,0),0.18^2 I_2\right),\\
Q &= \mathcal{N}\!\left(0,
\begin{bmatrix}
0.75^2+0.18^2 & 0\\
0 & 0.18^2
\end{bmatrix}\right)
\end{aligned}\)
&
\(2.40\)
\\
\midrule
Checkerboard
&
\(\begin{aligned}
P &= \tfrac12\mathcal{N}\!\left((-2.2,-1.7),0.35^2 I_2\right)
   + \tfrac12\mathcal{N}\!\left((2.2,1.7),0.35^2 I_2\right),\\
Q &= \tfrac12\mathcal{N}\!\left((-2.0,1.9),0.35^2 I_2\right)
   + \tfrac12\mathcal{N}\!\left((2.0,-1.9),0.35^2 I_2\right)
\end{aligned}\)
&
\(4.40\)
\\
\bottomrule
\end{tabular}
\caption{Datasets used to study the tradeoff between power and contraction. The displayed Gaussian laws are sampled and then clipped to the ball \(B(0,R)\).}
\label{tab:tradeoff_datasets}
\end{table}

\clearpage
\newpage
\begin{figure}[H]
    \centering
    \includegraphics[width=\linewidth]{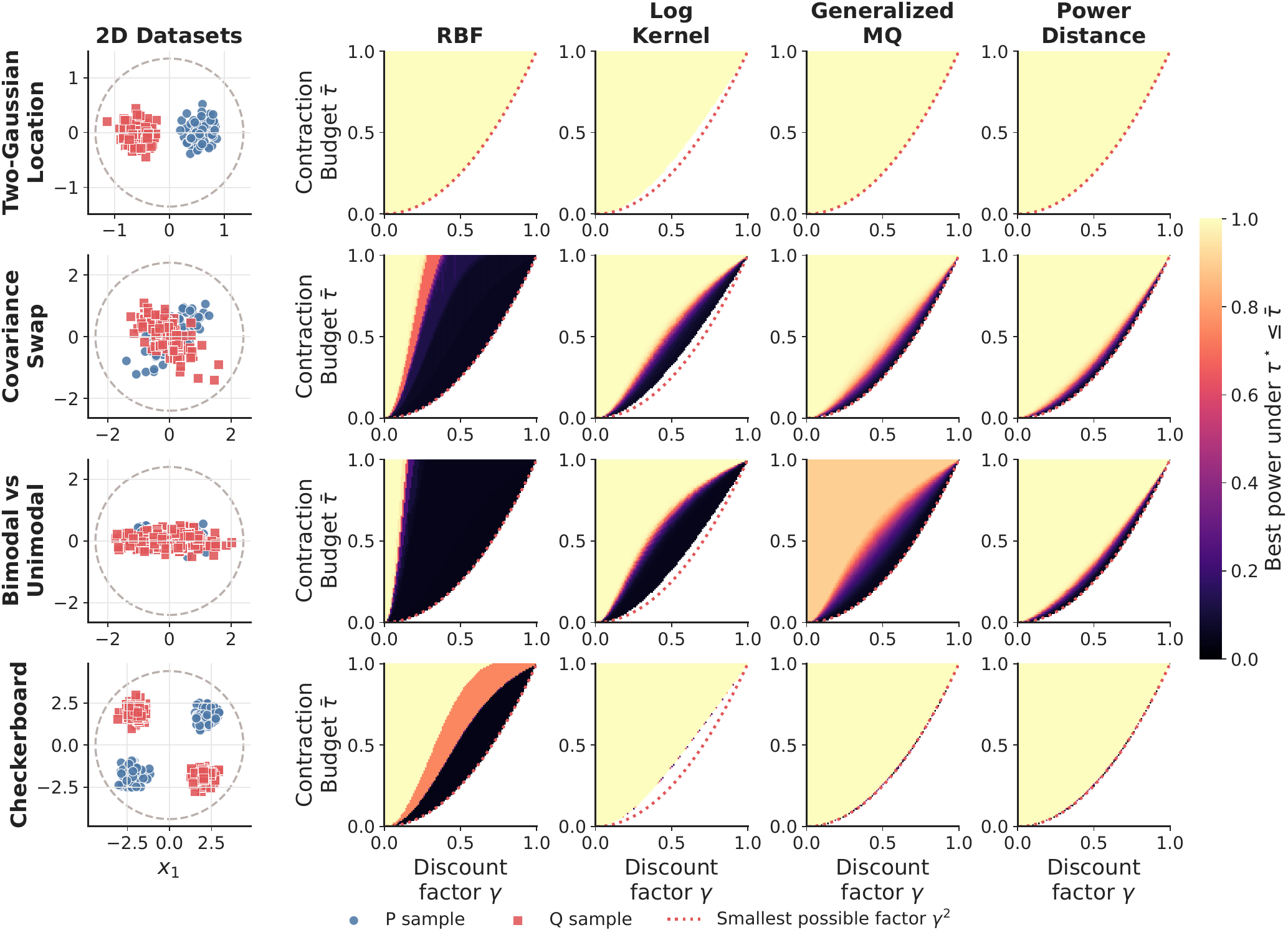}
    \caption{Two-dimensional tradeoff experiments. The left column shows one sampled dataset for each task, with dashed circles marking the clipping radius \(R\). The remaining columns show, for each kernel family, the best empirical test power achievable under the constraint \(\tau^\star_{k_\theta,E}(\gamma)\leq \bar{\tau}\). The dotted curve is the smallest possible contraction factor \(\gamma^2\).}
    \label{fig:empirical_envelope_2d_kernel_comparison}
\end{figure}

\newpage

\subsection{Higher dimensions}\label{app:tradeoff_dimension}
Return distributions are multivariate when rewards are vector valued \citep{zhang2021distributional,wiltzer2024foundations,debes2026multivariate}, so the two samples compared by the test can live in higher dimensions. We therefore check whether the tradeoff persists as the dimension grows, on two pairs of distributions defined in any dimension, a mean shift and a checkerboard on the corners of a cube (Table~\ref{tab:tradeoff_dimension}). We use the power-distance kernels \(k(x,y)=-\|x-y\|^{\alpha}\), whose smallest contraction factor across dimensions is \(\gamma^{\alpha}\) (Proposition~\ref{prop:main_known_positive_kernels}). The constraint of Appendix~\ref{app:tradeoff_sweep} is then \(\gamma^{\alpha}\le\bar\tau\), with no estimation on a ball.

\begin{figure}[H]
    \centering
    \includegraphics[width=\linewidth]{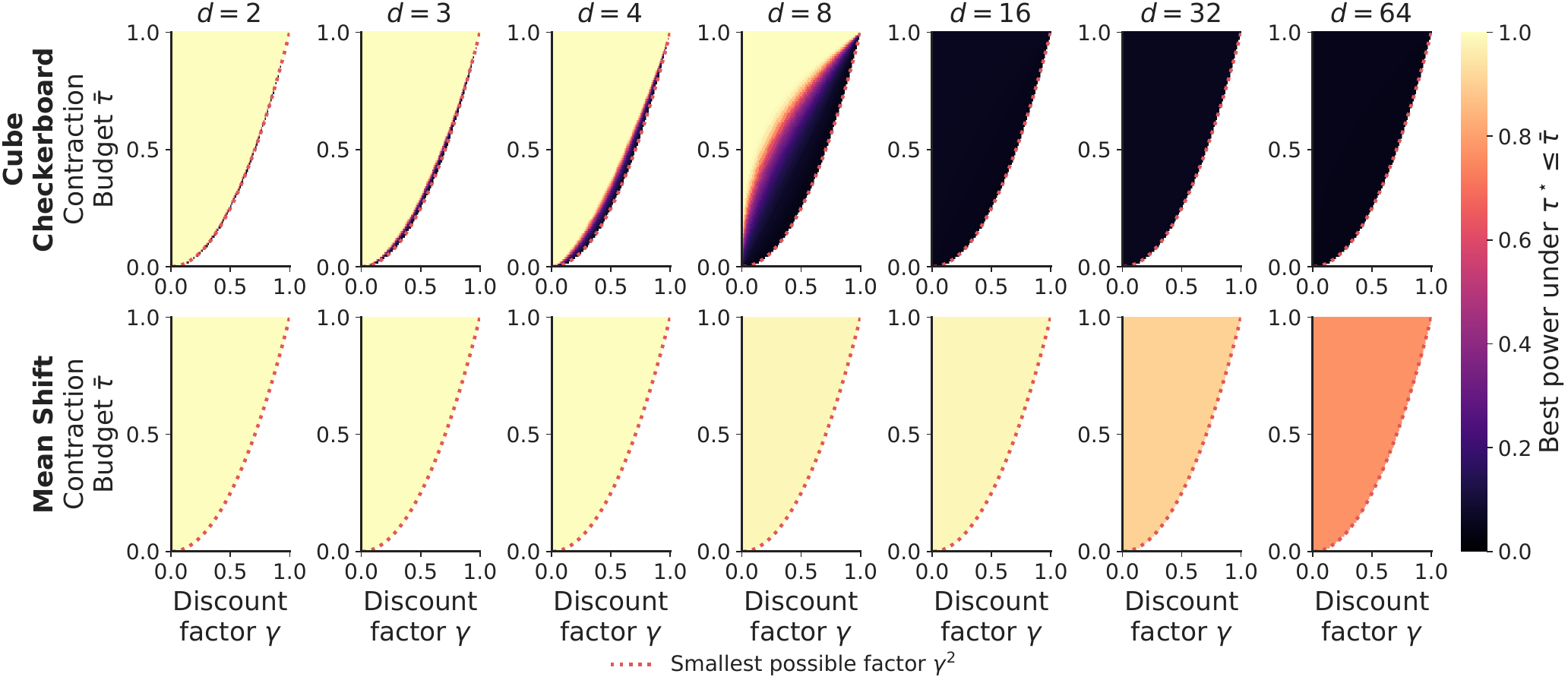}
    \caption{Tradeoff experiments in higher dimensions for the power-distance kernels \(k(x,y)=-\|x-y\|^{\alpha}\), with the settings of Table~\ref{tab:tradeoff_dimension}. Each column is a dimension \(d\). Each panel shows the best empirical test power achievable under the constraint \(\gamma^{\alpha}\le\bar\tau\). The dotted curve is the smallest possible contraction factor \(\gamma^2\).}
    \label{fig:empirical_envelope_dimension}
\end{figure}

Figure~\ref{fig:empirical_envelope_dimension} shows the results. For the mean shift, the power does not depend on \(\alpha\), so faster contraction costs no power in any dimension. For the checkerboard, the two distributions have the same mean. The kernels with larger \(\alpha\) contract faster and are closer to comparing means, and they lose their power first. As the dimension grows, they lose it at smaller \(\alpha\), so the tradeoff becomes stronger. As the dimension grows, each distribution also spreads over \(2^{d-1}\) corners, many more than the \(100\) points of a sample. By dimension \(16\), no kernel detects the difference, so the tradeoff can no longer be observed.

\begin{table}[H]
\centering
\small
\setlength{\tabcolsep}{4pt}
\begin{tabular}{@{}p{3.2cm}p{9.4cm}@{}}
\toprule
\textbf{Dataset} & \textbf{Distribution} \\
\midrule
Cube Checkerboard & \(P\) and \(Q\) uniform on the corners \(s/\sqrt{d}\), \(s\in\{-1,1\}^d\), of a cube, colored like a checkerboard, with the corners having an even number of \(-1\) for \(P\) and those having an odd number for \(Q\) \\
\midrule
Mean Shift & \(P=\mathcal{N}(-0.4\,u,I_d)\) and \(Q=\mathcal{N}(0.4\,u,I_d)\), with \(u=(1,\dots,1)/\sqrt{d}\) \\
\bottomrule
\end{tabular}

\vspace{8pt}
\begin{tabular}{@{}lp{3.6cm}p{6.0cm}@{}}
\toprule
\textbf{Setting} & \textbf{Value} & \textbf{Description} \\
\midrule
Dimension \(d\) & \(2,3,4,8,16,32,64\) & Dimension of the samples. \\
\midrule
Power \(\alpha\) & \(\operatorname{lin}(0.20,2.00,50)\) & Parameter of \(k_\alpha(x,y)=-\|x-y\|^\alpha\). \\
\midrule
Sample sizes \((n,m)\) & \((100,100)\) & Number of points in each sample. \\
\bottomrule
\end{tabular}
\caption{Distributions and settings of the experiments in higher dimensions. The outer repetitions, the permutation budget and the level are those of Table~\ref{tab:tradeoff_hparams}.}
\label{tab:tradeoff_dimension}
\end{table}

\clearpage

\section{Properties of radial kernels built from Bernstein functions}\label{app:properties_bernstein_based_kernels}
\paragraph{Preamble.}
This section studies when the radial kernels built from Bernstein functions induce a metric through \(\MMD\), rather than only a pseudometric. For kernels of the form
\[
k(x,y)=C-\phi(\|x-y\|^2),
\]
the key issue is whether the Bernstein part \(\phi(\|x-y\|^2)\) is \emph{strictly} conditionally negative definite. Lemma~\ref{lemma:Bernstein_CSND} shows that this happens exactly when the L\'evy measure \(\Lambda\) is nonzero. Theorem~\ref{thm:bernstein_mmd_metric} then combines this with the general criterion for MMD to be a metric and shows that \(\MMD_k\) is a metric if and only if \(\Lambda\not\equiv0\).

\begin{lemma}[Quadratic forms of squared Euclidean distances]\label{lem:euclidean_squared_distance_identity}
Let \(x_1,\dots,x_n\in\R^d\) and \(w_1,\dots,w_n\in\R\) satisfy \(\sum_{i=1}^n w_i=0\).
Then
\[
\sum_{i,j=1}^n w_i w_j\,\|x_i-x_j\|^2
=
-2\Bigl\|\sum_{i=1}^n w_i x_i\Bigr\|^2
\ \le\ 0.
\]
Moreover, the left-hand side equals \(0\) if and only if \(\sum_{i=1}^n w_i x_i=0\).
\end{lemma}
\begin{proof}
Expand
\[
\|x_i-x_j\|^2=\|x_i\|^2+\|x_j\|^2-2\langle x_i,x_j\rangle.
\]
Hence
\begin{align*}
\sum_{i,j=1}^n w_i w_j\,\|x_i-x_j\|^2
&=
\sum_{i,j=1}^n w_i w_j\,\|x_i\|^2
+
\sum_{i,j=1}^n w_i w_j\,\|x_j\|^2
-2\sum_{i,j=1}^n w_i w_j\,\langle x_i,x_j\rangle.
\end{align*}
Using \(\sum_{j=1}^n w_j=0\),
\[
\sum_{i,j=1}^n w_i w_j\,\|x_i\|^2
=
\sum_{i=1}^n \Bigl(w_i\|x_i\|^2\sum_{j=1}^n w_j\Bigr)=0,
\quad
\sum_{i,j=1}^n w_i w_j\,\|x_j\|^2
=
\sum_{j=1}^n \Bigl(w_j\|x_j\|^2\sum_{i=1}^n w_i\Bigr)=0.
\]
Therefore
\[
\sum_{i,j=1}^n w_i w_j\,\|x_i-x_j\|^2
=
-2\sum_{i,j=1}^n w_i w_j\,\langle x_i,x_j\rangle.
\]
Finally,
\[
\sum_{i,j=1}^n w_i w_j\,\langle x_i,x_j\rangle
=
\Bigl\langle \sum_{i=1}^n w_i x_i,\ \sum_{j=1}^n w_j x_j\Bigr\rangle
=
\Bigl\|\sum_{i=1}^n w_i x_i\Bigr\|^2,
\]
which proves the identity and the inequality. The last claim follows since
\(\bigl\|\sum_{i=1}^n w_ix_i\bigr\|^2=0\) only when \(\sum_{i=1}^n w_ix_i=0\).
\end{proof}

\begin{lemma}[When a radial kernel built from a Bernstein function is CSND]\label{lemma:Bernstein_CSND}
Let \(\phi\colon[0,\infty)\to[0,\infty)\) be a Bernstein function with \(\phi(0)=0\), admitting the L\'evy--Khintchine representation
\[
\phi(t)\;=\;b\,t\;+\;\int_{(0,\infty)}\bigl(1-e^{-t\,s}\bigr)\,\Lambda(d s),
\]
where \(b\ge0\) and \(\Lambda\) is a nonnegative measure satisfying \(\displaystyle\int_{(0,\infty)}\min\{1,s\}\,\Lambda(d s)<\infty\).  Define the radial kernel
\[
k(x,y)\;=\;\phi\bigl(\|x-y\|^2\bigr)
\]
on \(\R^d\).  Then \(k\) is conditionally \emph{strictly} negative definite (CSND) on \(\R^d\) if and only if \(\Lambda\) is not identically zero (equivalently, \(\phi\) is not of the form \(\phi(t)=b\,t\)).  Moreover, if \(\Lambda\equiv0\), then \(k\) is CND but not CSND.
\end{lemma}
\begin{proof}
We split into two cases.

\noindent\textbf{Case 1: Zero L\'evy measure, \(\Lambda\equiv0\).}
Then \(\phi(t)=b\,t\) and
\[
k(x,y)=b\,\|x-y\|^2.
\]
For any finite choice \(x_1,\dots,x_n\in\R^d\) and weights \(w_1,\dots,w_n\) with \(\sum_i w_i=0\),
\[
\sum_{i,j=1}^n w_i w_j\,k(x_i,x_j)
=
b\sum_{i,j=1}^n w_i w_j\,\|x_i-x_j\|^2
=
-2b\,\Bigl\|\sum_{i=1}^n w_i x_i\Bigr\|^2
\le 0
\]
by Lemma~\ref{lem:euclidean_squared_distance_identity}. Hence \(k\) is CND.
Moreover, if \(\sum_i w_i x_i=0\) with \(\sum_i w_i=0\) and \(w\) nonzero, then the quadratic form vanishes, so \(k\) is not CSND.

\noindent\textbf{Case 2: Nonzero L\'evy measure, $\Lambda\not\equiv0$.}
From the L\'evy--Khintchine form,
\[
k(x,y)= b\,\|x-y\|^2
  + \int_{(0,\infty)}\bigl(1-e^{-s\|x-y\|^2}\bigr)\,\Lambda(ds).
\]
Fix pairwise distinct $x_1,\dots,x_n\in\R^d$ and weights $w_1,\dots,w_n$ with $\sum_i w_i=0$, not all zero, and set
\[
S := \sum_{i,j=1}^n w_i w_j\,k(x_i,x_j).
\]

\emph{Drift part.}
By Lemma~\ref{lem:euclidean_squared_distance_identity},
\[
b\sum_{i,j} w_i w_j \|x_i-x_j\|^2
\;=\;
-2b\Bigl\|\sum_{i=1}^n w_i x_i\Bigr\|^2
\;\le\;0.
\]

\emph{Interchanging the finite sum and the integral.}
For each $(i,j)$ and all $s>0$, set $u:=s\|x_i-x_j\|^2\ge 0$. Since
\[
1-e^{-u}\le 1
\qquad\text{and}\qquad
1-e^{-u}\le u,
\]
we have
\[
0 \le 1-e^{-s\|x_i-x_j\|^2} \le \min\{1,\; s\|x_i-x_j\|^2\}.
\]
Moreover,
\[
\min\{1,\,s\|x_i-x_j\|^2\}\le \max\{1,\|x_i-x_j\|^2\}\,\min\{1,s\},
\]
so
\[
\int_{(0,\infty)} \!\bigl|1-e^{-s\|x_i-x_j\|^2}\bigr|\,\Lambda(ds)
\;\le\;
\max\{1,\|x_i-x_j\|^2\} \int_{(0,\infty)} \min\{1,s\}\,\Lambda(ds)
\;<\;\infty.
\]
Hence each integral is finite, and since there are only finitely many pairs $(i,j)$, we may exchange the finite sum and the integral,
\[
\sum_{i,j} w_i w_j \!\int_{(0,\infty)} \!\bigl(1-e^{-s\|x_i-x_j\|^2}\bigr)\,\Lambda(ds)
= \int_{(0,\infty)} \Bigl[\sum_{i,j} w_i w_j \bigl(1-e^{-s\|x_i-x_j\|^2}\bigr)\Bigr]\Lambda(ds).
\]

\emph{Inner sum decomposition.}
For any fixed $s>0$,
\[
\sum_{i,j} w_i w_j \bigl(1-e^{-s\|x_i-x_j\|^2}\bigr)
= \underbrace{\sum_{i,j} w_i w_j}_{=0}
  - \sum_{i,j} w_i w_j\,e^{-s\|x_i-x_j\|^2}
= -\,Q(s),
\]
where
\[
Q(s) := \sum_{i,j=1}^n w_i w_j\,e^{-s\|x_i-x_j\|^2}.
\]

\emph{Strict positivity of \(Q(s)\).}
Fix \(s>0\). By \cite[Cor.~2.1(c)]{micchelli1986interpolation}, for pairwise distinct
\(x_1,\dots,x_n\in\R^d\),
\[
\sum_{i,j=1}^n w_i w_j\,e^{-s\|x_i-x_j\|^2} \;>\; 0
\qquad\text{for every nonzero }(w_1,\dots,w_n)\in\R^n.
\]
Therefore \(Q(s)>0\) for all \(s>0\) whenever the weights are not all zero, which is the case for the weights fixed above.

\emph{Conclusion.}
Combining the pieces,
\[
S \;=\; -2b\Bigl\|\sum_i w_i x_i\Bigr\|^2 \;-\; \int_{(0,\infty)} Q(s)\,\Lambda(ds).
\]
The first term is $\le 0$. Moreover, since $\Lambda\not\equiv 0$ and $Q(s)>0$ for all $s>0$, we have
\[
\int_{(0,\infty)} Q(s)\,\Lambda(ds) > 0.
\]
Hence $S<0$, and therefore $k$ is conditionally strictly negative definite.

\end{proof}

\begin{theorem}[When radial kernels built from Bernstein functions make MMD a metric]\label{thm:bernstein_mmd_metric}
Let $d\ge 1$ and let $k:\R^d\times\R^d\to\R$ be symmetric.
Assume that there exist a constant $C\in\R$ and a Bernstein function
$\phi:[0,\infty)\to[0,\infty)$ with $\phi(0)=0$ such that
\[
k(x,y)\;=\;C-\phi\bigl(\|x-y\|^2\bigr).
\]
By Theorem~\ref{thm:schoenberg-bernstein}, $k$ is CPD, so $\MMD_k$ is defined as a quadratic form, as in Definition~\ref{def:mmd}.
Write the L\'evy--Khintchine representation of $\phi$ as
\[
\phi(t)\;=\;b\,t\;+\;\int_{(0,\infty)}\bigl(1-e^{-t\,s}\bigr)\,\Lambda(ds),
\qquad b\ge0,
\]
where $\Lambda$ is a nonnegative measure with $\int_{(0,\infty)}\min\{1,s\}\,\Lambda(ds)<\infty$.

Then $\MMD_k$ satisfies the metric axioms except possibly identity of indiscernibles
(as in Proposition~\ref{prop:mmd-metric}). Moreover,
\[
\boxed{
\MMD_k \text{ is a metric if and only if } \Lambda\not\equiv 0.}
\]
\end{theorem}

\begin{proof}
\noindent\paragraph{Step 1: Reducing identity of indiscernibles to strictness.}
Since $k$ is CPD, Proposition~\ref{prop:mmd-metric} implies that $\MMD_k$ satisfies nonnegativity, symmetry,
and the triangle inequality. It is a metric if and only if identity of indiscernibles holds, which by the same proposition
is equivalent to $k$ being CSPD.

\noindent\paragraph{Step 2: From CSPD of \(k\) to CSND of the Bernstein part.}
Adding a constant does not change whether a kernel is CPD or CSPD, hence
\[
k \text{ is CSPD}
\quad\Longleftrightarrow\quad
-\phi(\|x-y\|^2)\ \text{is CSPD}
\quad\Longleftrightarrow\quad
\phi(\|x-y\|^2)\ \text{is CSND}.
\]

\noindent\paragraph{Step 3: When the Bernstein part is CSND.}
Lemma~\ref{lemma:Bernstein_CSND} proves this criterion on finite sets of points. For the general case required by Definition~\ref{def:cpd-cspd}, Theorem~4.1 of \cite{guella2021generalizationenergydistancebernstein} shows that \(\phi(\|x-y\|^2)\) is CSND when \(\Lambda\not\equiv 0\), and Lemma~\ref{lemma:Bernstein_CSND} already shows that it is not when \(\Lambda\equiv 0\). Hence
\[
\phi(\|x-y\|^2)\ \text{is CSND on }\R^d
\quad\Longleftrightarrow\quad
\Lambda\not\equiv 0.
\]
Combining Steps~1--3 yields
\[
\MMD_k \text{ is a metric}
\quad\Longleftrightarrow\quad
\Lambda\not\equiv 0,
\]
which proves the boxed statement.
\end{proof}

\end{document}